\documentclass[runningheads]{llncs}

\usepackage{accv}

\usepackage{accvabbrv}

\usepackage{graphicx}
\usepackage{booktabs}
\usepackage{algorithmic}
\usepackage{pifont}
\usepackage{amsfonts}

\usepackage{url}
\usepackage{listings}
\usepackage{tipa}

\newcommand{\cmark}{\ding{51}}
\newcommand{\xmark}{\ding{55}}

\usepackage{hyperref}

\usepackage[ruled]{algorithm2e}

\usepackage[accsupp]{axessibility}

\usepackage{orcidlink}
\usepackage[most]{tcolorbox}

\begin{document}

\title{Dynamics-Inspired Diffusion for Foreground-Preserving Document Background Editing} 

\titlerunning{Dynamics-Inspired Diffusion for Document Background Editing}

\author{Taewon Kang\inst{1}\orcidlink{0009-0004-8534-6314} \and
Yu Shen\inst{2}\orcidlink{0000-0002-9693-0485} \and
Ming C. Lin\inst{1}\orcidlink{0000-0003-3736-6949}}

\authorrunning{T.~Kang et al.}

\institute{University of Maryland at College Park, United States \and Adobe Research, United States \\
\email{taewon@umd.edu, shenyu@adobe.com, lin@umd.edu}}

\maketitle

\begin{abstract}
  We revisit diffusion-based generation for structured visual content and identify a fundamental limitation of existing approaches: foreground preservation and background stylization are typically enforced through external interventions, such as hard masking or corrective post-processing, rather than arising from the generative process itself. Here, we define \emph{background} as the generative content outside designated foreground regions (e.g., text and layout elements), while preserving the structural integrity of the foreground. We propose a dynamical systems perspective on diffusion, in which controllable generation is formulated as trajectory shaping in latent space. Under this view, we introduce \emph{Auxiliary Context Diffusion} (ACD), a state-space control framework that integrates heterogeneous signals (layout-derived foreground indicators, document summaries, and style representations) directly into the diffusion dynamics. This formulation induces time-scale separation in the generative process, where foreground regions become dynamically stabilized while background regions remain expressive. To address stylistic drift across multi-page documents, we further introduce \emph{style directions} as persistent latent constraints that guide diffusion trajectories within a shared stylistic subspace. Unlike prior approaches that entangle style with prompt conditioning, our formulation enables reusable and consistent style control across pages. We validate the proposed perspective through controlled experiments on synthetic document benchmarks, demonstrating that trajectory-level control provides a unified and extensible mechanism for structured generation without retraining, hard masking, or corrective post-processing. These results suggest a new direction for controllable diffusion in document-centric and multimodal applications. 
\end{abstract}

\section{Introduction}

Diffusion models have demonstrated remarkable capability in generating visually rich imagery, yet their application to structured visual artifacts such as documents reveals a fundamental limitation: controllable generation is typically achieved through external interventions rather than being expressed as a property of the generative process itself. In particular, structured constraints such as foreground preservation are commonly enforced through explicit mechanisms (e.g., masking, inpainting, or post-processing), rather than arising naturally from the underlying generative dynamics. In document-centric background generation, the challenge is not merely to produce aesthetically pleasing images, but to do so in a way that preserves foreground content and maintains stylistic consistency across pages. Here, we define \emph{background} as the generative content outside designated foreground regions (e.g., text and layout elements), while preserving the structural integrity of the foreground. Existing approaches often produce visually plausible results, yet suffer from stylistic drift across pages and rely on explicit mechanisms such as masking, attenuation, or post-processing to preserve foreground regions. A key reason for this behavior lies in how diffusion models are conditioned. Most pipelines rely on textual prompts to control both content and style, implicitly assuming that language provides a sufficient anchor in the generative space. In practice, however, text descriptions specify semantic intent but do not uniquely determine a location in latent space. As a result, repeated generations conditioned on identical prompts may traverse different regions of the latent manifold, leading to variations in color, texture, and composition that accumulate across pages. More critically, foreground preservation in prior work is treated as an external constraint, rather than as an intrinsic property of the generative dynamics. This perspective introduces heuristic mechanisms that obscure the underlying process and limit the controllability and compositionality of generation. In particular, many approaches rely on hard spatial masks that enforce preservation through binary exclusion, where generation is either suppressed or active within a region. While effective, such all-or-nothing treatments do not fully align with the continuous nature of diffusion trajectories, and can lead to sharp transitions near foreground boundaries. A natural alternative is to generate backgrounds independently and overlay the original foreground. While our pipeline likewise renders the original content onto the generated canvas, the critical distinction lies in how the canvas is produced: independently generated backgrounds are agnostic to layout geometry and per-page semantics, yielding content that collides visually with foreground regions and drifts stylistically across independently sampled pages.

In this work, we instead propose a dynamical systems perspective on diffusion, where generation is formulated as trajectory shaping in latent space. From this viewpoint, the central question is not how to suppress generation near foreground regions through external constraints, but how to design the generative dynamics such that foreground preservation emerges naturally from the evolution of the system. \textbf{This shifts the problem from explicit spatial exclusion to trajectory-level control, where structure is enforced through the evolution of the generative process itself.} We achieve this by introducing structured control signals that induce time-scale separation in the diffusion process, where designated foreground regions become dynamically stabilized while background regions remain expressive. To address stylistic inconsistency across pages, we further decouple style from prompt-based conditioning. We introduce \emph{style directions} as persistent latent constraints that define coherent stylistic axes. By constraining diffusion trajectories to evolve within a shared stylistic subspace, these directions enable consistent visual appearance across pages while avoiding ambiguity inherent in text-based style specification, \textbf{which is often underdetermined in latent space}. 

More broadly, we formulate our approach as \emph{Auxiliary Context Diffusion (ACD)}, a state-space control framework in which heterogeneous context elements (e.g., layout-derived foreground indicators, document summaries, style directions, and backing latents) are integrated directly into the diffusion dynamics. Unlike prior approaches that rely on learned conditioning or heuristic interventions, ACD provides a unified and training-free mechanism for compositional control. This formulation admits a geometric interpretation, where diffusion trajectories evolve on a latent manifold subject to structured constraints, and foreground preservation emerges from controlled stochastic dynamics rather than explicit exclusion. \textbf{Importantly, this formulation suggests that structured control in diffusion models can be achieved without hard masking or corrective post-processing, but instead through principled manipulation of generative dynamics.} By reframing controllable diffusion as trajectory-level control, our framework produces document backgrounds that are both foreground-preserving and stylistically consistent across pages, without relying on hard-masking interventions or post-processing heuristics. \textbf{More broadly, this perspective offers a complementary view of controllable diffusion, framing it as a problem of dynamical system design rather than constraint satisfaction, and we employ it as a unifying lens for organizing the proposed controls.} Furthermore, it suggests broader applicability of trajectory-level control in diffusion models beyond document-centric generation. To summarize, our key contributions are as follows:

\begin{enumerate}

\item \textbf{A Dynamical Systems Perspective on Controllable Diffusion.}  
We reinterpret diffusion-based generation as trajectory shaping in latent space, where controllability is achieved through structured state-space dynamics rather than external interventions. This perspective reframes foreground preservation as an emergent property of the generative process, providing a principled alternative to hard masking and corrective post-processing, with the foreground indicator $\mathbf{m}$ acting as a soft, time-gated control signal rather than a binary exclusion mask.

\item \textbf{Auxiliary Context Diffusion as a Unified Control Framework.}  
We introduce Auxiliary Context Diffusion (ACD), a training-free state-space control formulation that integrates heterogeneous signals including layout-derived foreground indicators, document summaries, and latent style representations directly into the diffusion dynamics. This formulation enables compositional and interpretable control within a single generative process.

\item \textbf{Persistent Style Constraints for Multi-Page Consistency.}  
We propose \emph{style directions} as persistent latent constraints that guide diffusion trajectories within a shared stylistic subspace. This decouples style from prompt-based conditioning and enables consistent visual generation across pages without repeated or ambiguous text specifications.

\item \textbf{Trajectory-Level Control as a General Mechanism for Structured Generation.}  
We demonstrate that trajectory-level control provides a general mechanism for structured generation, highlighting its applicability beyond document-centric tasks and suggesting broader applicability for controllable generative modeling.

\item \textbf{Strong Empirical Gains with Consistent User Preference.}  
Our method achieves consistent and substantial improvements across both automatic and human-aligned metrics, including more than \textbf{2× improvement} in multi-page consistency (CLIP), \textbf{an 8\% improvement} in OCR-based text preservation compared to GPT-5, and \textbf{a 21.5\% relative improvement} in accessibility (WCAG), with \textbf{86.9\%} of users preferring our results overall.
\end{enumerate}

\section{Related Work} 
\subsection{Design Generation and Document Editing Systems} 
Generative models have been widely applied to visual design tasks such as background synthesis, layout generation, and poster creation~\cite{eshratifar2024salient,li2023planning,inoue2024opencole, wang2025designdiffusion,huang2024layerdiff,dalva2024layerfusion, li2023relation,hu2024ella,luo2024layoutllm,yang2024mastering, weng2024desigen,chen2025posta,zhang2025creatiposter,peng2025bizgen, zhang2025creatidesign,wu2025hybrid}. These systems generate designs from scratch or rely on explicit layout planning, and thus differ fundamentally from document-centric background editing, where layout and foreground content are fixed and must be preserved~\cite{kang2025text}. Interactive frameworks such as POSTA~\cite{chen2025posta}, CreatiPoster~\cite{zhang2025creatiposter}, and InstructPix2Pix~\cite{brooks2023instructpix2pix} depend on prompt-heavy, human-in-the-loop workflows that scale poorly to multi-page documents, while grounded control frameworks (GLIGEN~\cite{li2023gligen}, LayoutDiffusion~\cite{zheng2023layoutdiffusion}, HiCo~\cite{ma2024hico}) still require active user intervention. General-purpose editors such as BAGEL~\cite{deng2025emerging}, GPT-4o~\cite{openai2024gpt4o}, and GPT-5~\cite{openai2025introducinggpt5} operate directly on document pages but frequently intrude into text regions, alter textual content, or drift stylistically across pages. The most closely related work~\cite{kang2025text} addresses our exact setting through three external mechanisms: a fixed attenuation mask, a post-hoc readability module that composites semi-transparent backing shapes to meet contrast targets, and per-page LLM-generated instructions for cross-page coherence. Our formulation subsumes these interventions into the diffusion dynamics themselves: preservation arises from time-dependent trajectory control rather than a fixed mask, near-ceiling readability is attained without corrective backing shapes, and cross-page consistency is enforced by persistent latent style directions rather than repeated LLM prompting. Table~\ref{tab:acd_vs_prior} (Appendix~\ref{appendix:positioning}) summarizes this per-axis comparison; as~\cite{kang2025text} provides no public implementation or data, the comparison is necessarily mechanism-level rather than empirical.

\subsection{Diffusion Models for Structured Editing and Layout Control} 
Applying diffusion to structured documents raises persistent challenges in text readability and cross-page consistency; classical studies~\cite{scharff2000discriminability,scharff2003contrast, leykin2004automatic,zuffi2007human,zuffi2009understanding} show that textured backgrounds significantly degrade legibility. TextDiffuser and TextDiffuser-2~\cite{chen2023textdiffuser,chen2024textdiffuser} model text placement and SAWNA~\cite{10.1145/3721250.3743023} preserves empty regions, but both assume controllable text rendering or blank space rather than existing dense foreground. Relative to blank-region-preserving generation (SAWNA), the concrete deltas of our formulation are threefold: we operate on existing dense-foreground pages rather than reserved blank regions, apply time-dependent soft control rather than fixed nonreactive regions, and maintain persistent cross-page style directions. Mask-guided editing~\cite{avrahami2022blended,couairon2022diffedit,lugmayr2022repaint, ju2024brushnet} restricts edits spatially but degrades content under imperfect masks; attention-based control~\cite{hertz2022prompt,cao2023masactrl,chefer2023attend} and layered diffusion~\cite{huang2024layerdiff,li2023layerdiffusion, zhang2024transparent} assume clean layer separation that rarely holds in real documents; spatial conditioning~\cite{sun2024spatial,Chen_2024_WACV,Xie_2023_ICCV} relies on strong external signals without guaranteeing preservation. In contrast, our method constrains diffusion dynamics through latent state-space control, stabilizing sensitive regions while keeping the background expressive. 

\subsection{Content--Style Disentanglement in Generative Models} 
GAN-era work showed that latent directions encode interpretable stylistic attributes~\cite{shen2020interpreting,harkonen2020ganspace}. In diffusion models, StyleAligned~\cite{hertz2024style}, InST~\cite{zhang2023inversion}, and IP-Adapter~\cite{ye2023ip} improve style control by injecting external signals, but require reference images or added conditioning and do not extend to multi-page settings where style must remain fixed across independently generated pages; SDE-based frameworks~\cite{sohl2015deep} model global dynamics without persistent, reusable style constraints. We instead represent style as a persistent latent direction that directly modulates the trajectory, enabling reuse across pages without repeated prompting or parameter changes. Table~\ref{tab:method_comparison} (Appendix~\ref{appendix:positioning}) positions prior categories against the five simultaneous constraints of our task (operating on instantiated pages, soft foreground control, cross-page consistency, dense-text fidelity, and no retraining), none of which any existing category satisfies in full.

\section{Method}
\label{sec:method}

\begin{figure*}[t]
    \centering
    \includegraphics[width=0.9\linewidth]{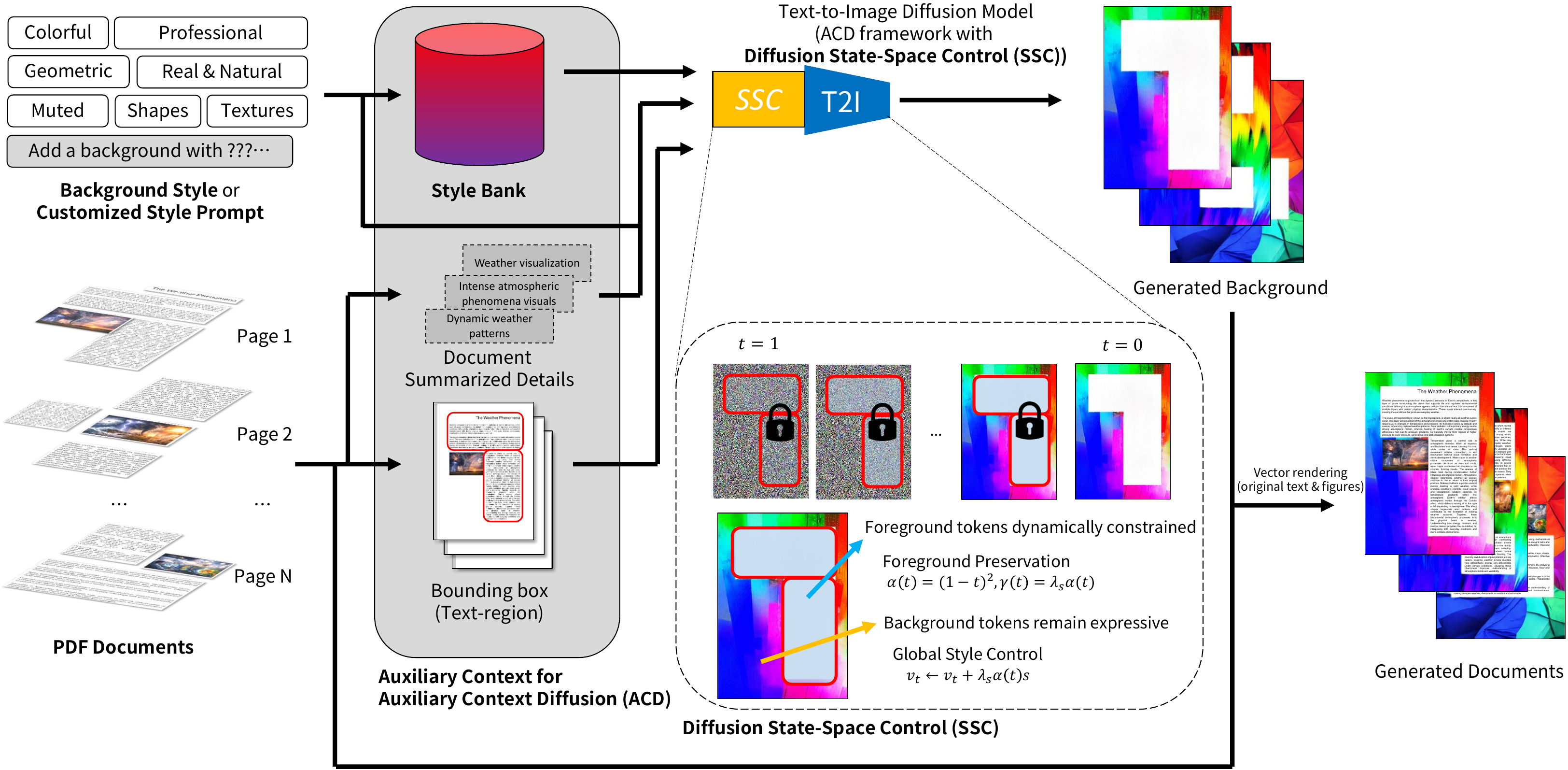}
    \caption{\textbf{Overview of Auxiliary Context Diffusion (ACD).}
    Layout-derived foreground regions, document summaries, and style directions are integrated as auxiliary controls that shape the diffusion trajectory, enabling foreground preservation and cross-page consistency.}
    \label{fig:method_figure_doc}
\end{figure*}

We formulate foreground-aware background generation as a problem of \emph{trajectory-level control} in diffusion models, as illustrated in Figure~\ref{fig:method_figure_doc}. Our pipeline generates a background \emph{canvas}: the diffusion process synthesizes background content while driving latent regions beneath foreground bounding boxes toward a stable, low-variance state, so that the canvas remains neutral precisely where document content resides. The final document is then produced by standard vector rendering, placing the original, unmodified text and figures onto the generated canvas; the original foreground is never re-synthesized, encoded into the latent, or altered. What our formulation eliminates is therefore not this rendering step, which is common to all document workflows, but \emph{hard-masked generation and corrective post-processing} (e.g., binary exclusion during sampling, boundary harmonization, or contrast adjustment applied after generation). Rather than relying on such interventions, we modify the diffusion dynamics such that foreground regions of the canvas become progressively stabilized while background regions remain expressive. This formulation shifts the focus from explicit spatial constraints to the design of generative dynamics.

\paragraph{Latent Representation.}
Let $x_t \in \mathbb{R}^{N \times d}$ denote the packed latent tokens at diffusion time $t \in [0,1]$, where $N$ is the number of visual tokens and $d$ is the latent dimensionality. We define a binary foreground indicator:
\begin{equation}
\mathbf{m} \in \{0,1\}^{N \times 1},
\end{equation}
where $\mathbf{m}_k = 1$ denotes tokens overlapping foreground regions. Unlike hard spatial masks used for explicit exclusion, our foreground indicator $\mathbf{m}$ acts as a soft, time-dependent control signal that modulates the diffusion dynamics. A hard mask enforces preservation by preventing updates within foreground regions, which can reduce interaction between foreground and background during generation and introduce sharp transitions near their boundaries. In contrast, $\mathbf{m}$ attenuates foreground updates progressively through the gating function $\alpha(t)$, allowing the degree of stabilization to increase gradually over time. This enables foreground preservation to emerge from the generative dynamics themselves, rather than being imposed as an external constraint, and supports smoother and more context-aware integration between foreground and background. Foreground indicators are derived from document layout analysis using off-the-shelf text detection and PDF parsing tools, producing bounding boxes that are rasterized into the binary mask $\mathbf{m}$.

\subsection{Auxiliary Context Diffusion}

We express diffusion as a controlled dynamical system:
\begin{equation}
\frac{d x_t}{dt}
=
- v_\theta(x_t,t)
+
\sum_{i} g_i \, u_i(x_t,t),
\label{eq:acd_core_final}
\end{equation}

where $v_\theta$ denotes the pretrained diffusion drift, $u_i$ are auxiliary control terms, and $g_i \geq 0$ are scalar gains weighting each control. In our instantiation:
\begin{align}
u_{\text{style}}(x_t,t) &= \lambda_s\,\alpha(t)\,s, \\
u_{\text{fg}}(x_t,t) &= - v_\theta(x_t,t) \odot \alpha(t)\mathbf{m}, \\
u_{\text{stab}}(x_t,t) &= \gamma(t)(b - x_t).
\end{align}

Here, $s \in \mathbb{R}^{d}$ is a normalized latent style direction applied with strength $\lambda_s$ (Sec.~3.2); $\alpha(t) = (1-t)^2$ is a late-emphasis gate that grows as sampling proceeds toward $t=0$; $\mathbf{m}$ is the foreground indicator defined above; $\odot$ denotes token-wise multiplication; $b$ is a fixed reference latent set to the latent-prior mean $b=0$ (Sec.~\ref{sec:stabilization}); and $\gamma(t) = \lambda_s\,\alpha(t)$ modulates the stabilization strength. This decomposition clarifies how style, structure, and stabilization jointly shape the diffusion trajectory. In discrete time, $u_{\text{style}}$, $u_{\text{fg}}$, and $u_{\text{stab}}$ are instantiated, respectively, as the additive steering of Eq.~\ref{eq:style_steer}, the multiplicative gating of Eq.~\ref{eq:fg_gating}, and the relaxation of Eq.~\ref{eq:stab_relax}, applied per step in Algorithm~1 (lines~7, 8, and~10).

\subsection{Style Bank}

We define a Style Bank:
\begin{equation}
\mathcal{S} = \{ s_i \in \mathbb{R}^{d} \mid i = 1, \dots, K \},
\end{equation}

where each $s_i$ is a normalized latent direction representing a visual style. Style directions are constructed offline using a small set of descriptive prompts and remain fixed during inference. Given a selected style:
\begin{equation}
v_t \leftarrow v_t + \lambda_s\,\alpha(t)\,s.
\label{eq:style_steer}
\end{equation}

The late-emphasis gate $\alpha(t)$ concentrates stylistic steering in the content-forming phase of sampling, and a norm-preserving rescaling of $v_t$ prevents velocity magnitude drift. For user-specified prompts beyond predefined categories, a style direction can be constructed on-the-fly using the same offline aggregation procedure.
\subsection{Diffusion as a State-Space Dynamical System}

Building on the controlled dynamics of Eq.~\ref{eq:acd_core_final}, we instantiate the foreground control with the time-dependent gating function:

\begin{equation}
\alpha(t) = (1 - t)^2,
\end{equation}

and modulate the velocity field:
\begin{equation}
v_t = v_\theta(x_t, t) \odot \bigl(1 - \alpha(t)\mathbf{m}\bigr).
\label{eq:fg_gating}
\end{equation}

\paragraph{Time-scale Separation.}
The resulting dynamics induce anisotropy:
\begin{equation}
\left\lVert \frac{d x_t^{(k)}}{dt} \right\rVert \ll 
\left\lVert \frac{d x_t^{(j)}}{dt} \right\rVert,
\quad \forall k \in \mathcal{I}_{\text{fg}},\ j \in \mathcal{I}_{\text{bg}}.
\end{equation}

This ensures that foreground tokens evolve on a slower time scale than background tokens.

\subsection{Latent Stabilization}
\label{sec:stabilization}

We further provide an energy-based interpretation, viewing standard 
diffusion as descending a data-plus-noise energy landscape. We introduce 
a foreground potential:
\begin{equation}
\mathcal{E}_{\text{fg}}(x_t) = \sum_{k:\mathbf{m}_k=1} \lVert x_t^{(k)} - b \rVert^2,
\end{equation}

where $b$ is a fixed reference latent (mean latent prior). In our instantiation we set $b = 0$, which is precisely the mean of the latent prior $\mathcal{N}(0, I)$ from which sampling is initialized. Contracting foreground-region tokens toward $b$ therefore does not erase content, as no document content is ever encoded in these tokens (see Sec.~\ref{sec:method}); rather, it anchors the canvas beneath foreground regions to a neutral, low-variance state onto which the original content is rendered with maximal legibility.

This yields the update:
\begin{equation}
x_t^{(k)} \leftarrow (1 - \gamma(t))x_t^{(k)} + \gamma(t)b,
\label{eq:stab_relax}
\end{equation}

with
\begin{equation}
\gamma(t) = \lambda_s \alpha(t).
\end{equation}

This relaxation drives foreground regions toward a stable, low-variance state. A geometric interpretation of the style directions, a temperature-based reading of this stabilization, and the corresponding formal analysis are provided in Appendix~\ref{sec:supp_proof}.

\begin{algorithm}[t]
\small
\caption{Auxiliary Context Diffusion (ACD)}
\begin{algorithmic}[1]
\STATE Initialize latent $x_T \sim \mathcal{N}(0,I)$
\STATE Set $b = 0$ \COMMENT{mean of the latent prior; see Sec.~\ref{sec:stabilization}}
\STATE Extract foreground mask $\mathbf{m}$
\STATE Select style direction $s$
\FOR{$t = T \rightarrow 0$}
    \STATE $v \leftarrow v_\theta(x_t, t)$
    \STATE $v \leftarrow v + \lambda_s \alpha(t) s$ \COMMENT{late-gated style steering}
    \STATE $v \leftarrow v \odot (1 - \alpha(t)\mathbf{m})$
    \STATE $x_t \leftarrow x_t - v \Delta t$
    \STATE $x_t^{(\text{fg})} \leftarrow (1-\gamma(t))x_t^{(\text{fg})} + \gamma(t)b$
\ENDFOR
\STATE \textbf{return} $x_0$
\end{algorithmic}
\end{algorithm}

\subsection{Unified View}

Our method combines style injection, foreground-aware modulation, and latent stabilization into a unified trajectory-level control mechanism. All operations are training-free and require no architectural modification. We emphasize that the dynamical-systems, geometric, and thermodynamic views developed above serve as a unifying lens for organizing these operations rather than as derived theory; the concrete instantiations are the gating, steering, and relaxation updates of Eqs.~\ref{eq:style_steer}--\ref{eq:stab_relax} and Algorithm~1. More broadly, ACD provides a general framework for incorporating heterogeneous context signals into diffusion dynamics, enabling controllable and compositional generation.

\begin{figure}[t]
    \centering
    \includegraphics[width=0.4\linewidth]{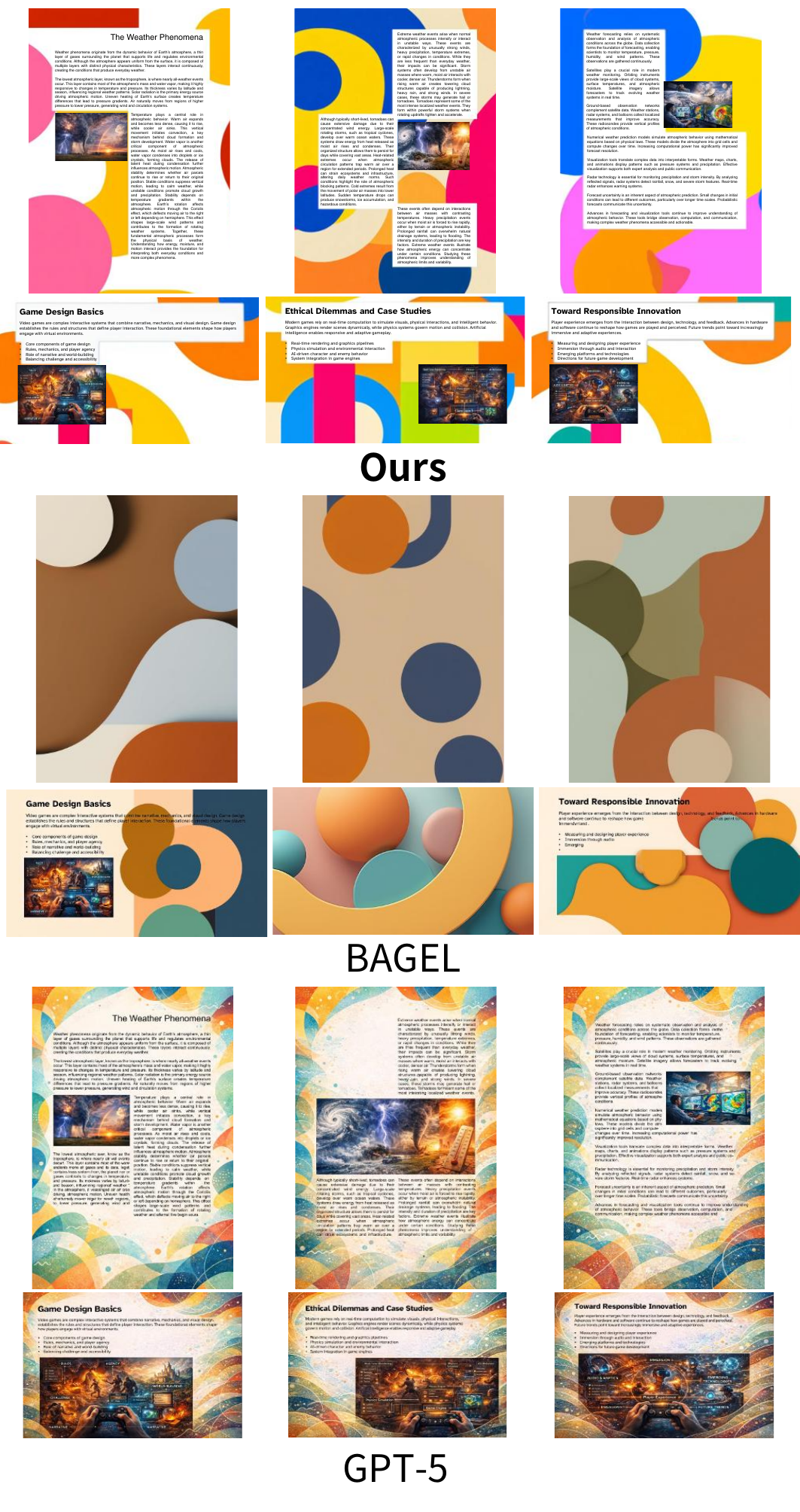}
    \caption{ \textbf{Representative qualitative comparison on document background editing.} We compare \textbf{Ours}, \textbf{BAGEL}, and \textbf{GPT-5} under different style conditions (rows). \textbf{BAGEL} often fails to preserve foreground text, as background textures overlap with text regions and reduce readability. \textbf{GPT-5} produces visually appealing designs but violates the editing constraint by altering the original layout (e.g., shifting text blocks), modifying or replacing existing figures, and introducing additional content not present in the input. In contrast, \textbf{Ours} preserves both layout and text fidelity while applying consistent background styles. \emph{For clarity, readers may focus on text regions and figure boundaries to observe layout preservation and content integrity.}}
    \label{fig:results_representative}
\end{figure}

\begin{figure}[t]
    \centering
        \includegraphics[width=0.38\linewidth]{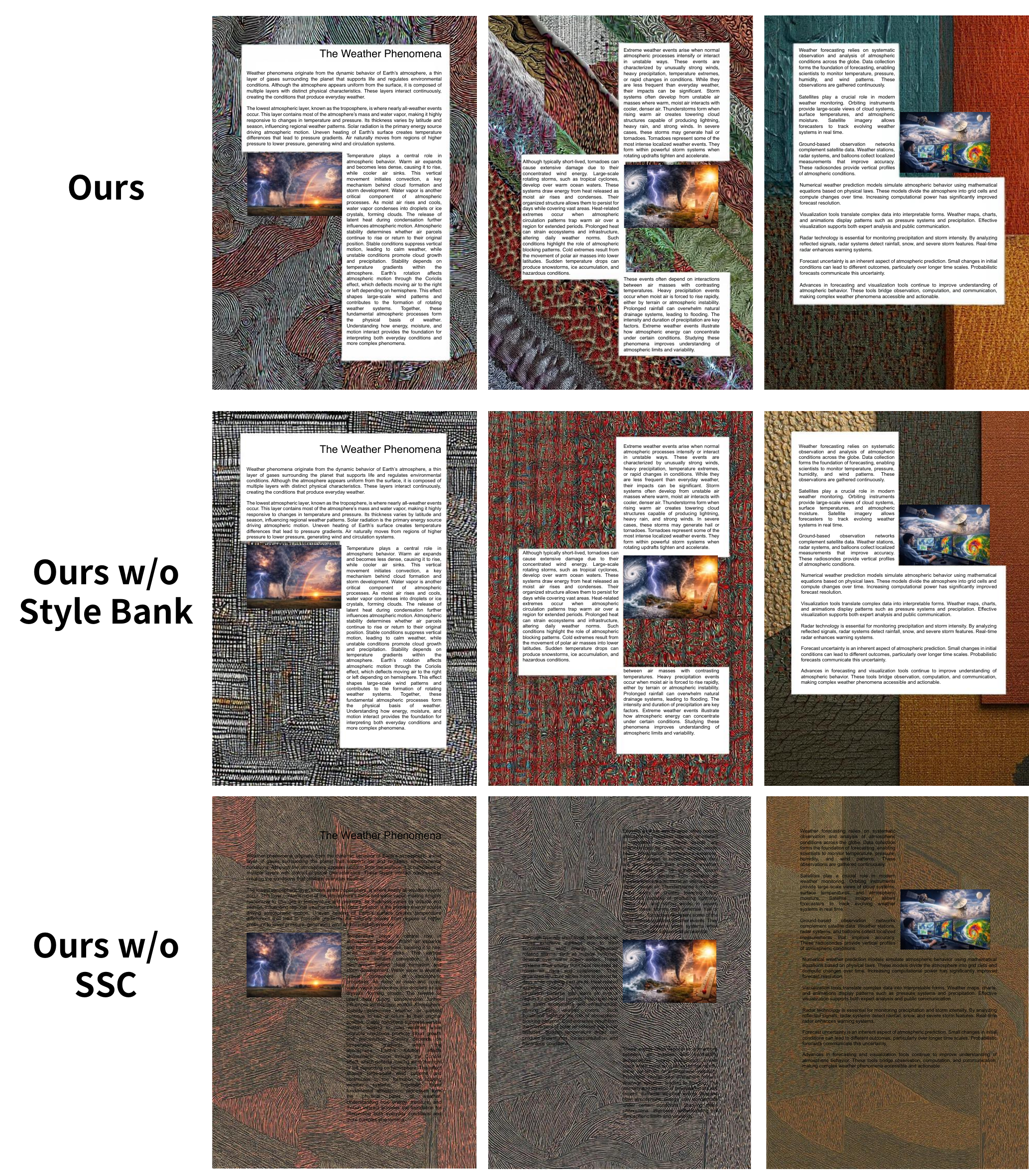}
    \caption{ \textbf{Ablation study on foreground-aware background generation.} \textbf{Ours} (top row) preserves text readability and layout fidelity while maintaining consistent visual styles across pages. Removing the Style Bank (\textbf{middle row}) leads to noticeable inconsistencies in background appearance across pages, despite preserving text readability. Disabling SSC (\textbf{bottom row}) causes background textures to intrude into text regions, significantly degrading readability and accessibility. \emph{Readers are encouraged to compare corresponding columns across rows, particularly focusing on text regions and cross-page style consistency.} }
    \label{fig:ablation}
\end{figure}

\begin{table*}[t]
\centering
\small
\resizebox{\textwidth}{!}{%
\begin{tabular}{lccccccccc}
    \toprule
    \textbf{Method} 
    & \textbf{Layout} $\uparrow$ 
    & \textbf{Color} $\uparrow$ 
    & \textbf{Graphic Style} $\uparrow$ 
    & \textbf{Compliance} $\uparrow$ 
    & \textbf{WCAG} $\uparrow$ (\%) 
    & \textbf{OCR Acc.} $\uparrow$ 
    & \textbf{CLIP MP Consistency} $\uparrow$ 
    & \textbf{CLIP Prompt Score} $\uparrow$ 
    & \textbf{LLM Voting} $\uparrow$ \\
    \midrule
    BAGEL & 3.7335 & 3.9735 & 3.8857 & 3.7292 & 82.95 & 0.363 & 0.6317 & 0.1567 & 3.8478 \\
    GPT-5 & 4.155 & \textbf{4.3272} & \textbf{4.321} & 4.2614 & 80.75 & 0.7217 & 0.3113 & 0.1657 & 4.2363 \\
    GPT-5 (Naive Overlay) & 3.8928 & 4.0486 & 4.0778 & 4.13 & 56.87 & 0.3077 & 0.6085 & 0.2631 & 3.9943 \\
    \textbf{Ours} & \textbf{4.24} & 4.07 & 4.14 & \textbf{4.74} & \textbf{98.12} & \textbf{0.779} & \textbf{0.6785} & \textbf{0.3144} & \textbf{4.2992} \\
    \midrule
    Ours w/o Style Bank & 4.1878 & 3.9492 & 4.055 & 4.70 & 98.08 & 0.7769 & 0.6661 & 0.3083 & 4.2335 \\
    Ours w/o SSC & 3.6885 & 3.7142 & 3.8442 & 4.2785 & 54.20 & 0.333 & 0.6667 & 0.2781 & 3.8764 \\
    \bottomrule
\end{tabular}}
\caption{\textbf{Quantitative results on document background generation evaluated by nine criteria.}  We report both human-aligned preference scores and automatic measurements to assess overall layout quality, color harmony, graphic style suitability, and guideline compliance, together with accessibility- and readability-focused metrics. Specifically, LLM-based ratings (Layout, Color, Graphic Style, Compliance, and LLM Voting) are aggregated on a 1--5 scale, while automatic metrics quantify WCAG contrast coverage, OCR-based text preservation, multi-page visual consistency, and prompt alignment. Higher values indicate better performance across all metrics. \emph{GPT-5 (Naive Overlay)} denotes the independent-generation-plus-overlay baseline: canvases generated by GPT-5 from the identical fixed style prompt per category, independently sampled per page, without layout awareness, document summaries, or the Style Bank, and composited with the original foreground via the same vector-rendering step as Ours (Sec.~\ref{sec:quantitative}, Appendix~\ref{appendix:naive_overlay}).}
\label{tab:quantitative_results}
\end{table*}

\section{Experiments and Results}

The implementation of this work follows a diffusion-based inference pipeline~\cite{deng2025emerging}, with all foreground preservation and style control realized through our Auxiliary Context Diffusion framework.

\subsection{Benchmarking Document Datasets}
\label{sec:dataset}

We evaluate our framework on a curated benchmark of academic-style documents and presentation slides designed to reflect realistic educational materials. Each instance consists of exactly three pages or slides with a coherent narrative progression, enabling evaluation of both single-page quality and multi-page consistency under semantic transitions. Pages combine long-form text, structured bullet points, and at least one figure arranged in non-trivial layouts, introducing layout diversity while preserving clear foreground structure. The benchmark is intentionally self-constructed and synthetic: all content is authored or generated specifically for this benchmark, ensuring reproducibility and \textbf{compliance with copyright requirements}, enabling precise control over layout structure, text density, and page-level semantic progression, and avoiding confounding factors such as OCR noise and document degradation. {\em This design isolates the core challenges of background generation}: foreground legibility, layout preservation, and multi-page stylistic coherence. The current benchmark consists of three-page documents; evaluation on longer sequences and on real-world documents remains complementary future work. The full corpus composition, the legal and reproducibility constraints motivating the synthetic design, and the intended scope of evaluation are detailed in Appendix~\ref{appendix:dataset_new} and Appendix~\ref{appendix:dataset_rationale}.

\subsection{Qualitative Results}

We qualitatively compare our method with two representative document editing baselines, BAGEL~\cite{deng2025emerging} and GPT-5~\cite{openai2025introducinggpt5}. Unlike poster generation or layout synthesis approaches, our task focuses on \emph{editing existing document pages}, where foreground text regions and layout are fixed and must be strictly preserved. This setting fundamentally differs from design-from-scratch pipelines such as POSTA and CreatiPoster, which assume flexible layouts and sparse text, and are therefore not directly applicable to document editing scenarios with fixed foreground content. Accordingly, we restrict comparison to methods that can directly operate on existing document pages without re-synthesizing layout or text; we summarize this conceptual positioning in Table~\ref{tab:method_comparison} (Appendix~\ref{appendix:positioning}), and discuss the excluded masking- and removal-based method families in Appendix~\ref{appendix:qualitative}. A simple alternative is to generate a background independently and overlay the original foreground; since the original content is preserved by construction, such a pipeline constitutes an oracle upper bound on text fidelity. It is, however, structurally blind to the document: the background is generated without awareness of layout geometry or page-level semantics, and independently sampled pages exhibit stylistic drift. Our formulation targets precisely the axes that direct overlay cannot address: layout-aware canvas stabilization, semantic grounding via document summaries, and persistent style directions for cross-page consistency. We evaluate this overlay pipeline empirically as \emph{GPT-5 (Naive Overlay)} in Table~\ref{tab:quantitative_results} and Sec.~\ref{sec:quantitative}, with qualitative examples in Appendix~\ref{appendix:naive_overlay}.
Figures~\ref{fig:results_representative} show representative qualitative comparisons on academic-style PDFs (A4) and slides (16:9). Backgrounds are generated by reflecting the textual content already present in each page, ensuring that visual motifs are semantically aligned with the document content while remaining non-intrusive to foreground text. As illustrated in Fig.~\ref{fig:results_representative}, our method consistently preserves readability and layout fidelity across styles, whereas BAGEL frequently introduces high-frequency textures or strong patterns that interfere with text regions. GPT-5 produces visually appealing backgrounds, but may alter layout structure or introduce additional content, violating the document editing constraint by modifying existing figures or hallucinating new text. Figures~\ref{fig:results_pdfs} and~\ref{fig:results_slides} further demonstrate that our framework maintains stylistic coherence across a broad range of styles, including Geometric, Muted, Professional, and Real \& Natural. Notably, our method better preserves multi-page consistency, with background motifs and color palettes evolving smoothly across pages without abrupt changes.

\subsection{Quantitative Evaluation}
\label{sec:quantitative}

We evaluate our framework against BAGEL and GPT-5, following prior document background generation protocols. All metrics are computed at the document level, with higher values indicating better performance. Metrics are grouped into three categories: (i) Design Quality (Layout, Color, Graphic Style, Compliance), (ii) Readability (WCAG contrast coverage, OCR accuracy), and (iii) Multi-page Consistency (CLIP-based similarity, LLM-based voting). Design scores are obtained via an LLM-based evaluator (GPT-4o), following prior work in document design assessment~\cite{kang2025text,zhang2025creatiposter}, where LLM judgments have been shown to correlate with human preference. Readability and consistency are measured automatically. As shown in Table~\ref{tab:quantitative_results}, our method achieves the best overall performance, leading on seven of the nine metrics, while GPT-5 leads on Color and Graphic Style. Our method attains the highest Compliance score, near-perfect WCAG coverage (98.12\%), and the highest OCR accuracy. It also significantly outperforms both baselines in multi-page consistency, particularly in CLIP-based measures, indicating stronger document-level visual coherence. Additional analyses are provided in Appendix~\ref{appendix:quantitative}.

\paragraph{Naive overlay baseline.}
To isolate the contribution of the pipeline structure from that of the proposed controls, Table~\ref{tab:quantitative_results} includes the independent-generation-plus-overlay baseline, \emph{GPT-5 (Naive Overlay)}, evaluated on the full benchmark under the identical protocol. The overlay structure alone recovers much of the consistency over full-page editing (CLIP MP $0.3113 \rightarrow 0.6085$), yet a substantial gap to Ours remains ($0.6085$ vs.\ $0.6785$, which is $5.6\times$ the \emph{w/o Style Bank} delta), and readability collapses without layout awareness (WCAG $56.87\%$, OCR $0.3077$). A detailed decomposition of this gap and qualitative examples are provided in Appendix~\ref{appendix:naive_overlay}.

\paragraph{LLM judge vs.\ human raters.}
The GPT-4o judge favors GPT-5 on the appearance-only axes (Color and Graphic Style), whereas the 30 human raters, who were shown the original document alongside each result, overwhelmingly weighted document usability, i.e., layout preservation and readability (86.9\% overall preference; Sec.~\ref{sec:user_study}). The judge-free, deterministic readability metrics (WCAG, OCR) side with the human raters. The disagreement thus reflects a difference in what is being judged (the aesthetics of an image versus the fitness of a document) rather than an inconsistency in the evaluation. We therefore designate the human study and the automatic readability metrics as primary evidence, and the LLM-based design scores as secondary.

\begin{figure}[thb]
    \centering
    \includegraphics[width=0.6\linewidth]{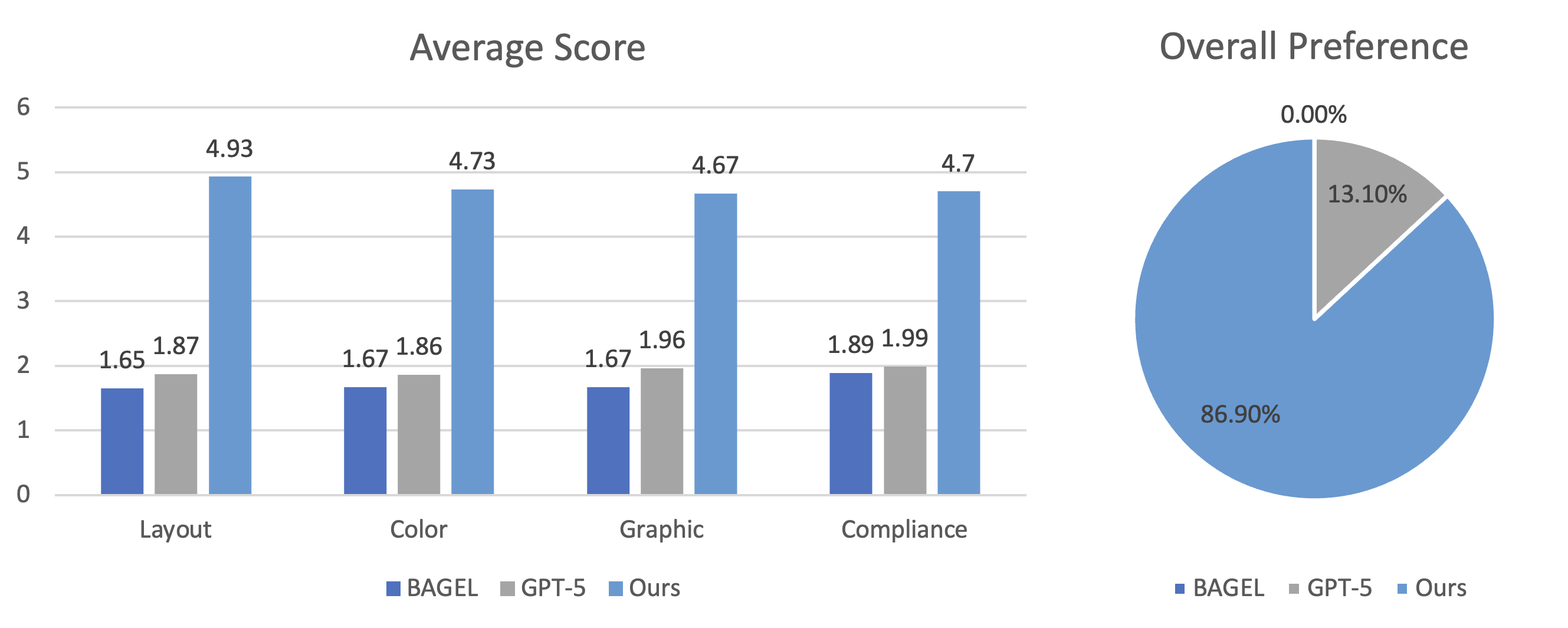}
    \caption{
    \textbf{Results of the user study.} Thirty participants compared three anonymized background generation methods across four evaluation criteria: Layout preservation, Color harmony, Graphic style consistency, and Prompt compliance (left). Our approach consistently received the highest average ratings across all dimensions. In the overall preference comparison (right), \textbf{86.9\%} of votes favored our method, compared to 13.1\% for GPT-5.}
    \label{fig:user_study}
\end{figure}

\subsection{User Study}
\label{sec:user_study}

We conducted a user study with 30 participants to evaluate document-centric background generation quality. For each task, participants were shown the original document along with three anonymized results generated by BAGEL, GPT-5, and \emph{Ours}. The presentation order was randomized. Participants rated each output on four criteria (\emph{Layout preservation}, \emph{Color harmony}, \emph{Graphic style consistency}, and \emph{Prompt compliance}) using a 5-point Likert scale, and selected an overall preferred result. As shown in Fig.~\ref{fig:user_study}, our method achieved the highest mean scores across all criteria (4.67–4.93), substantially outperforming BAGEL and GPT-5. In the overall preference evaluation, \textbf{86.9\%} of all votes favored \emph{Ours}, indicating strong user preference. Additional details are provided in Appendix~\ref{appendix:user_study}.

\subsection{Ablation Study}
\label{subsec:ablation}

We validate the contributions of our two core components: the \emph{Style Bank} and \emph{Diffusion State-Space Control (SSC)}. As shown in Fig.~\ref{fig:ablation}, Ours preserves foreground readability while producing visually coherent backgrounds with consistent style across pages. When removing the \textbf{Style Bank} (Ours w/o Style Bank), readability remains largely intact, but cross-page stylistic coherence degrades due to page-wise style drift. In contrast, removing \textbf{SSC} (Ours w/o SSC) leads to severe foreground intrusion and readability collapse, as background textures spill into text regions. These qualitative trends are consistent with Table~\ref{tab:quantitative_results}: removing the Style Bank mainly reduces multi-page consistency (CLIP MP Consistency $0.6785 \rightarrow 0.6661$), whereas removing SSC causes significant accessibility degradation (WCAG $98.12\% \rightarrow 54.20\%$, OCR $0.779 \rightarrow 0.333$). A detailed analysis is provided in Appendix~\ref{appendix:ablation}. We further observe that foreground-background transitions remain stable without requiring explicit boundary harmonization or post-processing, as shown in Fig.~\ref{fig:ablation}. While overlay pipelines whose backgrounds are generated without knowledge of foreground geometry may exhibit visual discontinuities at box boundaries, our trajectory-level control shapes the canvas around foreground regions during generation, reducing boundary artifacts. We further note that the auxiliary control terms in ACD do not impose measurable computational overhead during inference (Appendix~\ref{appendix:compute}).

\section{Conclusions}
We present a diffusion-based framework for foreground-preserving and stylistically consistent background generation in multi-layer, multi-page documents. Interpreting diffusion as a controllable stochastic process in a structured latent space, we decouple style from text conditioning through state-space control: cached style directions constrain trajectories to a shared stylistic subspace, ensuring cross-page consistency without repeated prompts. Beyond practical improvements, our formulation offers a new perspective on controllable diffusion. Unlike prior approaches that couple controllability to learned conditioning or heuristic interventions, it introduces explicit trajectory-level control mechanisms enabling persistent, reusable constraints across generations. The geometric interpretation clarifies how style directions induce structured latent subspaces for consistent generation, while the thermodynamic analogy offers intuition for why foreground stabilization emerges progressively during denoising, together providing a unifying, interpretable lens on structured control within diffusion dynamics. More broadly, we formulate this approach as \emph{Auxiliary Context Diffusion}, where optional context cues (layout information, document summaries, and style directions) modulate a shared diffusion process via interpretable controls, providing a principled, training-free foundation for document design. Empirically, this trajectory-level control yields strong improvements in foreground preservation, readability, and multi-page consistency over existing baselines. \textbf{Limitations and Future Works are in Appendix~\ref{sec:limitation}.}

\bibliographystyle{splncs04}
\bibliography{main}

\newpage
\appendix
\onecolumn

\section{Appendix}

\subsection{Ethics Statement}
\label{supplementary:ethics}

\begin{tcolorbox}[breakable, title=Ethics Statement]
All documents and slides used in our experiments, including both textual content and visual elements, were synthetically generated using large-scale foundation models (GPT-4o for multimodal image generation and GPT-5 for text generation). No external images, web-scraped data, or third-party copyrighted materials were incorporated.

The resulting dataset contains no personal, sensitive, or identifiable information related to real individuals, and does not include confidential or private documents. While the dataset itself does not involve human subjects, we conducted a separate user study to assess usability and preference of the generated document backgrounds. This study was reviewed and granted \textbf{IRB exemption as minimal-risk research} by our institution, confirming compliance with established human-subjects research ethics.

Content topics were intentionally selected to be non-sensitive and non-harmful (e.g., physics, engineering, media technology, design, and everyday systems). During document editing, the proposed system is designed to avoid introducing misleading factual information or altering the semantic intent of the original document. In particular, our research performs background synthesis exclusively, \emph{without modifying, obscuring, or removing existing text or foreground content}, thereby preserving readability and authorial intent.

We recognize that document background manipulation technologies could be misused, for instance to deceptively alter the appearance of documents. To mitigate such risks, our framework explicitly restricts edits to background regions, disallows direct manipulation of textual content, and assumes transparent user-driven editing workflows. \textbf{Overall, we intentionally constructed a fully synthetic dataset to proactively avoid potential ethical and legal concerns associated with real-world documents, including copyright restrictions, data ownership ambiguity, and unintended disclosure of sensitive information.} This design choice ensures clear data provenance and enables controlled, reproducible experimentation without reliance on externally sourced or copyrighted materials. Our dataset construction, experimental design, and system constraints therefore adhere to responsible AI research principles, including respect for intellectual property, transparency, and proactive mitigation of ethical risks.
\end{tcolorbox}

\subsection{Limitations and Future Work}
\label{sec:limitation} 
Soft, trajectory-level control may introduce minor artifacts near complex or densely packed boundaries; style directions are fixed once selected, limiting gradual or hierarchical style evolution; and the benchmark covers three-page synthetic documents, leaving longer sequences and real-world document collections to future evaluation; while the style-direction mechanism itself is length-agnostic (a fixed latent direction with no inter-page state), its behavior on substantially longer documents remains to be verified empirically. Future work may extend the framework to other display-oriented settings such as posters, infographics, comics, portraits, and structured compositions where foreground elements must remain stable under generative modification. Viewing diffusion as a controllable dynamical system opens new opportunities for reliable generative design, including interactive and iterative editing scenarios.

\subsection{Theoretical Analysis and Proofs}
\label{sec:supp_proof}

This supplementary material provides a formal theoretical analysis supporting the Method in the main paper. 

\subsubsection{Preliminaries and Notation}
We consider the continuous-time diffusion ODE
\begin{equation}
\frac{d x_t}{dt} = - v_\theta(x_t,t), \quad t \in [0,1],
\end{equation}
where $x_t \in \mathbb{R}^{N \times d}$ denotes packed latent tokens. 
Let $\mathbf{m} \in \{0,1\}^{N}$ be the foreground indicator and define index sets 
$\mathcal{I}_{\text{fg}}$ and $\mathcal{I}_{\text{bg}}$ accordingly.

We assume $v_\theta$ is locally Lipschitz in $x$ for fixed $t$, which ensures existence and uniqueness of trajectories.

\subsubsection{Controlled Dynamics and Time-Scale Separation}
\label{subsec:proof_timescale}

We consider the foreground-gated dynamics
\begin{equation}
\frac{d x_t}{dt} = - v_\theta(x_t,t) \odot (1 - \alpha(t) \mathbf{m}),
\end{equation}
with $\alpha(t) = (1-t)^2$.

\paragraph{Proposition 1 (Foreground Time-Scale Separation).}
For any $k \in \mathcal{I}_{\text{fg}}$ and $j \in \mathcal{I}_{\text{bg}}$, the ratio of update magnitudes satisfies
\begin{equation}
    \frac{\lVert \dot{x}_t^{(k)} \rVert}{\lVert \dot{x}_t^{(j)} \rVert} \le 1-\alpha(t),
\end{equation}
and therefore
\begin{equation}
    \lim_{t \to 0} 
    \frac{\lVert \dot{x}_t^{(k)} \rVert}{\lVert \dot{x}_t^{(j)} \rVert} = 0.
\end{equation}

\paragraph{Derivation.}
By construction, foreground velocities are scaled by the factor $(1-\alpha(t))$, while background velocities are unaffected:
\begin{align}
\dot{x}_t^{(k)} &= - (1-\alpha(t)) v_\theta(x_t^{(k)},t), \\
\dot{x}_t^{(j)} &= - v_\theta(x_t^{(j)},t).
\end{align}
Since $\alpha(t) \to 1$ as $t \to 0$, the induced dynamics exhibit explicit time-scale separation, with foreground states evolving on a slower manifold.

\subsubsection{Stability of Foreground Latents}
\label{subsec:proof_stability}

Foreground stabilization applies the relaxation
\begin{equation}
x_t^{(k)} \leftarrow (1-\gamma(t)) x_t^{(k)} + \gamma(t) b,
\end{equation}
with $\gamma(t)=\lambda_s \alpha(t)$.

\paragraph{Proposition 2 (Asymptotic Convergence).}
For any $k \in \mathcal{I}_{\text{fg}}$, the foreground latent converges to the backing latent $b$ as $t \to 0$.

\paragraph{Stability Argument.}
Define the Lyapunov function
\begin{equation}
V(x_t^{(k)}) = \lVert x_t^{(k)} - b \rVert^2.
\end{equation}
Under the relaxation update,
\begin{equation}
V(x_{t-\Delta t}^{(k)}) = (1-\gamma(t))^2 V(x_t^{(k)}),
\end{equation}
which is strictly decreasing for $\gamma(t) \in (0,1)$. 
Hence, $b$ is an asymptotically stable equilibrium for all foreground states.

\subsubsection{Invariance of Style-Induced Submanifolds}
\label{subsec:proof_style}

From a geometric perspective, the accumulated effect of style steering is a consistent displacement $\Phi_s^{\lambda}(x) = x + \lambda_s s$ applied uniformly across tokens. Foreground stabilization can be interpreted as restricting variation within the induced subspace: it reduces the effective degrees of freedom of foreground tokens, biasing them toward a consistent latent configuration, while background tokens continue to explore the full space. We emphasize that this provides an interpretation rather than a strict geometric constraint; the formal statement is as follows.

Let $s \in \mathbb{R}^d$ be a fixed style direction and define the affine submanifold
\begin{equation}
\mathcal{M}_s = \{ x = y + \lambda s \mid y \in \mathbb{R}^d, \lambda \in \mathbb{R} \}.
\end{equation}

\paragraph{Proposition 3 (Affine Invariance).}
If $x_0 \in \mathcal{M}_s$, then repeated application of style injection and diffusion updates approximately preserves the affine structure of $\mathcal{M}_s$, up to orthogonal deviations.

\paragraph{Geometric Argument.}
Style injection applies a translation along $s$. 
The diffusion drift can be decomposed into components parallel and orthogonal to $s$:
\begin{equation}
v_\theta(x) = v_\parallel(x) + v_\perp(x),
\end{equation}
where $v_\parallel(x) \in \mathrm{span}(s)$ and $v_\perp(x) \perp s$.

Under the assumption that the style-parallel component is approximately preserved by the pretrained diffusion drift, the affine structure induced by style injection remains stable throughout the generation process. Orthogonal components allow expressive variation while preserving the dominant style direction.

\subsubsection{Energy-Based Interpretation}
\label{subsec:proof_energy}

Define the foreground energy
\begin{equation}
\mathcal{E}_{\text{fg}}(x) 
= \sum_{k \in \mathcal{I}_{\text{fg}}} \lVert x^{(k)} - b \rVert^2.
\end{equation}

\paragraph{Proposition 4 (Energy Dissipation).}
Foreground stabilization induces monotonic dissipation of $\mathcal{E}_{\text{fg}}$.

\paragraph{Energy Argument.}
Substituting the relaxation update yields
\begin{equation}
\mathcal{E}_{\text{fg}}(x_{t-\Delta t}) 
= (1-\gamma(t))^2 \mathcal{E}_{\text{fg}}(x_t),
\end{equation}
which implies exponential decay of foreground energy and convergence toward a low-entropy equilibrium.

\paragraph{Temperature Interpretation.}
The stabilization of Eq.~\ref{eq:stab_relax} can further be read as a temperature modulation:
\begin{equation}
T_{\text{fg}}(t) = (1 - \gamma(t))\,T(t),
\end{equation}
with
\begin{equation}
\lim_{t \to 0} T_{\text{fg}}(t) = 0.
\end{equation}
This suggests that foreground regions converge toward a stable, low-variance state. We emphasize that this thermodynamic view provides intuition rather than a strict physical model.

\subsubsection{Unified Theoretical Statement}

\paragraph{Theorem 1 (Controlled Diffusion with Stable Foreground).}
The proposed state-space control framework induces 
(i) time-scale separation between foreground and background dynamics,
(ii) asymptotically stable foreground equilibria,
and (iii) invariant global style submanifolds,
while preserving expressive background generation.

\paragraph{Argument.}
The result follows directly from the time-scale separation property, Lyapunov stability of foreground states, and geometric invariance of style-induced affine submanifolds.
\hfill $\square$

\paragraph{Remark.}
Diffusion acts here as a numerical solver for a controlled dynamical system. 
The underlying theoretical structure is independent of the specific generative backbone.

\subsection{Experiment Input Prompts}
\label{app:prompts}

To evaluate our research under a wide range of stylistic preferences, we adopt a set of representative background design prompts spanning seven visual categories: \emph{geometric}, \emph{shapes}, \emph{textures}, \emph{colorful}, \emph{muted}, \emph{professional}, and \emph{real and natural objects}. The prompts used in our experiments are listed in Table~\ref{tab:input_prompts}. These prompts are drawn from an existing publicly available benchmark introduced in prior document-centric background generation work, ensuring reproducibility and consistency with prior experimental settings.

This setting reflects realistic document editing scenarios, where users express coarse stylistic intent while expecting the system to respect and align with the existing document content.

\begin{table*}[t]
\centering
\begin{tabular}{|l|p{0.5\linewidth}|}
\hline
\textbf{Category} & \textbf{Example user prompt} \\
\hline
Geometric & ``Add a background with a modern abstract design composed of layered geometric forms, featuring clean repetitions and harmonious symmetry for a structured visual effect.'' \\
\hline
Shapes & ``Add a background featuring a playful yet balanced composition of varied shapes, combining bold curves and soft angles to create natural depth.'' \\
\hline
Textures & ``Add a background with richly layered textures, where smooth and coarse surfaces interact to produce tactile depth and visual interest.'' \\
\hline
Colorful & ``Add a background with a vivid, lifelike scene filled with diverse colors under natural lighting, creating a vibrant and dynamic atmosphere.'' \\
\hline
Muted & ``Add a background with a softly lit, realistic setting using a desaturated color palette that conveys calmness and understated elegance.'' \\
\hline
Professional & ``Add a background with a refined and realistic design, emphasizing clean lines, minimal clutter, and subtle details for a polished appearance.'' \\
\hline
Real and natural objects & ``Add a background inspired by real-world elements, incorporating bright yet natural details to create an inviting and authentic visual setting.'' \\
\hline
\end{tabular}
\caption{Representative input prompts used in the experiments, adopted from an existing public prompt set used in prior document-centric background generation studies. Prompts provide high-level stylistic guidance, while the textual content of the input PDF supplies semantic grounding for background generation.}
\label{tab:input_prompts}
\end{table*}

\subsection{Supplementary Benchmarking Datasets}
\label{appendix:dataset_new}

To evaluate the robustness and generality of our proposed background generation framework, we construct a new set of benchmarking datasets consisting of academic-style documents and presentation slides. Unlike several existing document-centric background generation datasets, which predominantly focus on a limited range of social or policy-oriented topics, our benchmarks intentionally span diverse, content-neutral domains such as physics, engineering, media technology, design, and everyday systems. This design choice emphasizes layout preservation, text readability, and visual consistency, rather than domain-specific semantics.

Each dataset instance consists of exactly three pages (for documents) or three slides (for presentations). The three-page structure is deliberately chosen to reflect realistic educational materials, where content naturally progresses from introduction to elaboration and finally to synthesis or conclusion. This structure allows us to evaluate not only single-page background quality, but also multi-page coherence under semantic shifts.

\paragraph{Document corpus.}
The academic document corpus contains seven PDF documents, each written in a textbook- or university handout–style format. All documents use dense paragraph-based prose and are formatted in A4 size. Each document follows a consistent three-part thematic organization, with page-level semantic progression designed to stress-test background consistency across related but distinct content.

The seven document topics are:
\begin{itemize}
    \item \textbf{The History of Musical Instruments} \\
    P1. Ancient \& Traditional Instruments, 
    P2. Mechanical \& Acoustic Innovations, 
    P3. Digital \& Electronic Instruments

    \item \textbf{The Science of Cooking} \\
    P1: Heat, Chemistry, and Flavor, 
    P2: Cooking Techniques Around the World, 
    P3: Modern Gastronomy \& Food Technology
    
    \item \textbf{Architecture Through the Ages} \\
    P1. Classical \& Ancient Structures, 
    P2. Industrial \& Modern Architecture, 
    P3. Contemporary \& Sustainable Design
    
    \item \textbf{Exploring the Ocean Depths} \\
    P1. Ocean Zones and Geography, 
    P2. Marine Life Adaptations, 
    P3. Deep-Sea Exploration Technology
    
    \item \textbf{The Evolution of Transportation} \\
    P1. Early Human \& Animal Transport, 
    P2. Industrial Transportation Systems, 
    P3. Future Mobility Concepts
    
    \item \textbf{Understanding Weather Phenomena} \\
    P1. Atmospheric Basics, 
    P2. Extreme Weather Events, 
    P3. Forecasting and Visualization Tools
    
    \item \textbf{The Art of Typography and Design} \\
    P1. Origins of Written Symbols, 
    P2. Print and Graphic Design, 
    P3. Digital Typography and Interfaces
\end{itemize}

Each document page includes long-form textual paragraphs and at least one figure, placed in non-trivial spatial configurations (e.g., adjacent to text blocks or offset within columns). This layout diversity is intended to challenge background generation methods to preserve foreground legibility while maintaining visual harmony across pages.

\paragraph{Slide corpus.}
In parallel, we construct a set of seven academic-style slide decks, each formatted in 16:9 aspect ratio and consisting of exactly three slides. Each deck follows a strict narrative scaffold:
\emph{Slide 1 (Introduction)}, \emph{Slide 2 (Body)}, and \emph{Slide 3 (Conclusion)}.
This structure mirrors common lecture and presentation practices, where high-level concepts are introduced, expanded through technical details, and summarized through implications or future directions.

The seven slide topics are:
\begin{itemize}
    \item \textbf{How Cameras Work}
    \item \textbf{Inside a Modern Video Game}
    \item \textbf{The World of Maps and Navigation}
    \item \textbf{Robots in Everyday Life}
    \item \textbf{The Basics of Sound and Music}
    \item \textbf{Visual Effects in Movies}
    \item \textbf{How the Internet Delivers Information}
\end{itemize}

Each slide contains a title, a short explanatory paragraph, exactly four bullet points, and one or more images positioned below or beside the text. This combination of prose and bullet points reflects realistic instructional slides while introducing layout complexity through mixed text density and figure placement.

\paragraph{Content provenance and copyright compliance.}
All textual content and images in both corpora are generated using GPT-based text generation and multimodal image synthesis. No third-party copyrighted material, datasets, or templates are used. All documents and slides are free-form in layout and content, ensuring full compliance with copyright and intellectual property constraints.

\paragraph{Design rationale.}
The datasets are intentionally designed to be \emph{domain-agnostic} and \emph{layout-diverse}. We vary:
(i) textual density (long paragraphs vs.\ compact bullet lists),
(ii) semantic focus across pages or slides,
(iii) spatial placement of figures relative to text,
and (iv) visual tone across multi-page units.
These variations allow us to systematically evaluate whether background generation remains text-preserving, layout-aware, and stylistically consistent under realistic, non-template-constrained conditions.

\paragraph{Font selection and accessibility compliance.}
Font usage in the proposed datasets is deliberately designed to balance accessibility compliance, measurement rigor, and realistic document diversity. For presentation slides, we exclusively adopt \textbf{Atkinson Hyperlegible}, a font explicitly developed for accessibility and low-vision readability, in accordance with the principles of the United States Americans with Disabilities Act (ADA) and Section~508 accessibility guidelines.\footnote{\url{https://www.section508.gov/develop/fonts-typography/}} This choice ensures maximal visual clarity and eliminates font-induced variability in instructional materials.

For the PDF document corpus, we intentionally diversify font usage across three ADA-compliant font families: \textbf{Atkinson Hyperlegible}, \textbf{Verdana}, and \textbf{Helvetica}. This design reflects realistic academic document practices, where accessibility-oriented, screen-optimized, and industry-standard fonts coexist. At the same time, all selected fonts satisfy Section~508 recommendations for legibility, stroke clarity, and sans-serif typography, avoiding decorative or high-contrast serif styles that could introduce confounding visual artifacts.

\paragraph{Font-aware and reproducible readability evaluation.}
All WCAG contrast measurements are computed directly on rasterized page images at a fixed DPI using a consistent, programmatic procedure. Specifically, we extract text regions from embedded PDF text when available (otherwise from OCR bounding boxes), and estimate foreground--background contrast per region via pixel-based segmentation. This evaluation protocol is deterministic given the rendered pages and avoids any GPT-based judgment. While font choice can influence rasterization, limiting the font pool to Section~508-aligned sans-serif families and using a fixed rendering configuration reduces uncontrolled variation and strengthens fairness compared to benchmarks that do not explicitly constrain typography.

\paragraph{Intended evaluation use.}
These supplementary datasets are used to evaluate:
\emph{(a)} preservation of foreground content such as text and figures,
\emph{(b)} alignment between background visuals and page-level semantics,
\emph{(c)} multi-page visual coherence under topic progression, and
\emph{(d)} automated readability, contrast, and OCR-based metrics.
Together, they provide a controlled yet realistic testbed for analyzing background generation quality beyond single-page settings.

\subsection{Benchmark Dataset Scope and Design Rationale}
\label{appendix:dataset_rationale}

We clarify the scope and design rationale of our benchmarking datasets to avoid potential misinterpretation of their role and intended use. The datasets employed in this work are intentionally self-constructed and synthetic, not as a simplification of real-world conditions, but as a controlled evaluation environment aligned with the specific objective of foreground-preserving background generation.

\paragraph{Legal and reproducibility constraints.}
Constructing a publicly redistributable benchmark from real-world documents presents substantial legal and ethical challenges. The majority of academic papers, textbooks, magazines, web documents, and scanned PDFs are protected by copyright and cannot be redistributed or visually modified as part of an open benchmark. While U.S. federal government documents are largely exempt from copyright restrictions, they represent a narrow and domain-specific subset that does not reflect the diversity of educational and presentation-oriented materials targeted in this work. Large-scale crawling or scraping of documents further introduces unresolved licensing ambiguity. To ensure full copyright compliance, reproducibility, and long-term availability, we therefore adopt a self-constructed dataset in which all textual and visual content is synthetically generated and freely redistributable.

\paragraph{Task alignment versus uncontrolled real-world noise.}
The goal of this work is not document recovery, OCR robustness, or layout reconstruction from degraded inputs, but \emph{foreground-preserving background generation} for document editing. This task assumes that the foreground text and layout are known and must remain readable and unaltered while new background content is synthesized.

Highly cluttered or densely packed real-world documents, particularly scanned PDFs with minimal margins, overlapping artifacts, or severe degradation, leave little semantic or spatial capacity for meaningful background generation. Such conditions confound background interference with unrelated degradation factors and obscure whether failures arise from background synthesis itself or from upstream document quality. In contrast, documents with intentional margins, structured layouts, and moderate text density more accurately reflect realistic use cases such as academic slides, lecture handouts, posters, and presentation materials, where background editing is both meaningful and practically relevant.

\paragraph{Controlled complexity as an evaluation principle.}
Rather than maximizing visual noise, our benchmark emphasizes controlled complexity. By systematically varying layout structure, text density, font accessibility, figure placement, and page-level semantic progression, the dataset isolates the core challenges of background generation: preserving foreground legibility, respecting layout constraints, and maintaining stylistic coherence across pages.

Introducing uncontrolled artifacts from real-world scanning or degraded rendering would entangle these factors with orthogonal sources of noise, reducing interpretability of evaluation outcomes. Our design philosophy therefore prioritizes clarity of attribution over uncontrolled realism. We view this benchmark as complementary to real-world document collections, which are better suited for studying robustness to degradation but not for isolating background generation behavior.

\paragraph{Scope of evaluation.}
Accordingly, the proposed benchmark is not intended to exhaustively represent all real-world document conditions. Instead, it provides a legally compliant, task-aligned, and reproducible testbed for analyzing background generation quality under conditions where background synthesis is meaningful and foreground preservation is critical. We believe this controlled scope is essential for making reliable and interpretable claims about background generation methods.

\subsection{Implementation Details}
\label{appendix:implementation}
Our implementation integrates our model directly into the diffusion sampling process and generates each page through trajectory-level control rather than hard constraints. Structured pages are processed to provide auxiliary contexts such as representative foreground indicators and compact semantic summaries. Foreground text regions are preserved by initializing latents with a background-aware noise scheme and progressively suppressing updates inside text regions using a late-emphasized schedule $\alpha(t)=(1-t)^2$. Additional stabilization is applied in later timesteps by softly pulling high-confidence text-interior tokens toward a fixed backing latent, with stronger attraction in interior regions and weaker blending near boundaries. Global visual appearance is controlled via a lightweight style steering mechanism, in which style directions are pre-constructed offline by aggregating latent representations from a small set of descriptive prompts, and are added to the diffusion trajectory with late-stage gating and bounded magnitude. All hyperparameters are chosen to define a stable operating regime rather than to finely tune performance, and the method is insensitive to moderate variations of these values, as they primarily regulate the relative timing and strength of trajectory-level guidance. Our system follows a diffusion-based inference pipeline~\cite{deng2025emerging}, with all foreground preservation and style control realized through trajectory-level modifications.

\paragraph{Style Bank construction.}
The Style Bank contains one direction per style category ($K=7$, matching the seven categories in Table~\ref{tab:input_prompts}). Each direction is constructed offline as follows: for a given category, a small set of descriptive prompts (the category prompt of Table~\ref{tab:input_prompts} together with short paraphrases) is encoded through the diffusion backbone's conditioning pathway; the resulting latent representations are aggregated by averaging and the mean is normalized to unit norm, yielding the direction $s_i \in \mathbb{R}^{d}$. Directions are computed once, cached, and remain fixed during inference; no per-page or per-document recomputation is performed. For user-specified prompts beyond the predefined categories, a direction is constructed on-the-fly by the same encode--average--normalize procedure.

\subsection{Conceptual Positioning and Comparison with Prior Work}
\label{appendix:positioning}

This section collects the positioning tables referenced in the main paper. Table~\ref{tab:method_comparison} positions existing method categories against the five simultaneous requirements of document-centric background editing, and Table~\ref{tab:acd_vs_prior} details the per-axis comparison with the most closely related work~\cite{kang2025text}.

\begin{table*}[t]
\centering
\small
\caption{\textbf{Conceptual positioning against existing method categories.} Our document-centric background editing setting imposes five requirements simultaneously. ACD is the only formulation that satisfies all of them, which motivates our restriction of direct empirical comparison to BAGEL~\cite{deng2025emerging} and GPT-5~\cite{openai2025introducinggpt5}, the only existing methods that operate directly on instantiated multi-page documents without re-rendering layout or text. \cmark/\xmark\ denote whether a category satisfies a requirement by construction; ``---'' indicates inapplicable.}
\label{tab:method_comparison}
\resizebox{\textwidth}{!}{%
\begin{tabular}{lccccc}
\toprule
\textbf{Method category (representatives)}
& \textbf{Operates on}
& \textbf{Foreground}
& \textbf{Multi-page}
& \textbf{Dense text}
& \textbf{Training-} \\
&  \textbf{existing pages}
&  \textbf{control mechanism}
&  \textbf{consistency}
&  \textbf{preservation}
&  \textbf{free} \\
\midrule
Layout / poster generation~\cite{chen2025posta,zhang2025creatiposter,peng2025bizgen,zhang2025creatidesign,wu2025hybrid}
& \xmark\ (regenerates layout) & via re-layout & \xmark & \xmark & --- \\
Mask-guided inpainting~\cite{ju2024brushnet,lugmayr2022repaint,avrahami2022blended,couairon2022diffedit}
& partial & binary spatial mask & \xmark & \xmark & \cmark \\
Text-aware diffusion~\cite{chen2023textdiffuser,chen2024textdiffuser,10.1145/3721250.3743023}
& partial & assumes blank / rendered text & \xmark & \xmark & --- \\
Layered diffusion~\cite{huang2024layerdiff,li2023layerdiffusion,zhang2024transparent}
& \xmark\ (requires clean layers) & layer separation & \xmark & \xmark & \xmark \\
Attention control~\cite{hertz2022prompt,cao2023masactrl,chefer2023attend}
& partial & attention re-weighting & \xmark & \xmark & \cmark \\
General-purpose editing~\cite{deng2025emerging,openai2025introducinggpt5}
& \cmark & implicit / none & \xmark & partial & inference-only \\
\midrule
\textbf{Ours (ACD)}
& \cmark & soft, trajectory-level & \cmark & \cmark & \cmark \\
\bottomrule
\end{tabular}}
\end{table*}

\begin{table}[t]
\centering
\small
\caption{\textbf{Per-axis comparison with the most closely related work~\cite{kang2025text}.} Prior work realizes each requirement through an external intervention applied outside the sampling process, whereas ACD subsumes all three into the sampling dynamics themselves.}
\label{tab:acd_vs_prior}
\resizebox{0.85\linewidth}{!}{%
\begin{tabular}{lll}
\toprule
\textbf{Axis} & \textbf{Kang et al.~\cite{kang2025text}} & \textbf{Ours (ACD)} \\
\midrule
Foreground preservation & fixed attenuation mask & time-dependent trajectory gating (Eq.~\ref{eq:fg_gating}) \\
Readability & post-hoc semi-transparent backing shapes & stabilized canvas alone (WCAG 98.12\%) \\
Cross-page consistency & per-page LLM-generated instructions & persistent latent style direction \\
Per-page inference dependency & one LLM call per page & none \\
\bottomrule
\end{tabular}}
\end{table}

\subsection{Detailed Qualitative Results}
\label{appendix:qualitative}

\begin{figure*}[t]
    \centering
    \includegraphics[width=\linewidth]{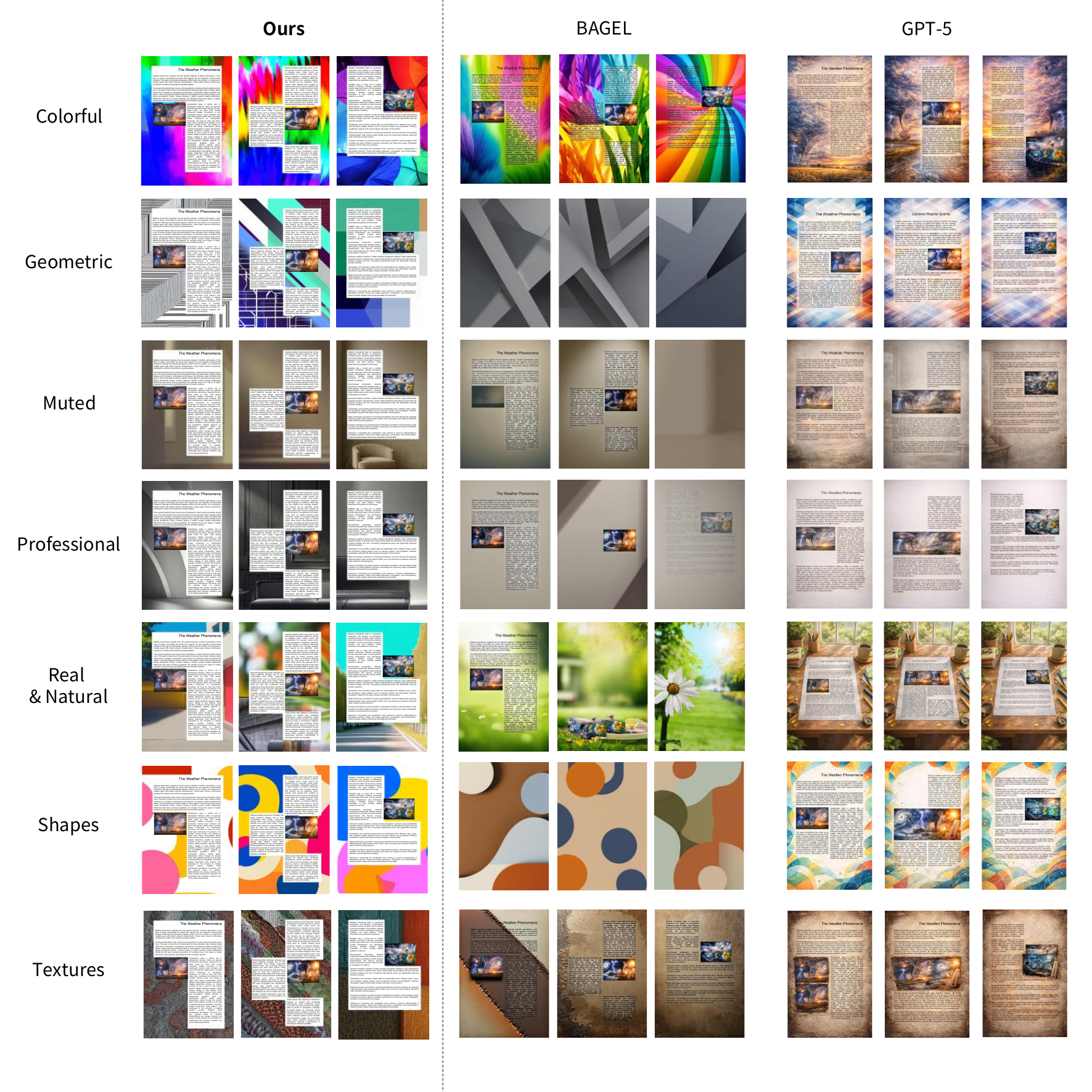}
    \caption{\textbf{Visual comparison on academic-format PDFs (A4).} Each row corresponds to a different style setting (\textit{Colorful, Geometric, Muted, Professional, Real \& Natural, Shapes, Textures}).}
    \label{fig:results_pdfs}
\end{figure*}

\begin{figure*}[t]
    \centering
    \includegraphics[width=\linewidth]{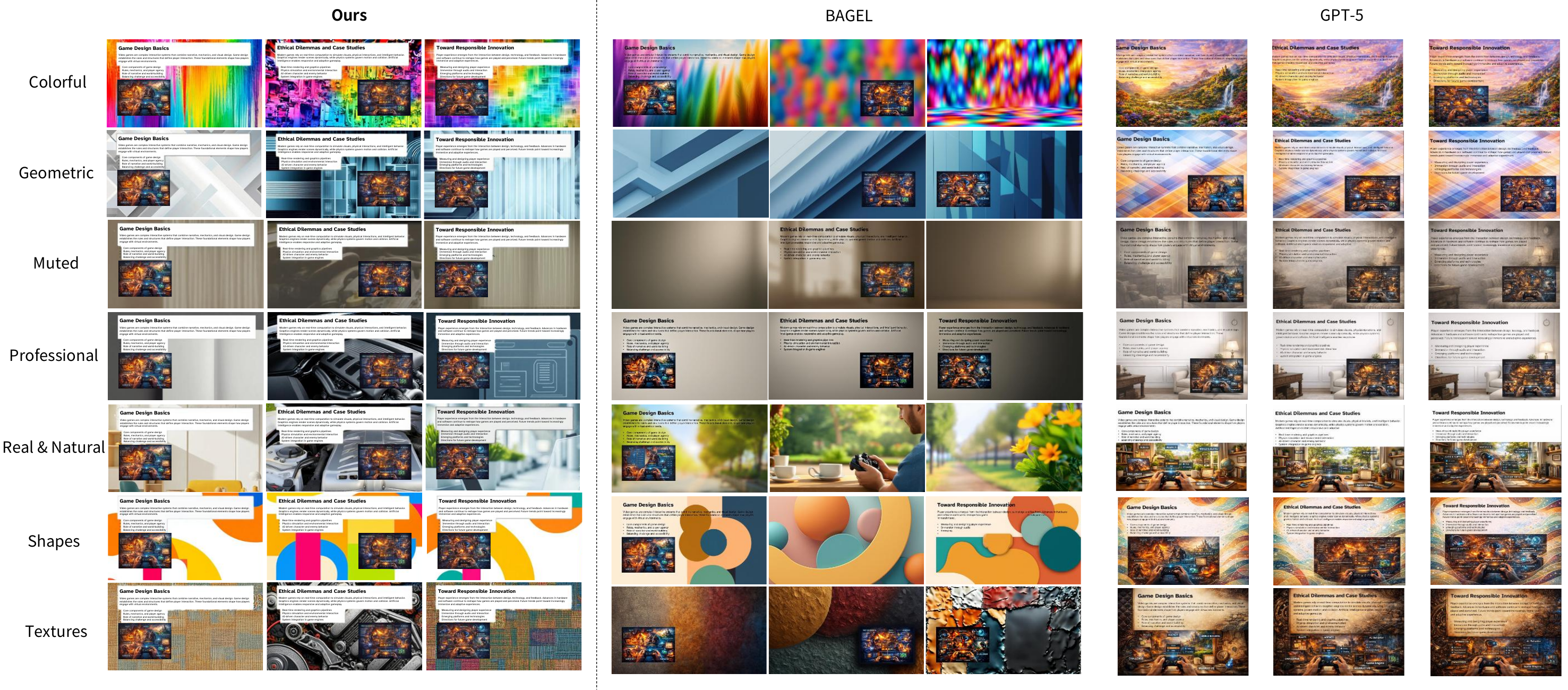}
    \caption{\textbf{Visual comparison on academic presentation slides (16:9).} Rows indicate style conditions (\textit{Colorful, Geometric, Muted, Professional, Real \& Natural, Shapes, Textures}).}
    \label{fig:results_slides}
\end{figure*}

\begin{figure*}[t]
    \centering
    \includegraphics[width=\linewidth]{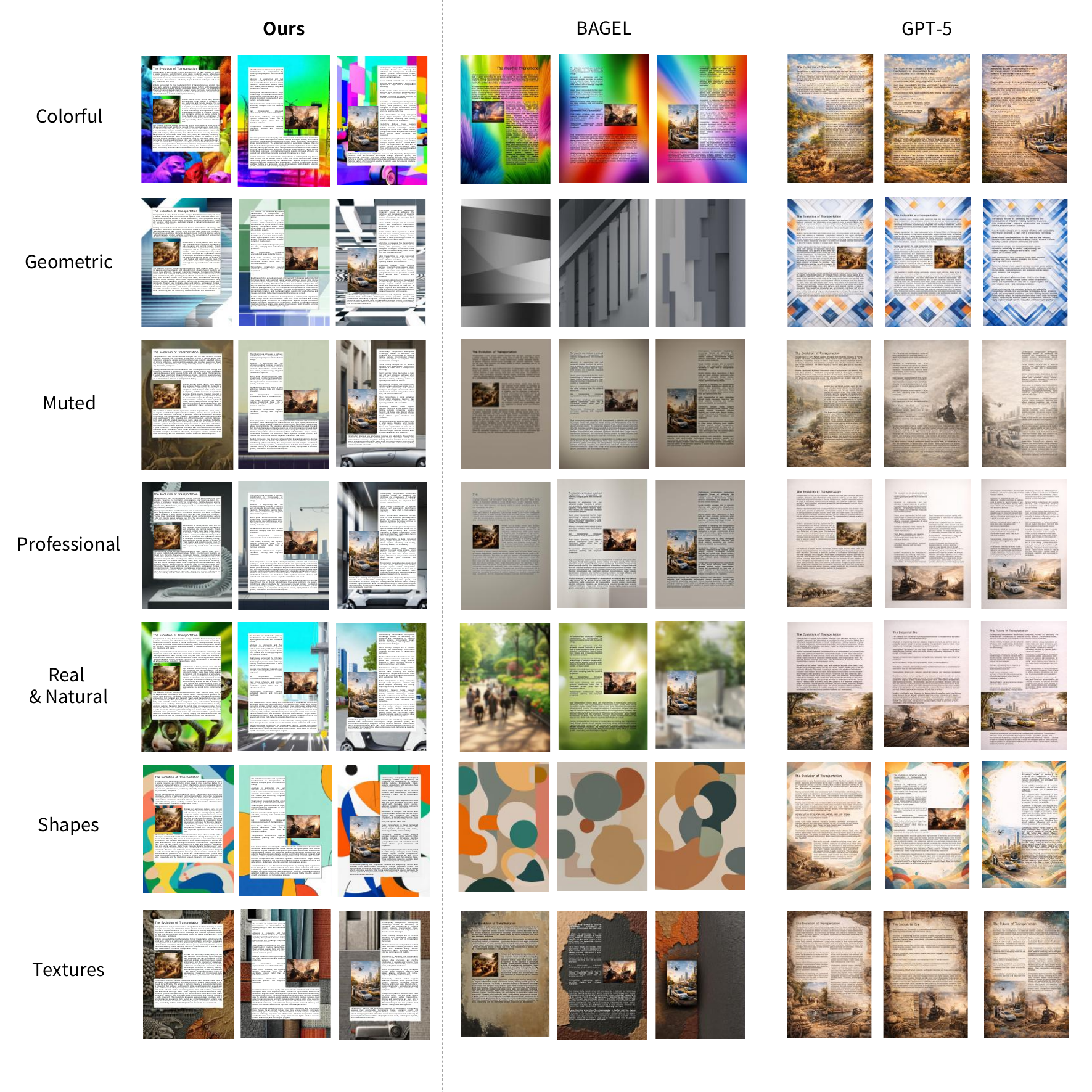}
    \caption{\textbf{Additional qualitative examples on academic PDFs (A4).} 
    Style conditions are arranged by rows (\textit{Colorful, Geometric, Muted, Professional, Real \& Natural, Shapes, Textures}).}
    \label{fig:results_pdfs_supp}
\end{figure*}

\begin{figure*}[t]
    \centering
    \includegraphics[width=\linewidth]{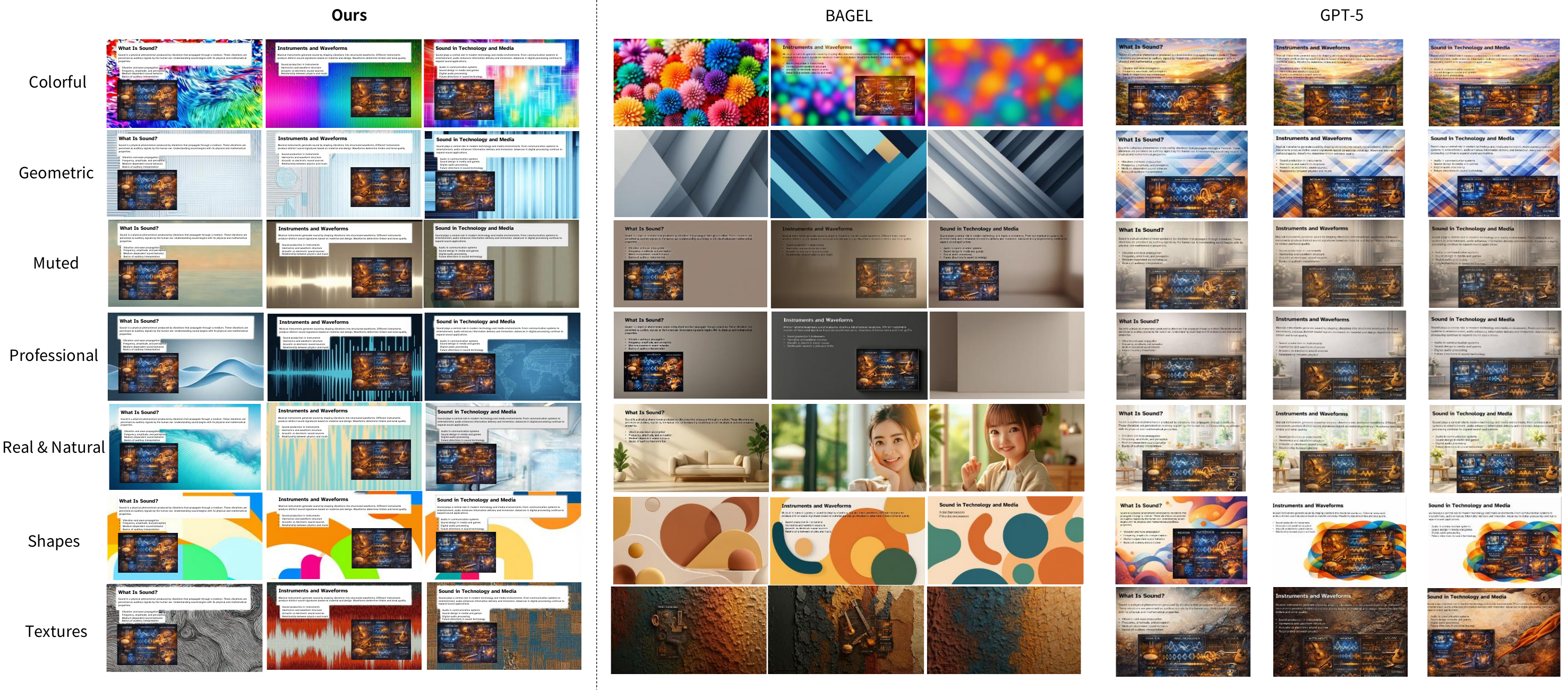}
    \caption{\textbf{Additional qualitative examples on academic slides (16:9).} 
    Rows correspond to the style settings (\textit{Colorful, Geometric, Muted, Professional, Real \& Natural, Shapes, Textures}).}
    \label{fig:results_slides_supp}
\end{figure*}

We present detailed qualitative comparisons between our framework and two document editing baselines, BAGEL~\cite{deng2025emerging} and GPT-5~\cite{openai2025introducinggpt5}, following an evaluation protocol adopted in prior document-centric background generation work. All methods are applied to identical input documents without manual post-processing.

A defining characteristic of our problem setting is that the input is an \emph{existing document page}, either a PDF or a slide, with fixed text regions and layout. The goal is to generate backgrounds that reflect the page content while strictly preserving foreground text and structural layout. This distinguishes our task as a document-centric \emph{editing} problem rather than generation from scratch.

Figures~\ref{fig:results_pdfs} and~\ref{fig:results_slides} present qualitative results on academic-style PDFs and slides across seven stylistic conditions: Colorful, Geometric, Muted, Professional, Real \& Natural, Shapes, and Textures. BAGEL often produces visually strong backgrounds with rich textures or patterns; however, these backgrounds frequently intrude into text regions or reduce local contrast, negatively affecting readability. GPT-5 generates aesthetically pleasing backgrounds with a coherent global appearance; however, its page-wise independent generation paradigm does not enforce foreground- or layout-preserving constraints. As a result, it often alters the original document structure, leading to inconsistencies in background motifs across pages, unintended modifications of text placement or figures, and occasional interference with foreground content. In some cases, GPT-5 also introduces additional visual or textual elements not present in the input, making it unsuitable for document-centric background editing tasks that require strict preservation of existing layout and content.

In contrast, our method consistently avoids foreground regions and maintains stable visual themes across pages. Backgrounds generated by our approach reflect the semantic content of each page while remaining visually subdued around text regions, resulting in improved readability and layout fidelity. These differences are particularly evident in texture-heavy and shape-based styles, where uncontrolled background variation tends to degrade text clarity in baseline methods.

Figures~\ref{fig:results_pdfs_supp} and~\ref{fig:results_slides_supp} provide additional qualitative examples, further confirming that our framework robustly balances stylistic expressiveness, semantic relevance, and readability preservation across diverse document formats and content domains.

\paragraph{Scope of comparison.} 
While recent research, including POSTA~\cite{chen2025posta} and CreatiPoster~\cite{zhang2025creatiposter}, as well as other design-oriented pipelines~\cite{peng2025bizgen, zhang2025creatidesign, wu2025hybrid}, has explored modular approaches for poster and graphic design generation, these methods are designed for flexible-layout, sparse-text settings and typically assume user-defined structural inputs. In contrast, our problem focuses on document-centric editing, where the layout and foreground content are fixed and must be strictly preserved. Applying such methods would require re-generating layout and textual content, which violates the core assumptions of our task and leads to an unfair comparison.

Similarly, methods based on explicit masking or foreground removal (e.g., TextDiffuser-style pipelines or mask-guided inpainting) assume either controllable text rendering or removable foreground regions, and therefore do not satisfy the constraint of preserving dense, pre-existing document content. \textbf{Direct comparison with such methods would require redefining the editing task itself, relaxing the foreground-preservation constraint that motivates our formulation, and therefore lies outside the scope of this work.} To ensure a fair and meaningful evaluation, we therefore restrict our comparison to methods that can directly operate on existing document pages without modifying layout or text content. Under these criteria, BAGEL and GPT-5 remain appropriate baselines for document-centric background generation.

\subsection{Naive Independent-Generation-plus-Overlay Baseline}
\label{appendix:naive_overlay}

\begin{figure*}[t]
    \centering
    \includegraphics[width=0.8\linewidth]{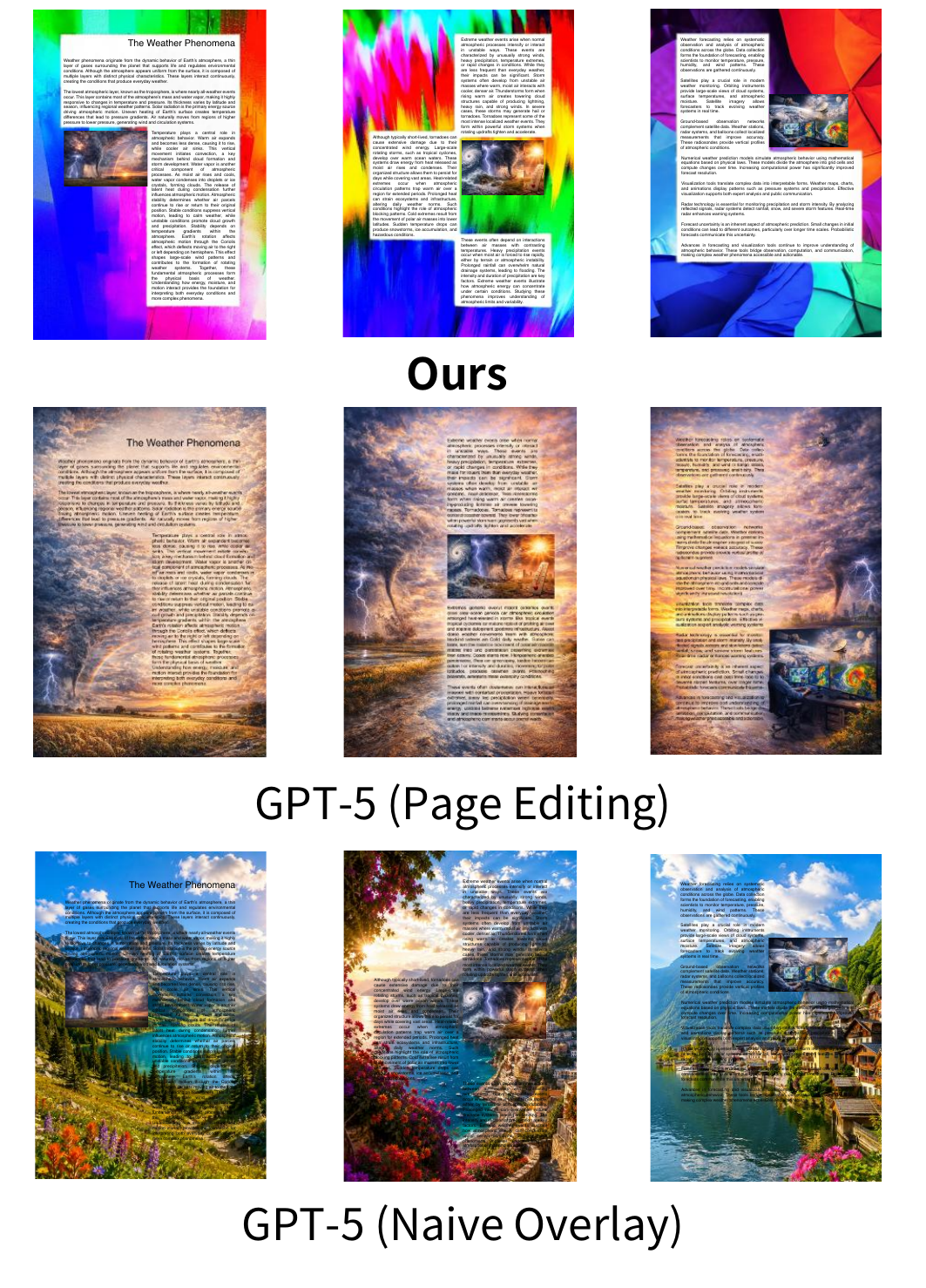}
    \caption{\textbf{Qualitative comparison with the naive independent-generation-plus-overlay baseline} on the Colorful style (same document, pages 1--3). \textbf{Top (Ours):} neutral regions beneath text are the stabilized canvas itself (Sec.~\ref{sec:stabilization}), not composited backing shapes; the palette is consistent across pages. \textbf{Middle (GPT-5, page editing):} independently edited pages diverge entirely in appearance (cf.\ CLIP MP 0.3113, Table~\ref{tab:quantitative_results}). \textbf{Bottom (GPT-5, Naive Overlay):} backgrounds intrude into text regions (the same failure mode as \emph{w/o SSC}) and pages drift stylistically despite the identical fixed prompt (P2 diverges in scene type, illumination, and palette).}
    \label{fig:naive_overlay}
\end{figure*}

This section details the \emph{GPT-5 (Naive Overlay)} baseline reported in Table~\ref{tab:quantitative_results}. The baseline instantiates the independent-generation-plus-overlay pipeline discussed in Sec.~4.2 in its most naive form: for each style category, background canvases are generated by GPT-5 from the identical fixed style prompt (Table~\ref{tab:input_prompts}), independently sampled per page in separate generation calls, at the target aspect ratio (A4 for documents, 16:9 for slides), with no access to layout geometry, document summaries, or the Style Bank. The original foreground is then composited onto each canvas via the same vector-rendering step used by our pipeline, and the composite pages are evaluated on the full benchmark (14 documents $\times$ 7 styles) under the identical protocol of Sec.~\ref{sec:quantitative}.

Figure~\ref{fig:naive_overlay} illustrates the two characteristic failure modes that the quantitative results capture. First, without layout awareness, salient background structures are placed beneath text regions, collapsing measured readability (WCAG 56.87\%, OCR 0.3077); even though the foreground text itself is preserved by construction, the composite page is no longer usable as a document. Second, despite sharing an identical prompt, independently sampled canvases drift in scene composition, illumination, and palette, and this drift is reflected in both CLIP MP Consistency (0.6085, below BAGEL's 0.6317) and LLM Voting (3.9943). These results confirm that the overlay structure alone does not account for our method's consistency and readability: layout-aware canvas stabilization and persistent style directions are both required.

\paragraph{Decomposition of the consistency gap.}
The quantitative results decompose the gap between our method and page-editing baselines into two parts. First, the background-only-canvas structure alone recovers much of the consistency over full-page editing (CLIP MP $0.3113 \rightarrow 0.6085$), confirming that part of the gain stems from generating canvases rather than editing pages. Second, a substantial gap to Ours nonetheless remains ($0.6085$ vs.\ $0.6785$, which is $5.6\times$ the \emph{w/o Style Bank} delta, together with LLM Voting $3.9943$ vs.\ $4.2992$), with naive consistency even falling below BAGEL ($0.6317$); this residual gap is closed by the proposed controls. Moreover, the readability collapse (WCAG $56.87\%$, OCR $0.3077$) mirrors the failure mode of \emph{w/o SSC}: although the foreground text is preserved by construction through vector rendering, the measured readability of the composite page is destroyed by background interference beneath text regions. For isolating the style directions specifically, the \emph{w/o Style Bank} ablation remains the cleaner control, as the naive baseline removes layout awareness, summaries, and style directions at once.

\subsection{Detailed Quantitative Analysis}
\label{appendix:quantitative}

This section provides a detailed analysis of the quantitative results reported in the main paper. We follow the same evaluation protocols and metric definitions adopted in prior document-centric background generation work to ensure fair comparison and reproducibility.

\paragraph{Design Quality.}
Design quality is evaluated along four dimensions: Layout, Color, Graphic Style, and Compliance, using an LLM-based evaluator with scores ranging from 1 to 5. Layout measures the structural balance between text and background elements, Color assesses palette harmony and suitability for document content, and Graphic Style captures stylistic coherence of visual elements. Compliance evaluates how well the generated background adheres to the input prompt while respecting document constraints.

Our research achieves the highest Layout score (4.24), indicating that trajectory-level guidance implicitly stabilizes background placement without explicit spatial constraints. While GPT-5 attains slightly higher scores in Color and Graphic Style, our method substantially outperforms both baselines in Compliance (4.74), reflecting stronger alignment between semantic intent and generated backgrounds. This suggests that our framework better preserves prompt adherence while avoiding foreground interference.

\paragraph{Readability.}
Readability is assessed using WCAG contrast coverage and OCR accuracy, which jointly capture perceptual accessibility and structural interference. WCAG coverage measures the percentage of text regions satisfying the WCAG~2.2 AA contrast threshold of 4.5:1, while OCR accuracy is computed at the character level using Tesseract.

Our method achieves 98.12\% WCAG compliance and the highest OCR accuracy (0.779), outperforming GPT-5 and BAGEL by a substantial margin. These results indicate that shaping diffusion trajectories to avoid foreground regions is more effective than post-hoc contrast adjustment or opacity tuning, as it reduces background-induced artifacts that degrade both contrast and text structure.

\paragraph{Multi-page Consistency.}
To evaluate cross-page coherence, we adopt two complementary measures: CLIP-based multi-page (MP) consistency and LLM-based voting. CLIP MP consistency computes cosine similarity between background embeddings of adjacent pages, capturing low-level visual coherence, while LLM voting assigns a score from 1 to 5 based on perceived thematic continuity across the document.

Our research achieves the highest CLIP MP consistency score (0.6785), significantly outperforming GPT-5, which exhibits poor consistency due to page-wise independent generation. Our method also attains the highest LLM voting score (4.2992), confirming that trajectory-level constraints preserve both structural similarity and high-level stylistic themes across pages.

Overall, these results validate that our framework provides a principled mechanism for document-centric background generation, jointly optimizing design quality, readability, and multi-page coherence without requiring hard masking or corrective post-processing.

\subsection{Detailed Ablation Study}
\label{appendix:ablation}
To disentangle the contributions of each module in our framework, we conduct ablation experiments under the same evaluation protocol and style conditions used throughout the main paper, while keeping all other settings fixed. Specifically, we remove (i) the \emph{Style Bank} and (ii) \emph{Diffusion State-Space Control (SSC)}. Quantitative results are summarized in Table~\ref{tab:quantitative_results}, and representative qualitative examples are shown in Fig.~\ref{fig:ablation}.

\paragraph{Baseline (Ours).}
Our full model achieves strong overall quality while explicitly balancing style expression and foreground preservation. As reported in Table~\ref{tab:quantitative_results}, \textbf{Ours} attains the best guideline compliance and accessibility scores (Compliance $4.74$, WCAG $98.12\%$, OCR $0.779$) together with the highest multi-page consistency (CLIP MP Consistency $0.6785$) and the strongest prompt alignment (CLIP Prompt Score $0.3144$). In Fig.~\ref{fig:ablation}, this manifests as backgrounds that remain visually expressive in non-text regions while keeping text regions readable and stable across pages.

\paragraph{Removing the Style Bank (Ours w/o Style Bank).}
The Style Bank provides a persistent style direction shared across pages, which is critical for suppressing cross-page stylistic drift. When we remove this component, we observe that readability-related metrics remain nearly unchanged, but style coherence degrades. Quantitatively, WCAG and OCR remain effectively preserved (WCAG $98.08\%$ vs.\ $98.12\%$, OCR $0.7769$ vs.\ $0.779$), and compliance stays high ($4.70$ vs.\ $4.74$), confirming that foreground preservation is primarily enforced by SSC rather than the Style Bank. In contrast, multi-page consistency drops (CLIP MP Consistency $0.6661$ vs.\ $0.6785$), accompanied by small decreases in appearance-related ratings (e.g., Color $3.9492$ vs.\ $4.07$, Graphic Style $4.055$ vs.\ $4.14$) and LLM Voting ($4.2335$ vs.\ $4.2992$). These trends align with Fig.~\ref{fig:ablation}, where the page-wise background appearance varies more noticeably without a persistent style anchor, even though foreground readability is largely maintained. Overall, this ablation isolates the Style Bank’s role as a \emph{cross-page style stabilizer} rather than a readability mechanism.

\paragraph{Removing SSC (Ours w/o SSC).}
SSC is responsible for controlling diffusion trajectories to preserve foreground readability. Disabling SSC causes a drastic collapse in accessibility and text preservation, consistent across both quantitative and qualitative evidence. As shown in Table~\ref{tab:quantitative_results}, removing SSC results in a sharp decline in WCAG coverage ($54.20\%$ vs.\ $98.12\%$) and OCR accuracy ($0.333$ vs.\ $0.779$), and also reduces compliance ($4.2785$ vs.\ $4.74$). Layout and overall human/LLM preferences also drop substantially (Layout $3.6885$ vs.\ $4.24$, LLM Voting $3.8764$ vs.\ $4.2992$), reflecting that the generated pages become less usable as documents. In Fig.~\ref{fig:ablation}, this failure mode is visually evident: background textures and colors intrude into foreground text regions, corrupting contrast and readability. Notably, CLIP MP Consistency remains relatively close to the full model ($0.6667$ vs.\ $0.6785$), which is expected since the Style Bank still provides a shared stylistic signal; however, stylistic consistency alone is insufficient when foreground content is not stabilized.

\paragraph{Takeaway.}
Together, these ablations highlight complementary roles: the \textbf{Style Bank} primarily improves \emph{cross-page stylistic consistency}, while \textbf{SSC} is essential for \emph{foreground preservation and accessibility}. The full model combines both, yielding the strongest overall performance and preventing the key failure modes observed in Fig.~\ref{fig:ablation} and Table~\ref{tab:quantitative_results}.

\subsection{Detailed User Study}
\label{appendix:user_study}

\paragraph{Participants.}
The study involved 30 participants. Participants were adults (18 years or older) recruited through online channels with open access, representing a general audience rather than a domain-specific expert group.

\paragraph{Survey Setup.}
Each participant completed 14 evaluation tasks, consisting of seven PDF pages and seven slides. For each task, participants were presented with the original document alongside three generated versions corresponding to BAGEL, GPT-5, and \emph{Ours}. The generated outputs were anonymized and labeled as Document A, B, and C, and \emph{the assignment and ordering of these labels were randomized for every task} to mitigate ordering and model-identity bias.

To support informed evaluation without requiring participants to read lengthy document text, two pieces of contextual information were provided:
\begin{itemize}
    \item \textbf{Input Prompt}, indicating the intended background style (e.g., \textit{Colorful}, \textit{Shapes}, \textit{Textures});
    \item \textbf{Content Summary}, a short semantic description of the document content.
\end{itemize}

Participants evaluated each generated output along four design dimensions (\emph{Layout preservation}, \emph{Color harmony}, \emph{Graphic style consistency}, and \emph{Prompt compliance}) using a 5-point Likert scale (1: strongly disagree, 5: strongly agree). No free-form comments were collected.

\paragraph{Evaluation Conditions.}
The 14 tasks covered seven background style categories used throughout the paper: \textit{Colorful}, \textit{Geometric}, \textit{Muted}, \textit{Professional}, \textit{Real \& Natural}, \textit{Shapes}, and \textit{Textures}. Each style was evaluated once on a PDF page and once on a slide.

\paragraph{Quantitative Results.}
Figure~\ref{fig:user_study} summarizes the quantitative results. Across all four dimensions, our method achieved the highest mean scores, ranging from \textbf{4.67 to 4.93}. In contrast, BAGEL and GPT-5 obtained substantially lower scores, typically below 2.0 on average, indicating frequent disruption of layout or reduced readability.

For the overall preference selection, participants chose a single favored result per task. Aggregated over all 14 tasks and all participants, \textbf{86.9\%} of preference votes were assigned to \emph{Ours}, while 13.1\% favored GPT-5. BAGEL received zero overall-preference votes; it was included in the aggregation rather than excluded, which is why it appears at 0.0\% in the overall-preference breakdown. This result highlights a strong and consistent user preference for our method in document-centric background editing scenarios.

\paragraph{Protocol and Ethics.}
The study required approximately 15--20 minutes per participant. No personally identifiable information was collected, and all responses were recorded anonymously. Participation was voluntary, and participants were free to withdraw at any time. The study protocol was reviewed by an Institutional Review Board at University of Maryland (Kuali-IRB Number 0294-1) and determined to be exempt from the requirements of 45 CFR 46 under Exempt Categories \textbf{45 CFR 46.104(d)(2)(i)} and \textbf{45 CFR 46.104(d)(2)(ii)}, as it involved anonymous survey procedures and did not include the collection of personally identifiable or sensitive information. The overall study design, evaluation dimensions, and presentation format are aligned with prior document-centric background generation~\cite{kang2025text} user studies, enabling direct comparability while using independently generated content and results.

\subsection{Computation Time and Memory Consumption}
\label{appendix:compute}

\begin{table}[htb!]
    \small
    \centering
    \resizebox{0.95\textwidth}{!}{
    \begin{tabular}{l|c|c}
    \toprule
    \textbf{Method (3-page document)} 
    & \textbf{Peak GPU memory (GB)} 
    & \textbf{Wall-clock time (sec)} \\
    \midrule
    BAGEL
    & 31.3461 GB / 40.000 GB
    & 246.00 sec \\
    \midrule
    GPT-5 (hosted API)
    & N/A
    & 324.00 sec \\
    \midrule
    Ours w/o SSC
    & 29.1953 GB / 40.000 GB
    & 243.00 sec \\
    \midrule
    Ours w/o Style Bank
    & 29.1953 GB / 40.000 GB
    & 243.00 sec \\
    \midrule
    \textbf{Ours (full ACD)}
    & \textbf{29.1953 GB / 40.000 GB}
    & \textbf{243.00 sec} \\
    \bottomrule
    \end{tabular}
    }
    \caption{\textbf{Deployment cost on a 3-page document.} All measurements are obtained on a single NVIDIA A100 GPU (40~GB) under the same hardware configuration. GPT-5 is invoked via its hosted API; the underlying GPU memory of the service is not externally observable and is reported as unavailable (N/A).}
    \label{tab:compute-module}
\end{table}

This appendix quantifies the deployment cost of ACD and reports reference numbers for comparable document editing systems. All measurements are obtained on a single NVIDIA A100 GPU (40~GB) and span the full pipeline from input parsing to final composition of a 3-page document.

\textbf{Inference cost of ACD.}
The complete ACD pipeline, comprising style direction injection, foreground-gated velocity modulation, and latent stabilization, produces background content for a 3-page document in 243 seconds, averaging 81 seconds per page. Peak GPU memory reaches 29.20~GB, well within the capacity of a single 40~GB device. The cost is determined almost entirely by the iterative sampling steps; the auxiliary control terms participate directly in those same steps rather than as separate stages.

\textbf{Cost of individual components.}
To attribute runtime to each auxiliary control term, we measure two ablated variants: ACD without State-Space Control (Style Bank only) and ACD without the Style Bank (SSC only). Both variants exhibit the same 243-second runtime and 29.20~GB peak memory as the full configuration. This is consistent with the algebraic form of the controls themselves: SSC's gating-plus-relaxation update and the Style Bank's single additive perturbation are realized entirely as in-place element-wise operations on the latent tensor, requiring no additional evaluations of the diffusion network. From a profiling perspective, enabling or disabling these terms does not alter the per-step compute graph in any measurable way.

\textbf{Baseline runtimes for context.}
BAGEL, a representative diffusion-based document editing baseline, completes the same 3-page task in 246 seconds and consumes 31.3461~GB of peak memory under identical hardware. GPT-5 is invoked through its hosted API, with client-side end-to-end latency reaching 324 seconds; the service does not externally surface its internal memory footprint, which we therefore report as unavailable. GPT-5's longer latency partly reflects its operating regime as a general-purpose multimodal model rather than as a constrained background-only editor, and is reported here as a deployment reference rather than as a method-level comparison.

\textbf{Implications.}
Two practical observations follow. First, ACD requires no post-hoc readability refinement stage or external LLM pass for multi-page coherence, because every control signal participates directly in the sampling dynamics. Second, the deployment cost of ACD scales primarily with the choice of diffusion sampler, leaving the framework readily portable to faster sampling schedules or backbones as they become available.

\newpage
\begin{figure}[t]
    \centering
    \includegraphics[width=\linewidth]{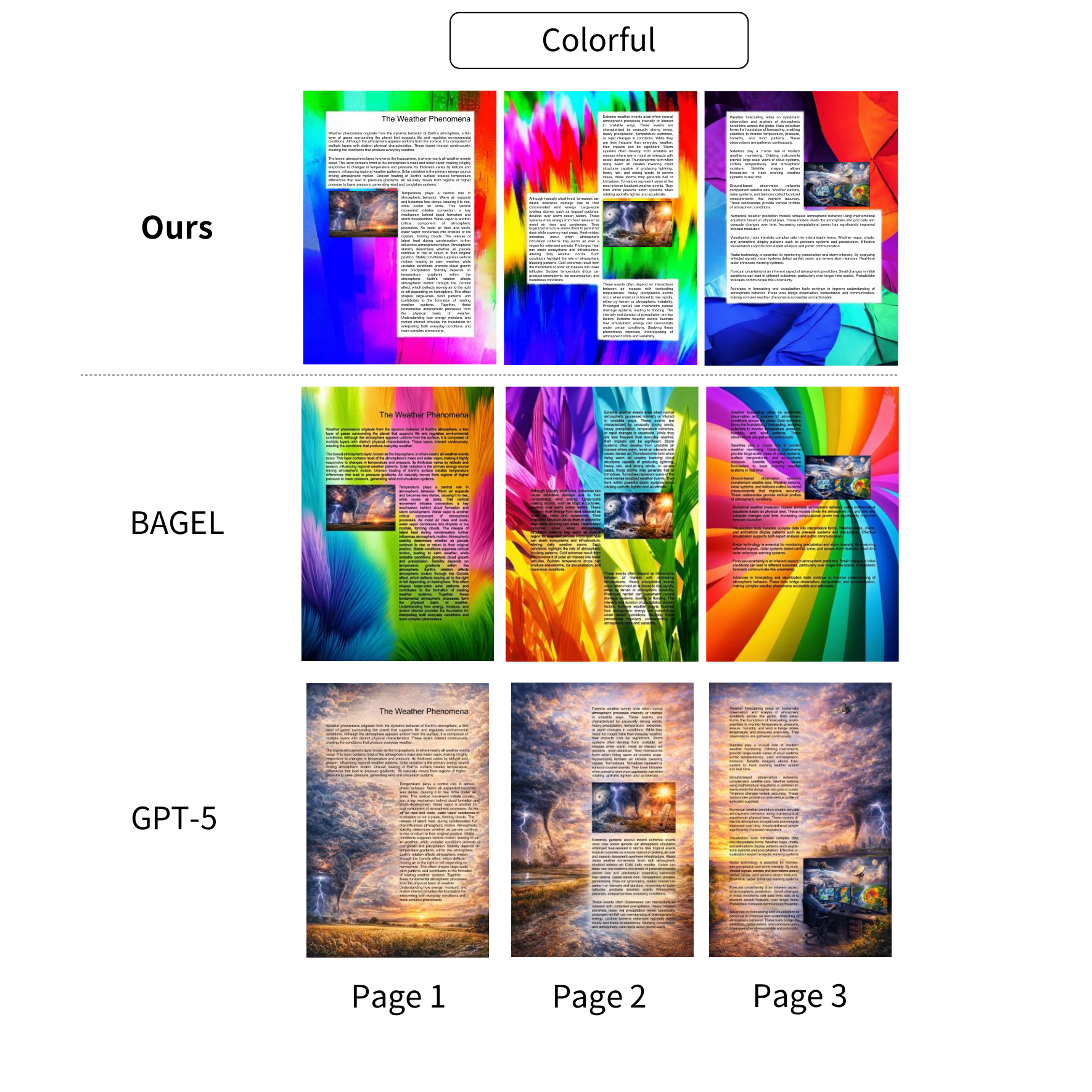}
    \caption{Qualitative comparison of background generation results for \emph{Colorful}-style PDFs.}
\end{figure}

\begin{figure}[t]
    \centering
    \includegraphics[width=\linewidth]{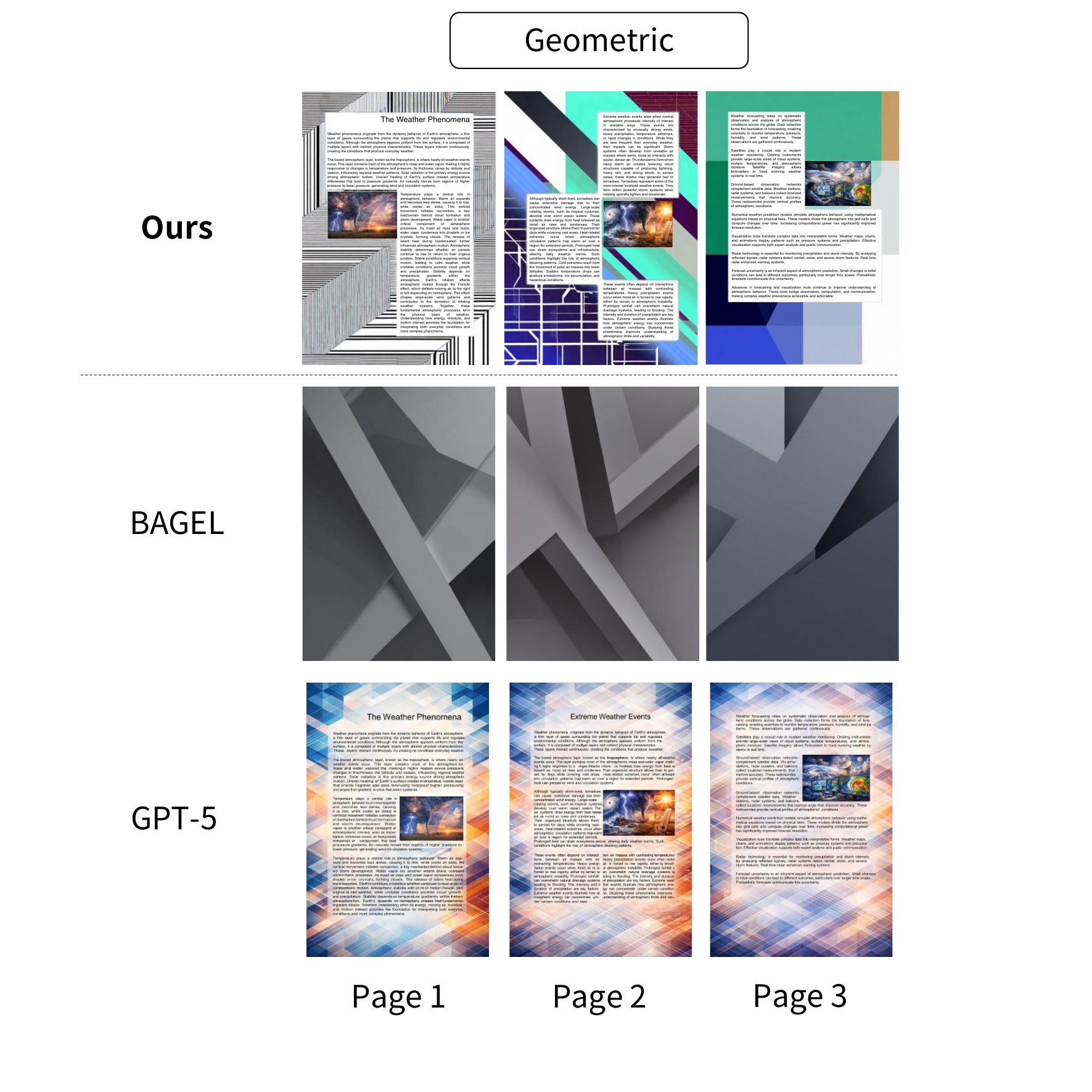}
    \caption{Qualitative comparison of background generation results for \emph{Geometric}-style PDFs.}
\end{figure}

\newpage
\begin{figure}[t]
    \centering
    \includegraphics[width=\linewidth]{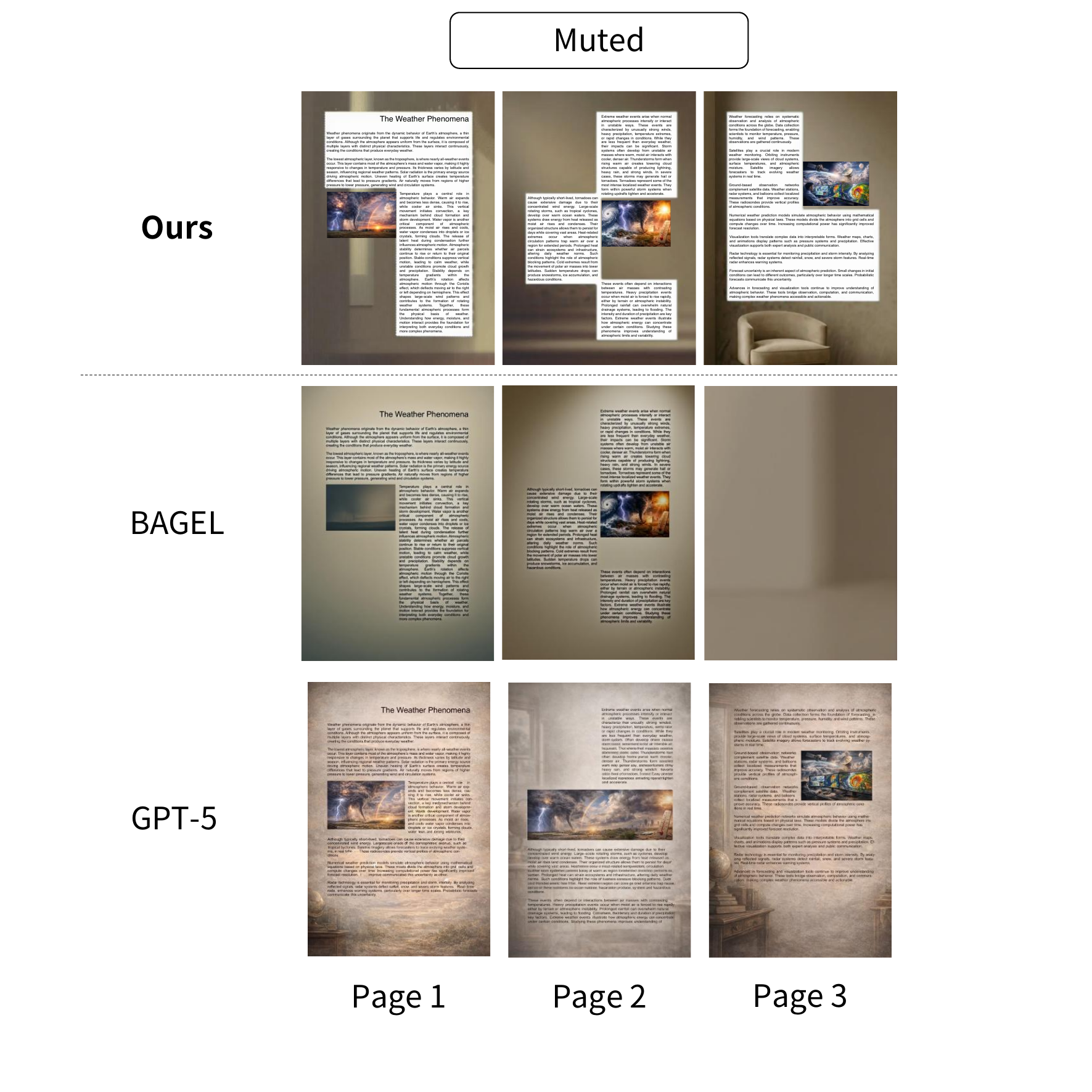}
    \caption{Qualitative comparison of background generation results for \emph{Muted}-style PDFs.}
\end{figure}

\newpage
\begin{figure}[t]
    \centering
    \includegraphics[width=\linewidth]{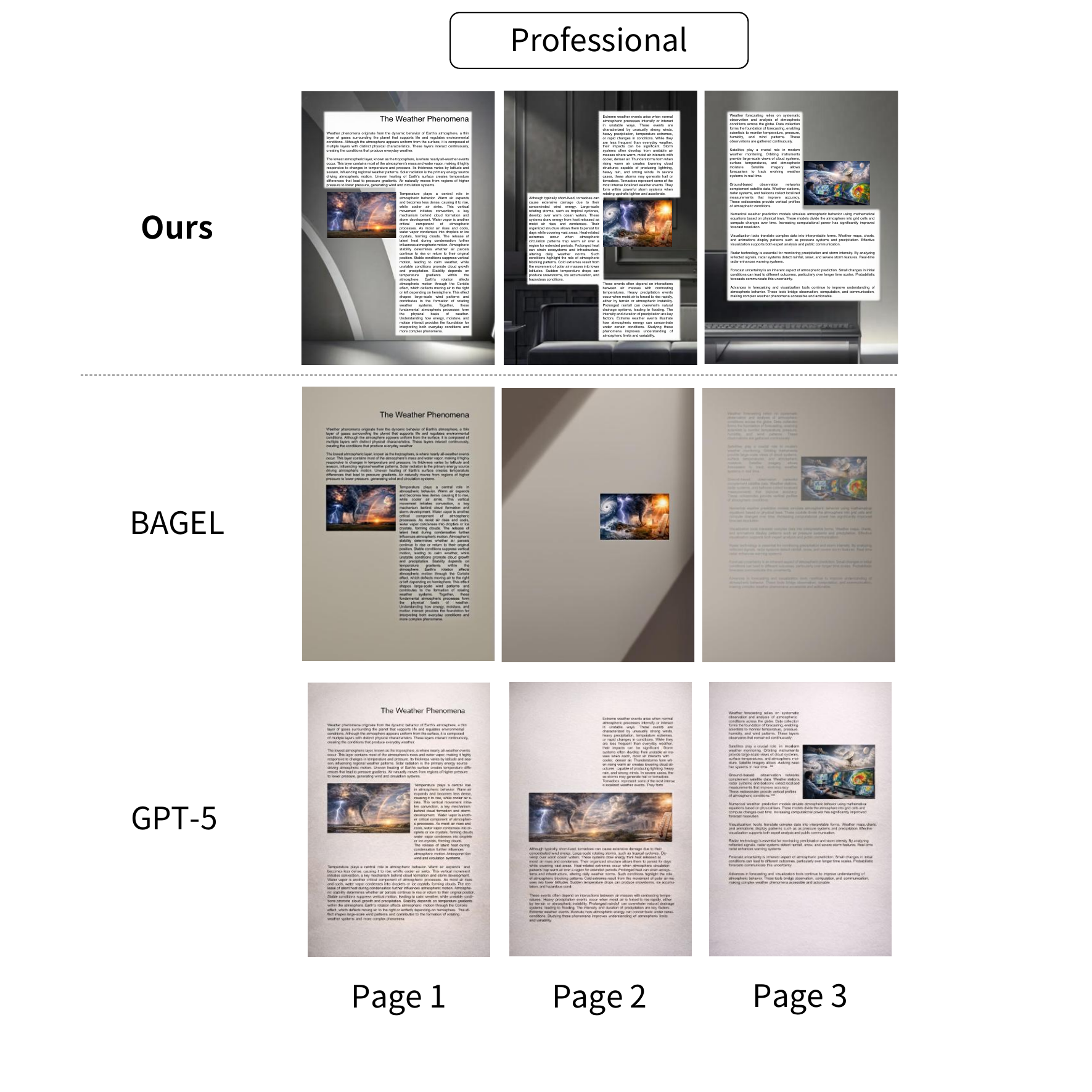}
    \caption{Qualitative comparison of background generation results for \emph{Professional}-style PDFs.}
\end{figure}

\newpage
\begin{figure}[t]
    \centering
    \includegraphics[width=\linewidth]{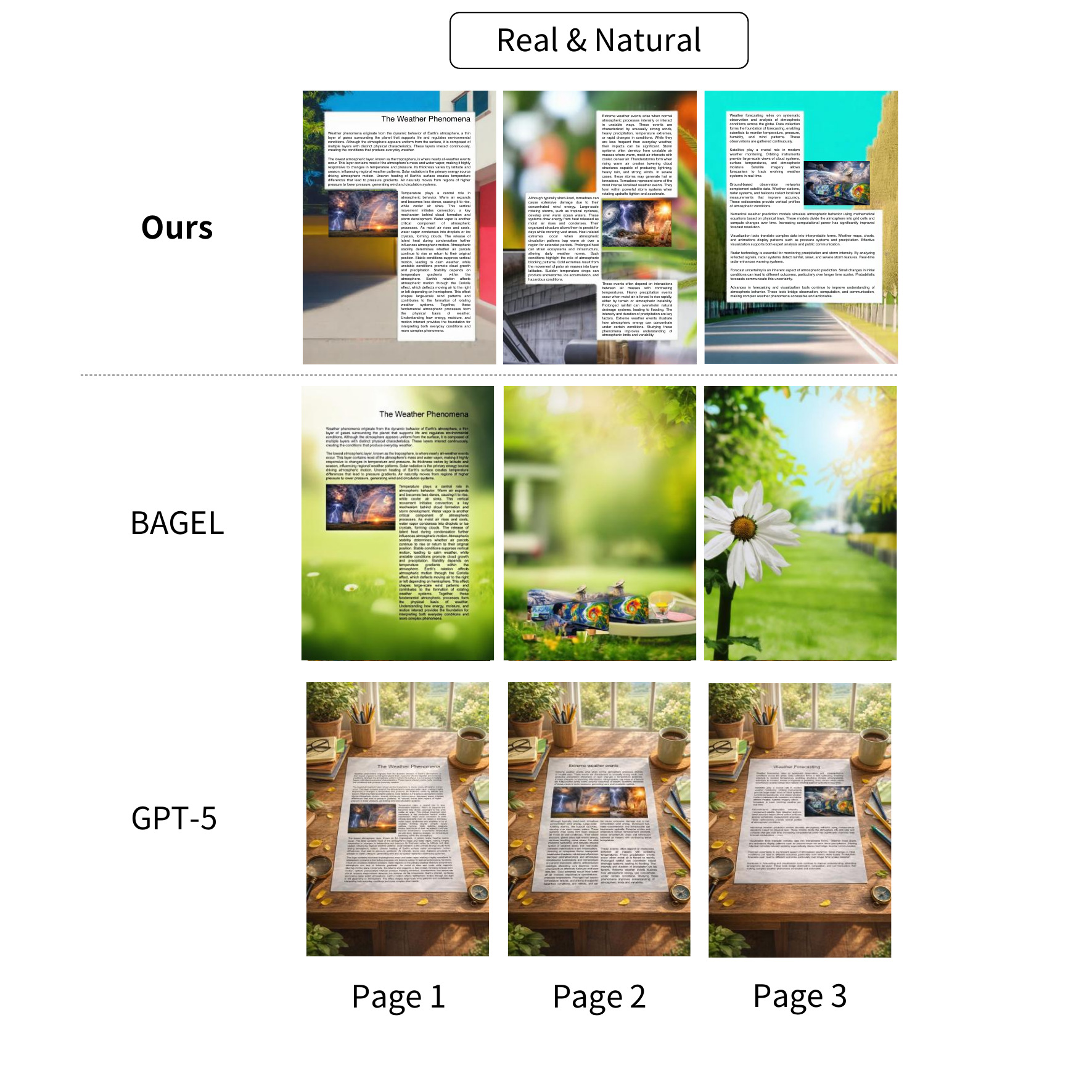}
    \caption{Qualitative comparison of background generation results for \emph{Real \& Natural}-style PDFs.}
\end{figure}

\newpage
\begin{figure}[t]
    \centering
    \includegraphics[width=\linewidth]{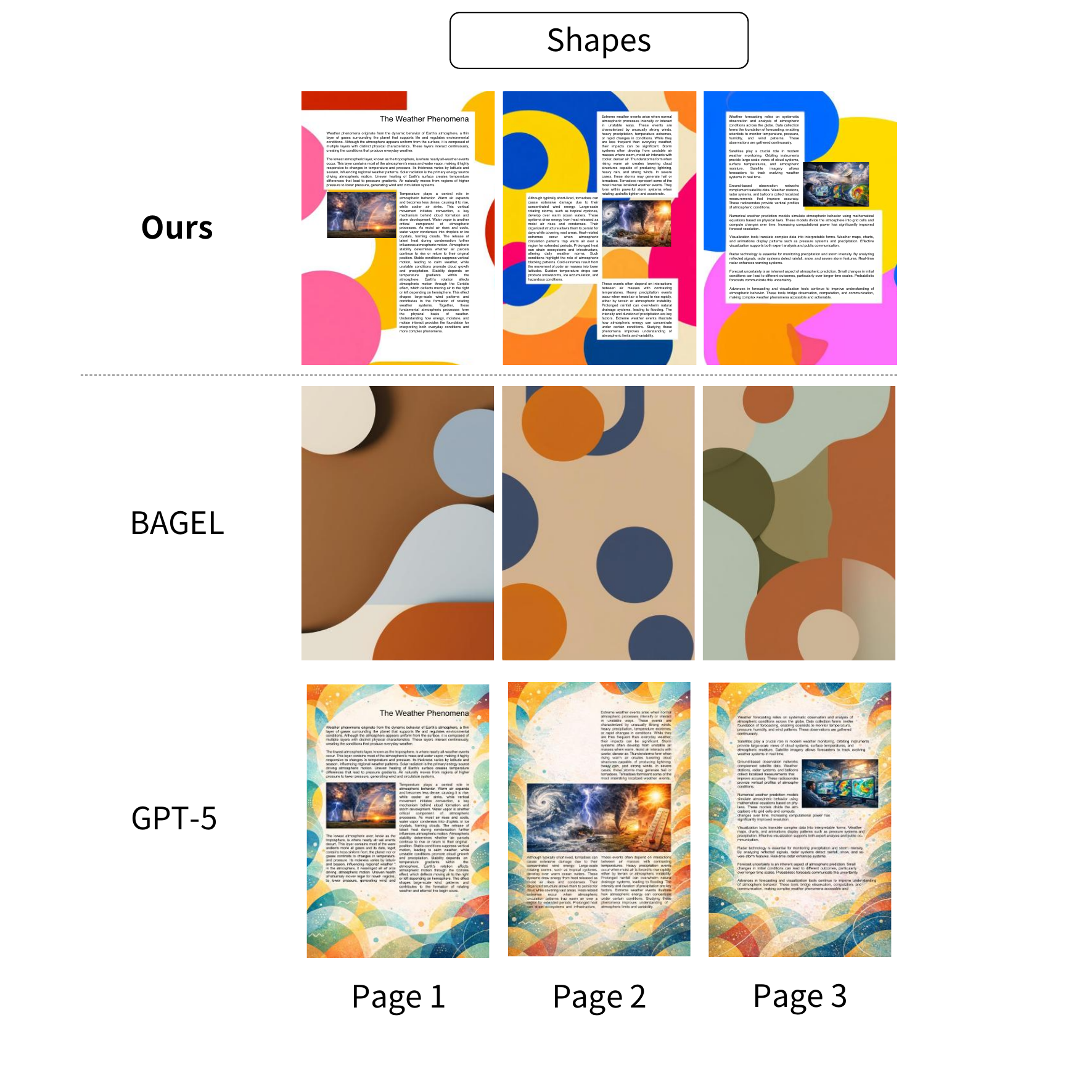}
    \caption{Qualitative comparison of background generation results for \emph{Shapes}-style PDFs.}
\end{figure}

\newpage
\begin{figure}[t]
    \centering
    \includegraphics[width=\linewidth]{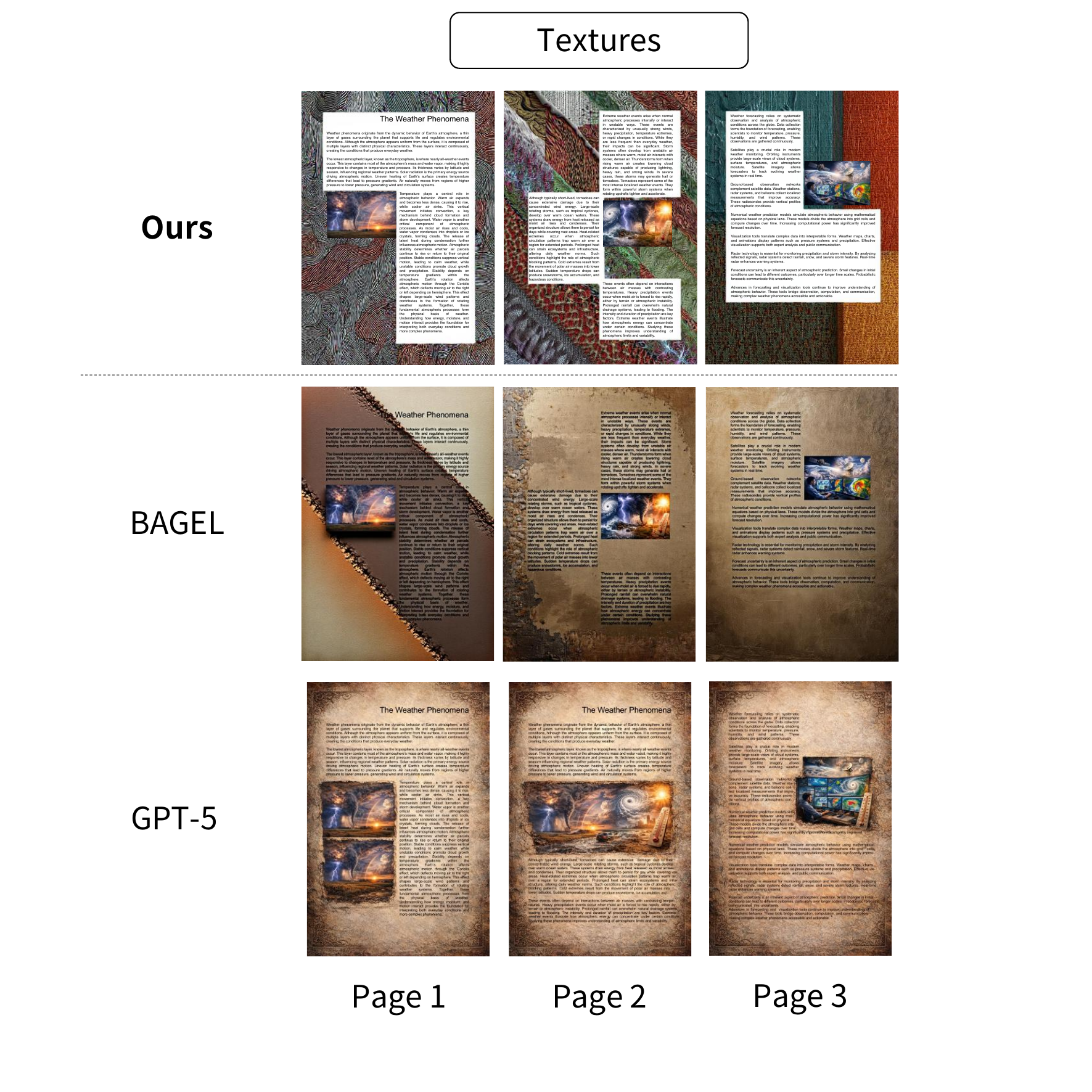}
    \caption{Qualitative comparison of background generation results for \emph{Textures}-style PDFs.}
\end{figure}

\newpage
\begin{figure}[t]
    \centering
    \includegraphics[width=\linewidth]{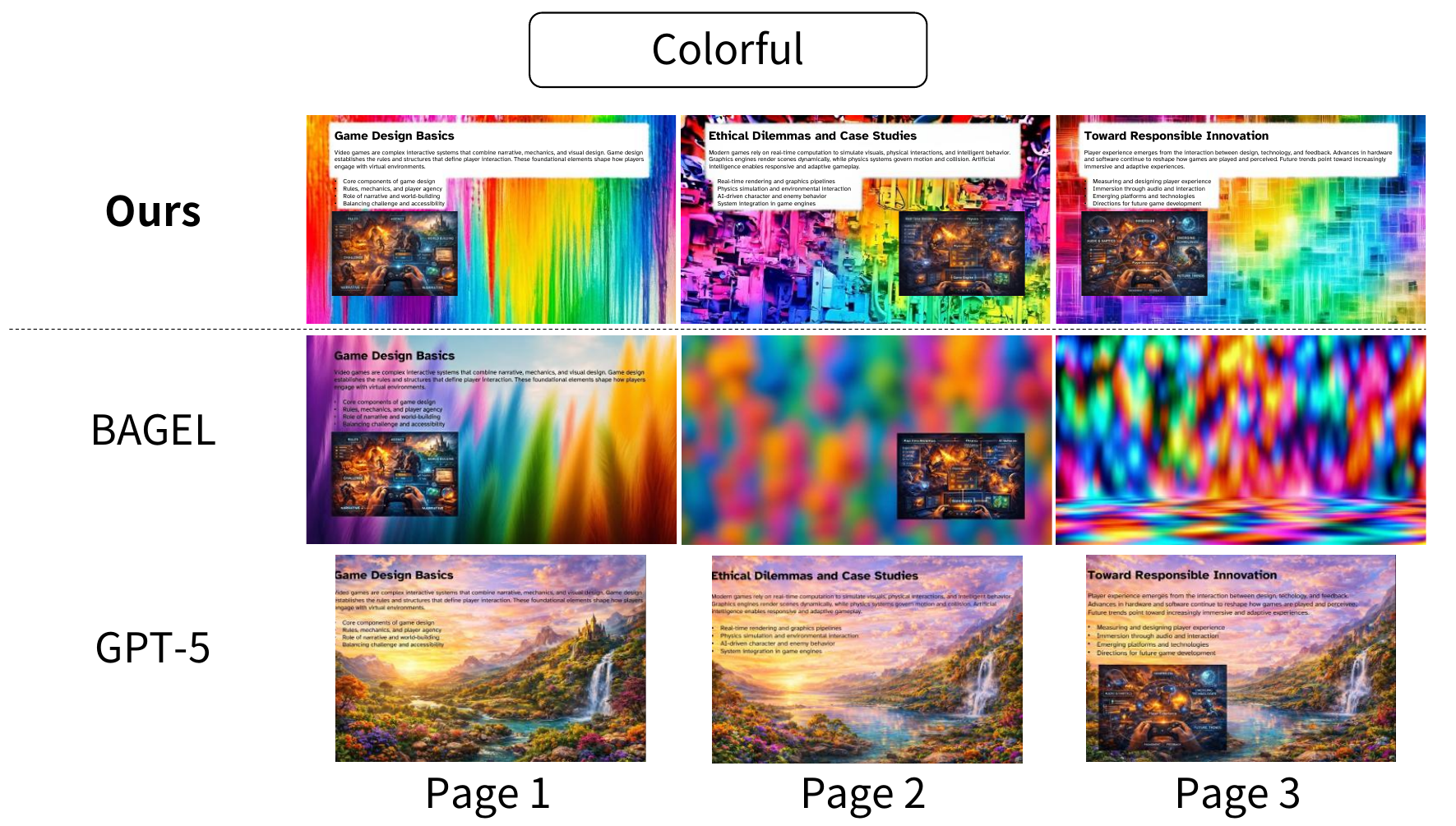}
    \caption{Qualitative comparison of background generation results for \emph{Colorful}-style slides.}
\end{figure}

\begin{figure}[t]
    \centering
    \includegraphics[width=\linewidth]{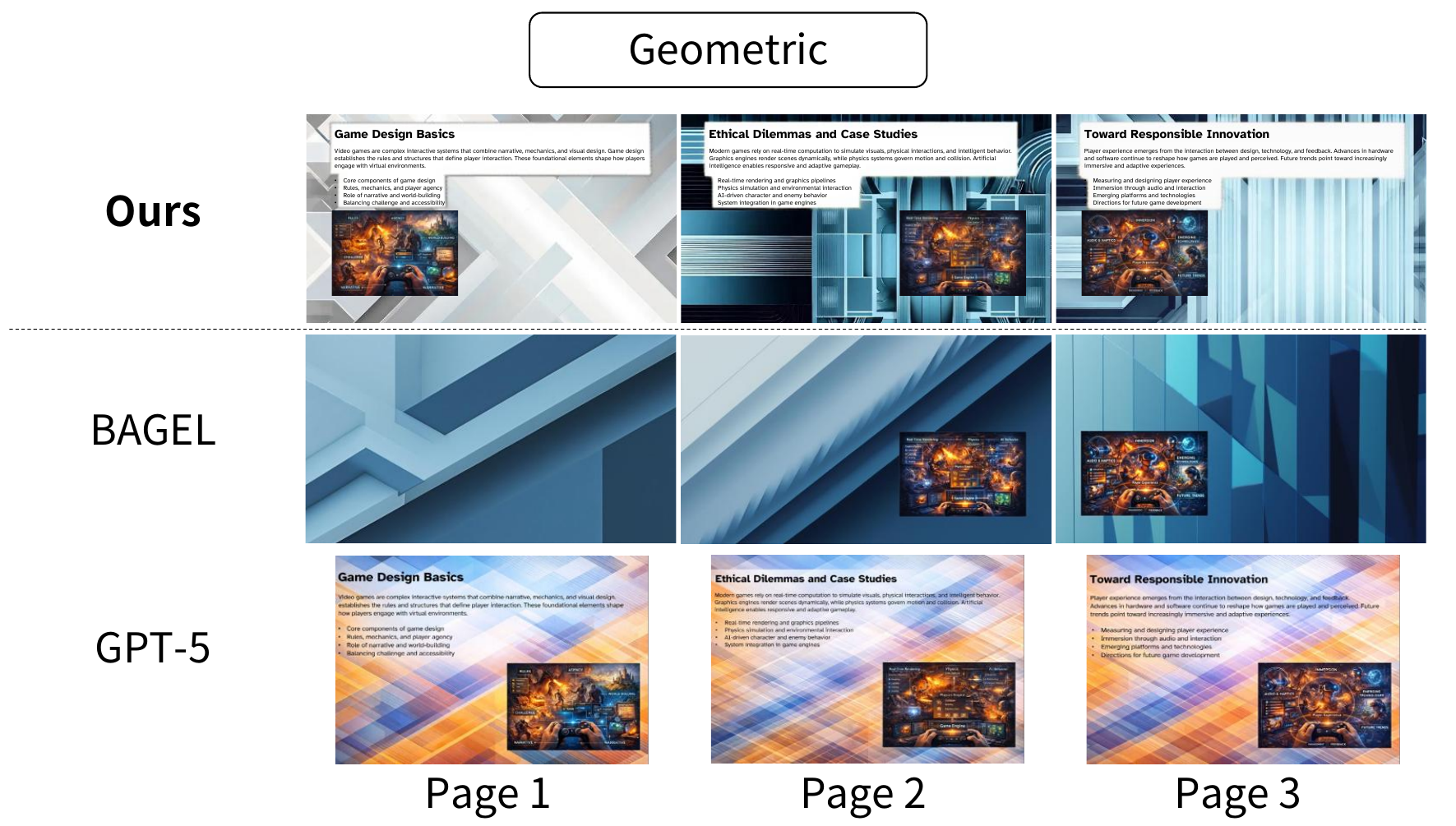}
    \caption{Qualitative comparison of background generation results for \emph{Geometric}-style slides.}
\end{figure}

\newpage
\begin{figure}[t]
    \centering
    \includegraphics[width=\linewidth]{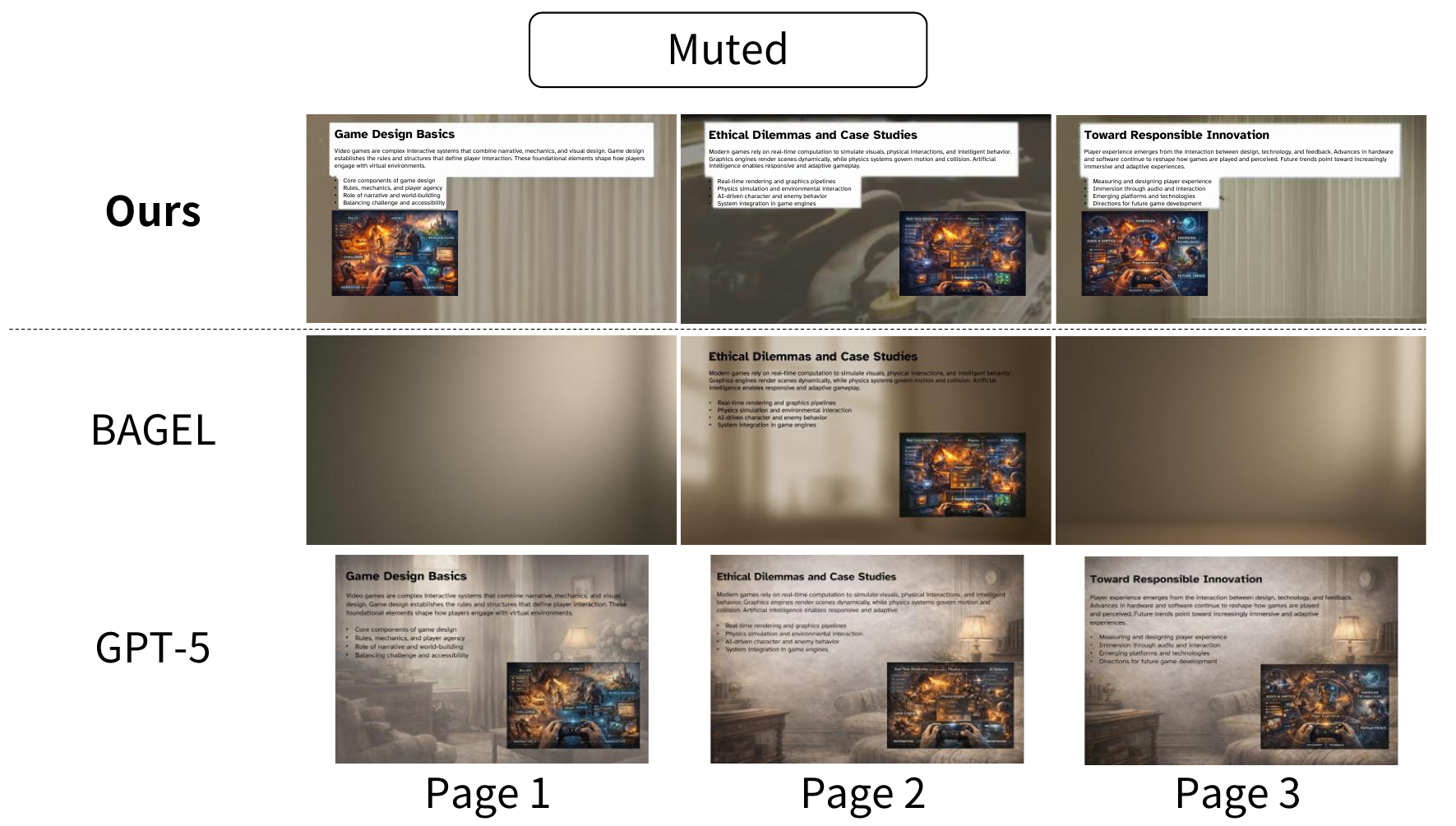}
    \caption{Qualitative comparison of background generation results for \emph{Muted}-style slides.}
\end{figure}

\begin{figure}[t]
    \centering
    \includegraphics[width=\linewidth]{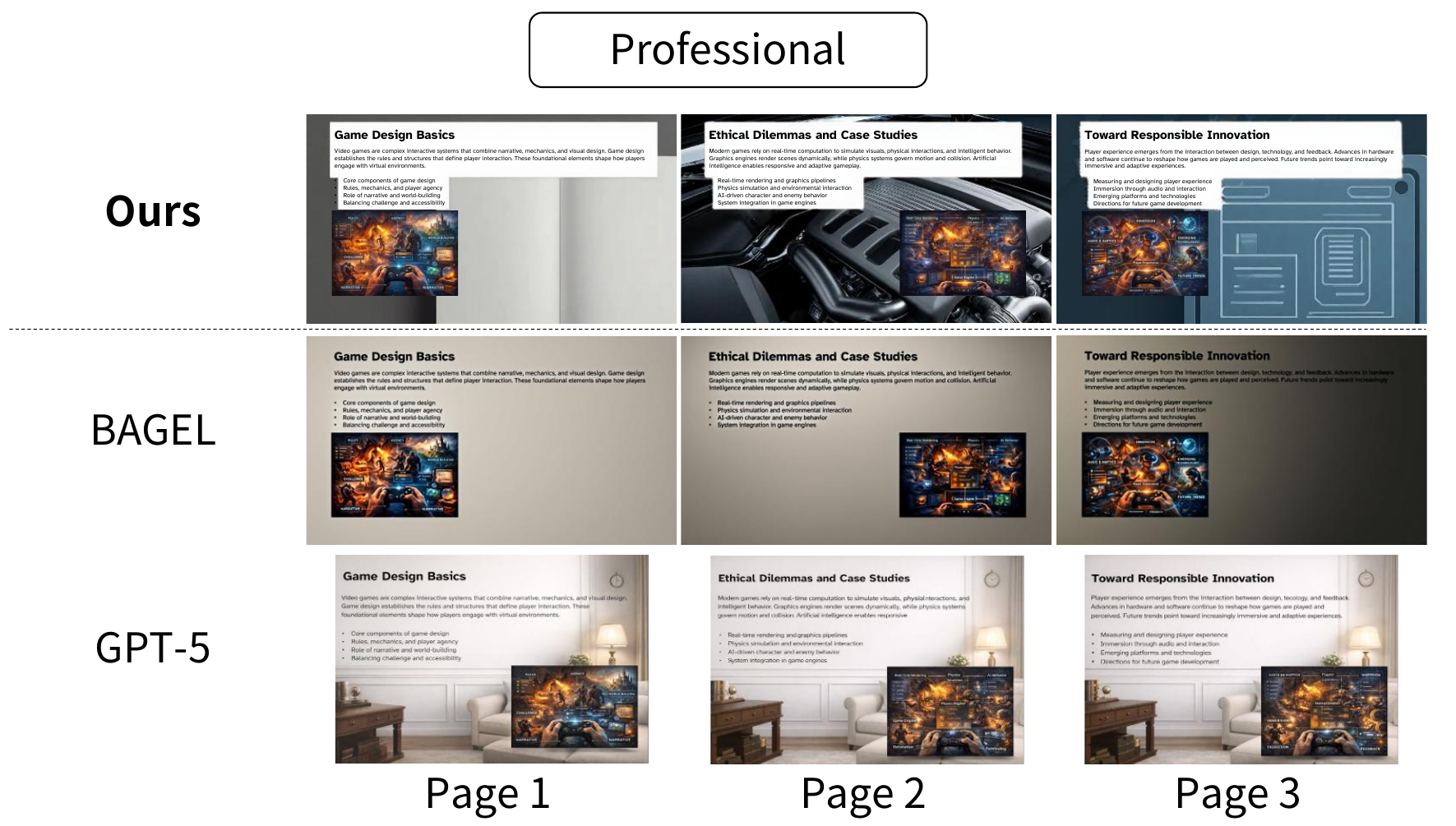}
    \caption{Qualitative comparison of background generation results for \emph{Professional}-style slides.}
\end{figure}

\newpage
\begin{figure}[t]
    \centering
    \includegraphics[width=\linewidth]{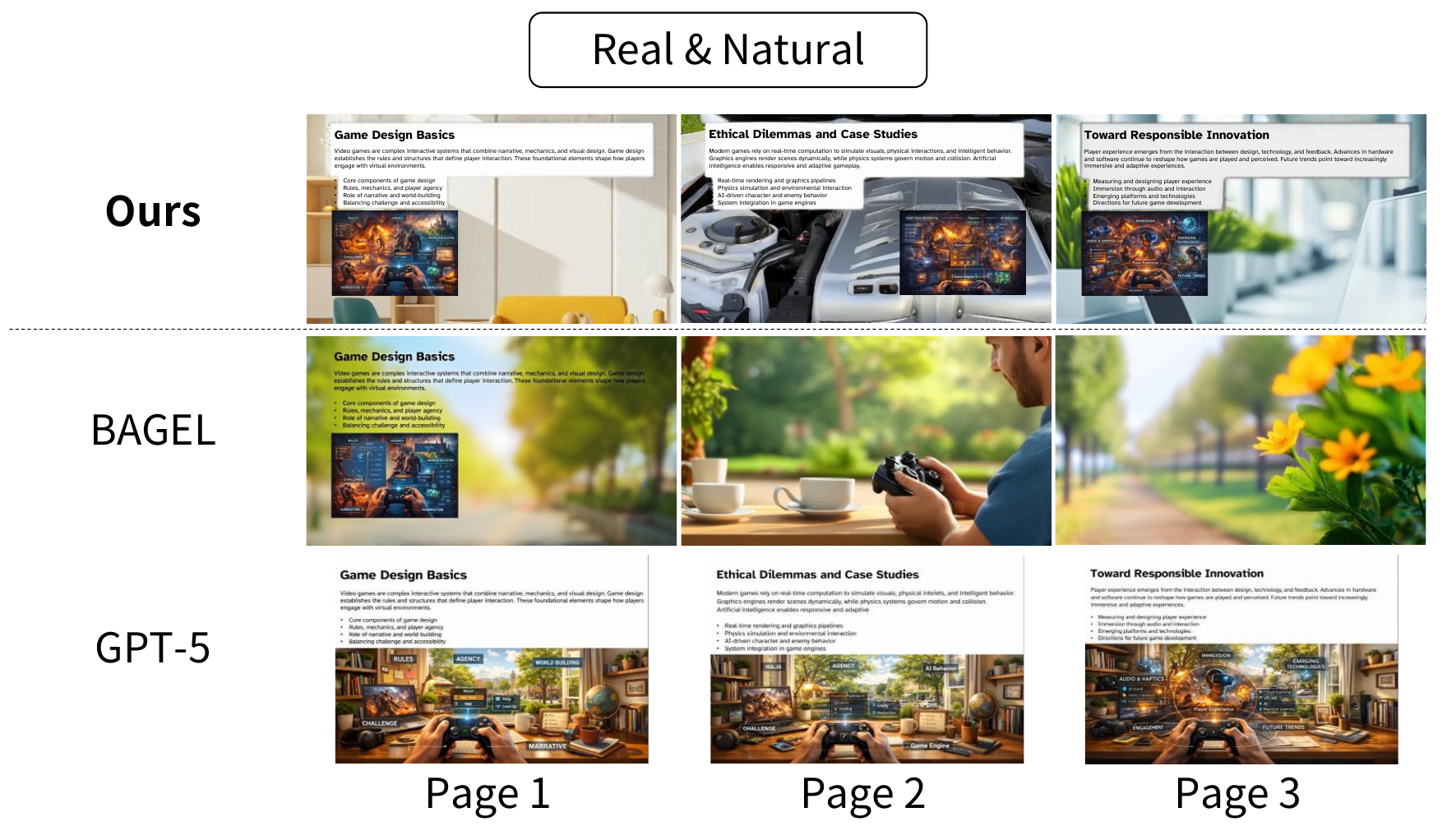}
    \caption{Qualitative comparison of background generation results for \emph{Real \& Natural}-style slides.}
\end{figure}

\begin{figure}[t]
    \centering
    \includegraphics[width=\linewidth]{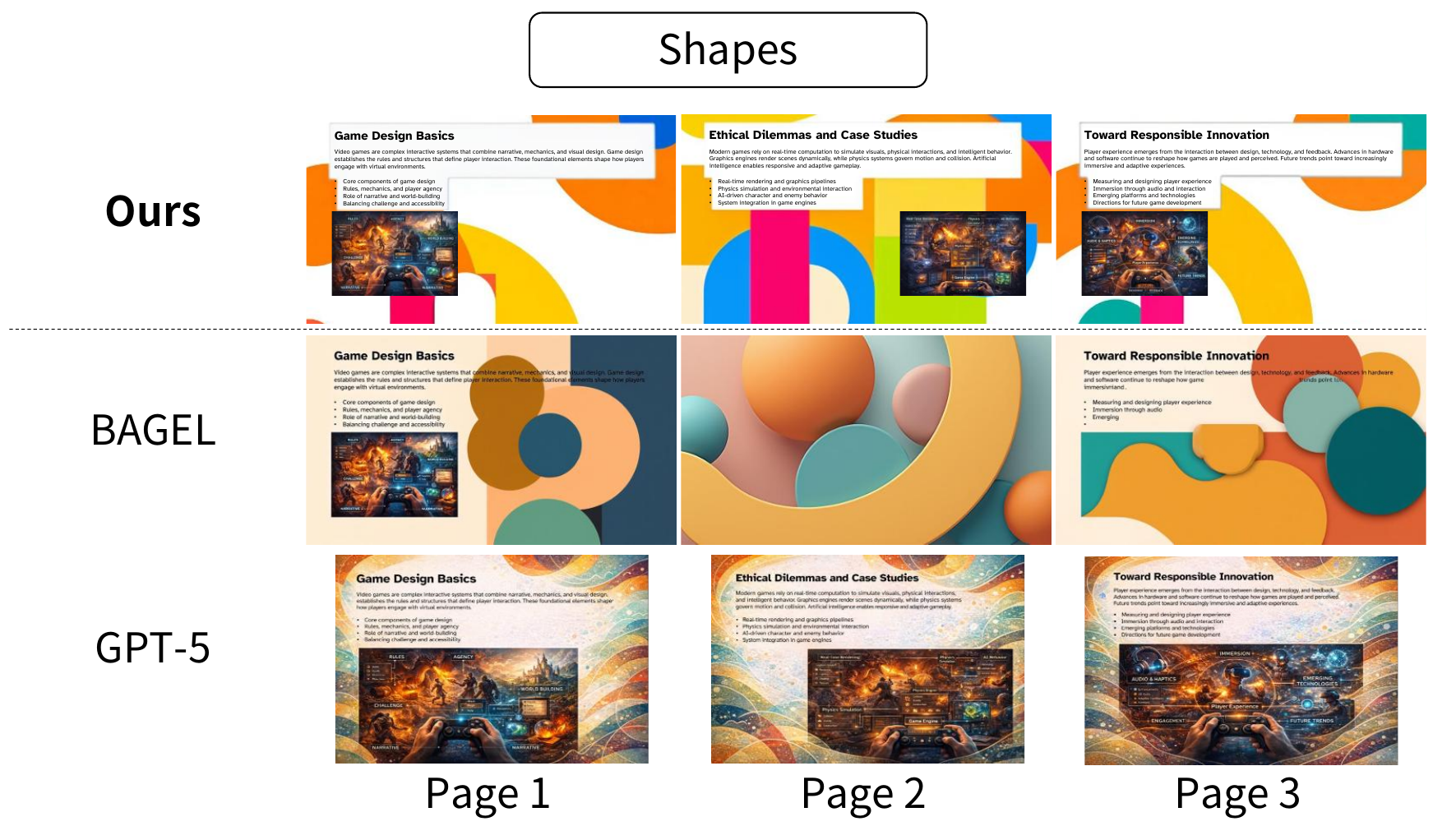}
    \caption{Qualitative comparison of background generation results for \emph{Shapes}-style slides.}
\end{figure}

\newpage
\begin{figure}[t]
    \centering
    \includegraphics[width=\linewidth]{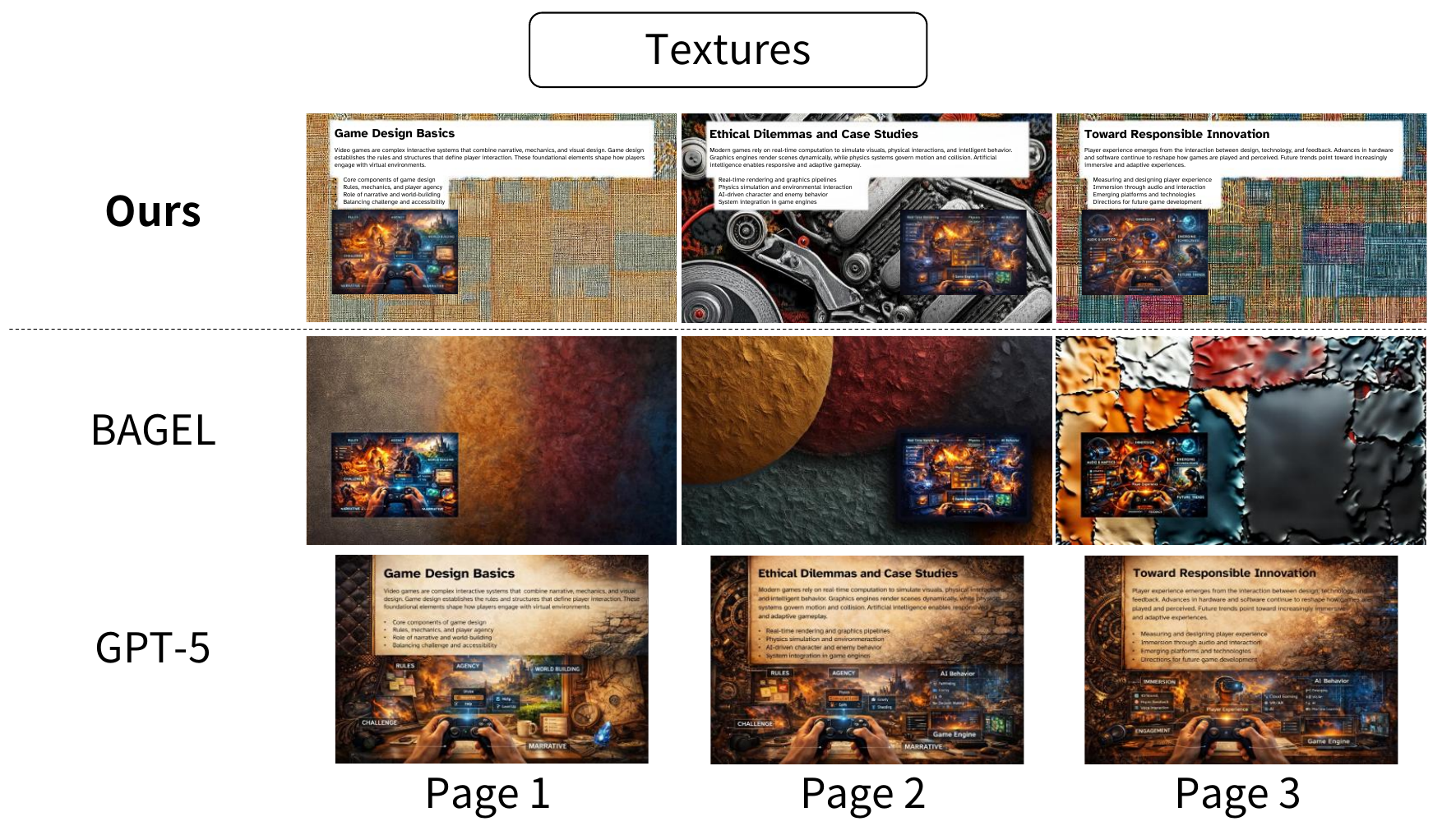}
    \caption{Qualitative comparison of background generation results for \emph{Textures}-style slides.}
\end{figure}

\newpage
\begin{figure}[t]
    \centering
    \includegraphics[width=\linewidth]{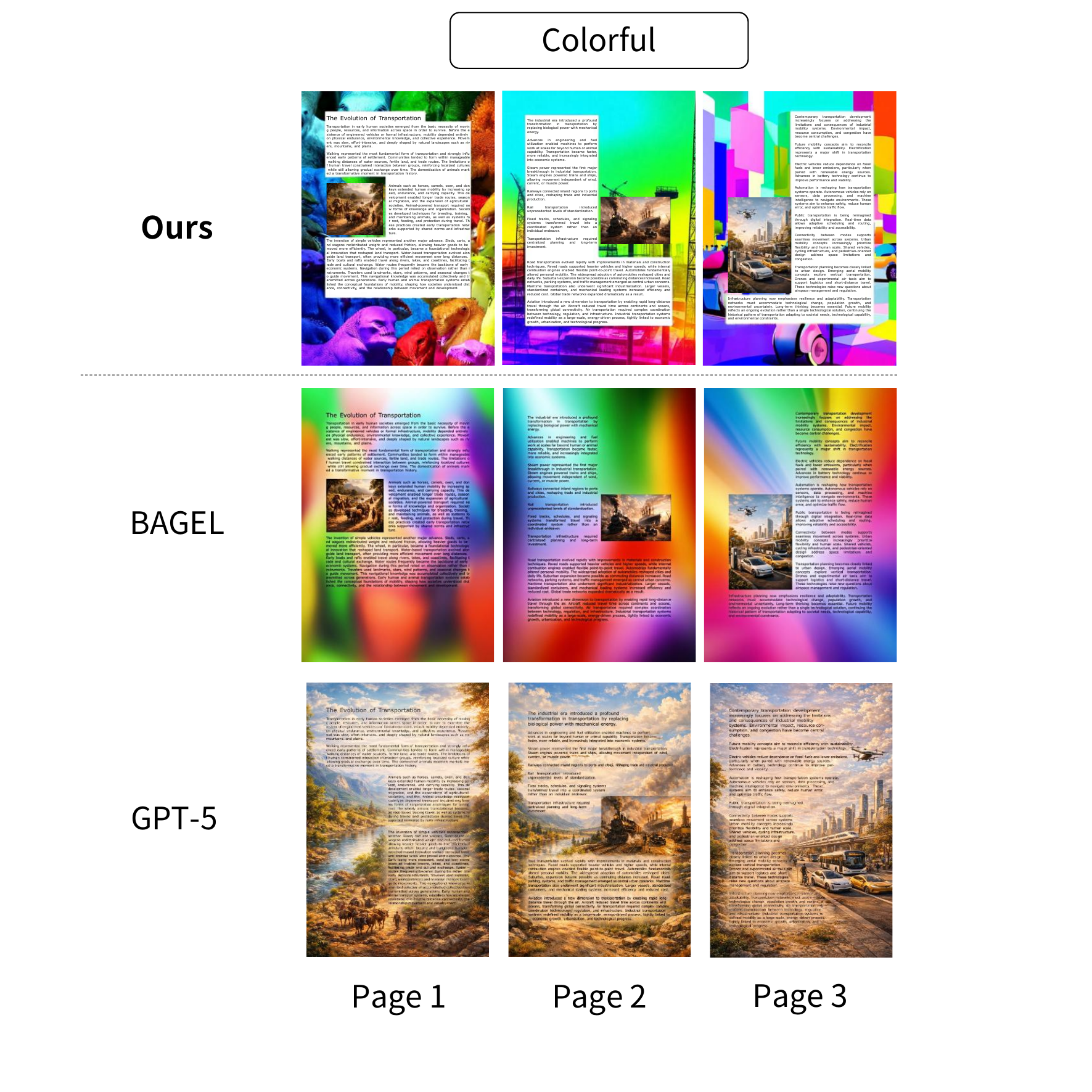}
    \caption{Qualitative comparison of background generation results for \emph{Colorful}-style PDFs.}
\end{figure}

\newpage
\begin{figure}[t]
    \centering
    \includegraphics[width=\linewidth]{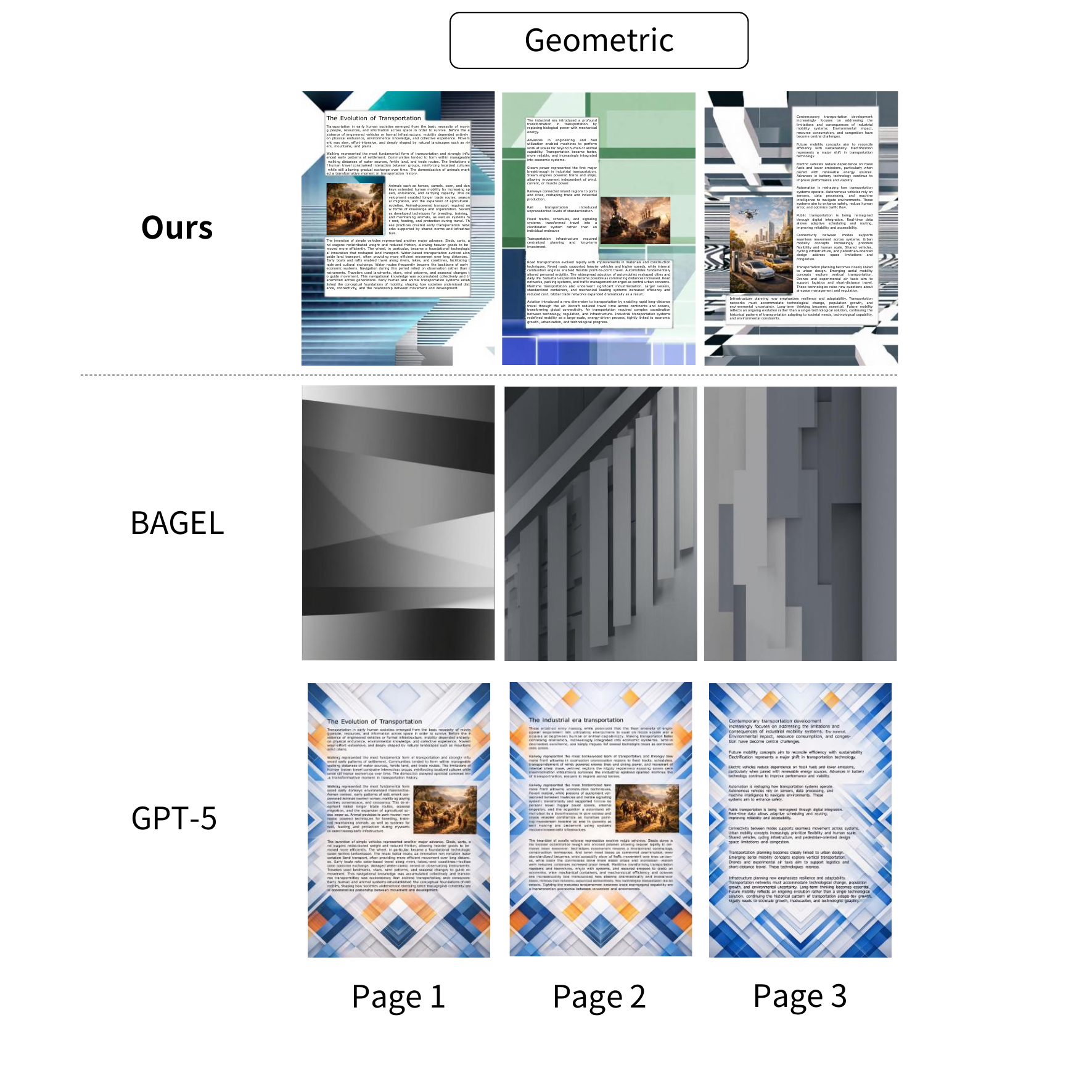}
    \caption{Qualitative comparison of background generation results for \emph{Geometric}-style PDFs.}
\end{figure}

\newpage
\begin{figure}[t]
    \centering
    \includegraphics[width=\linewidth]{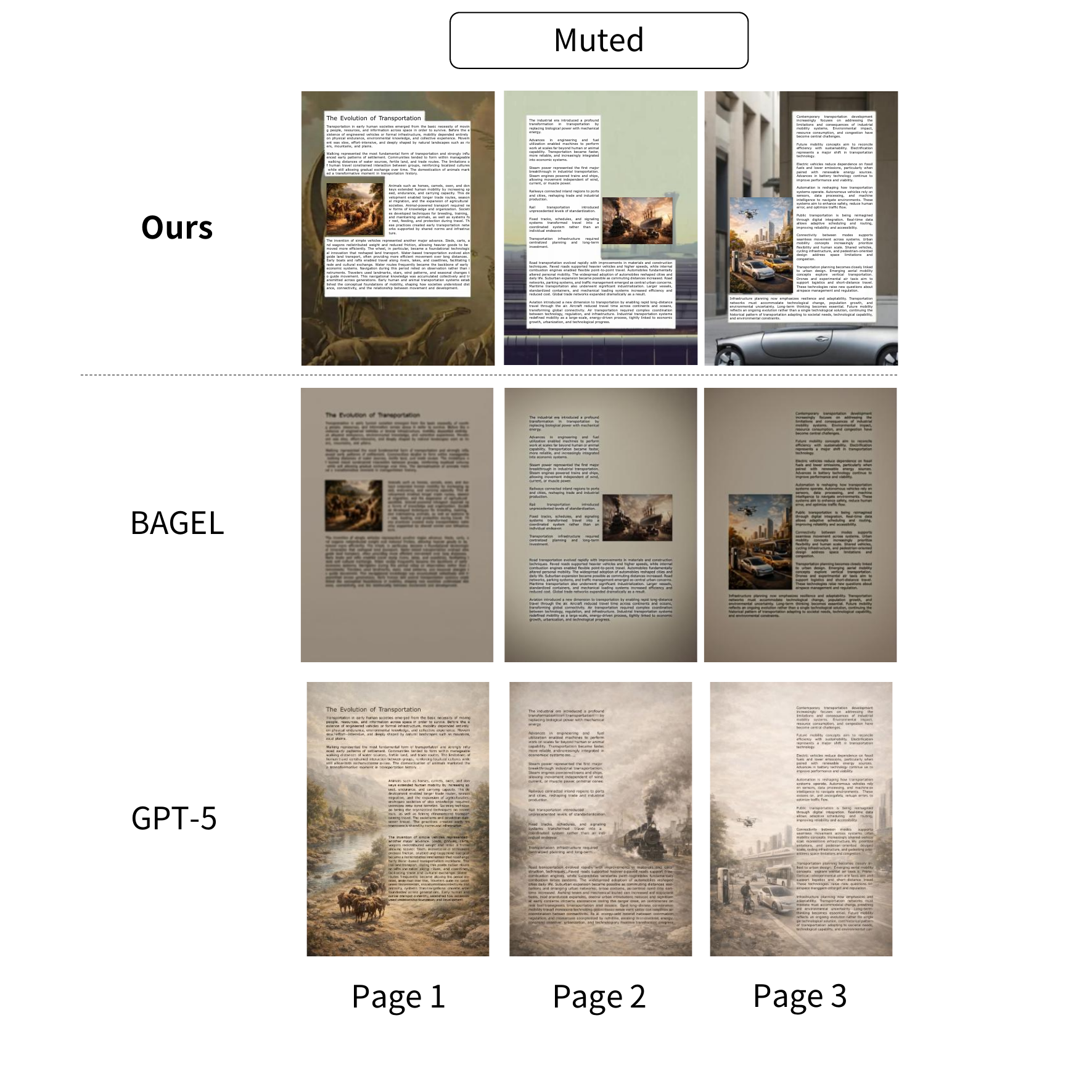}
    \caption{Qualitative comparison of background generation results for \emph{Muted}-style PDFs.}
\end{figure}

\newpage
\begin{figure}[t]
    \centering
    \includegraphics[width=\linewidth]{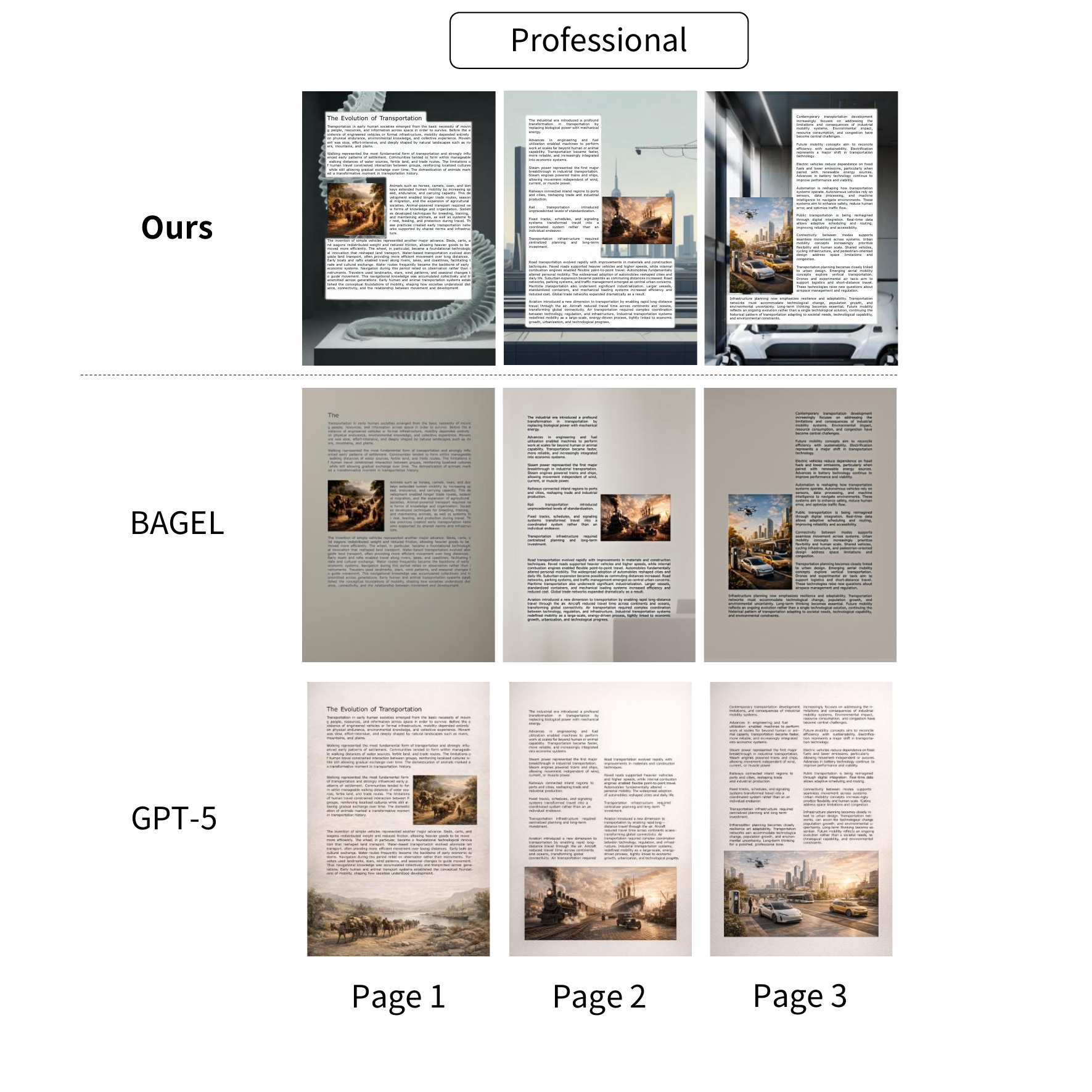}
    \caption{Qualitative comparison of background generation results for \emph{Professional}-style PDFs.}
\end{figure}

\newpage
\begin{figure}[t]
    \centering
    \includegraphics[width=\linewidth]{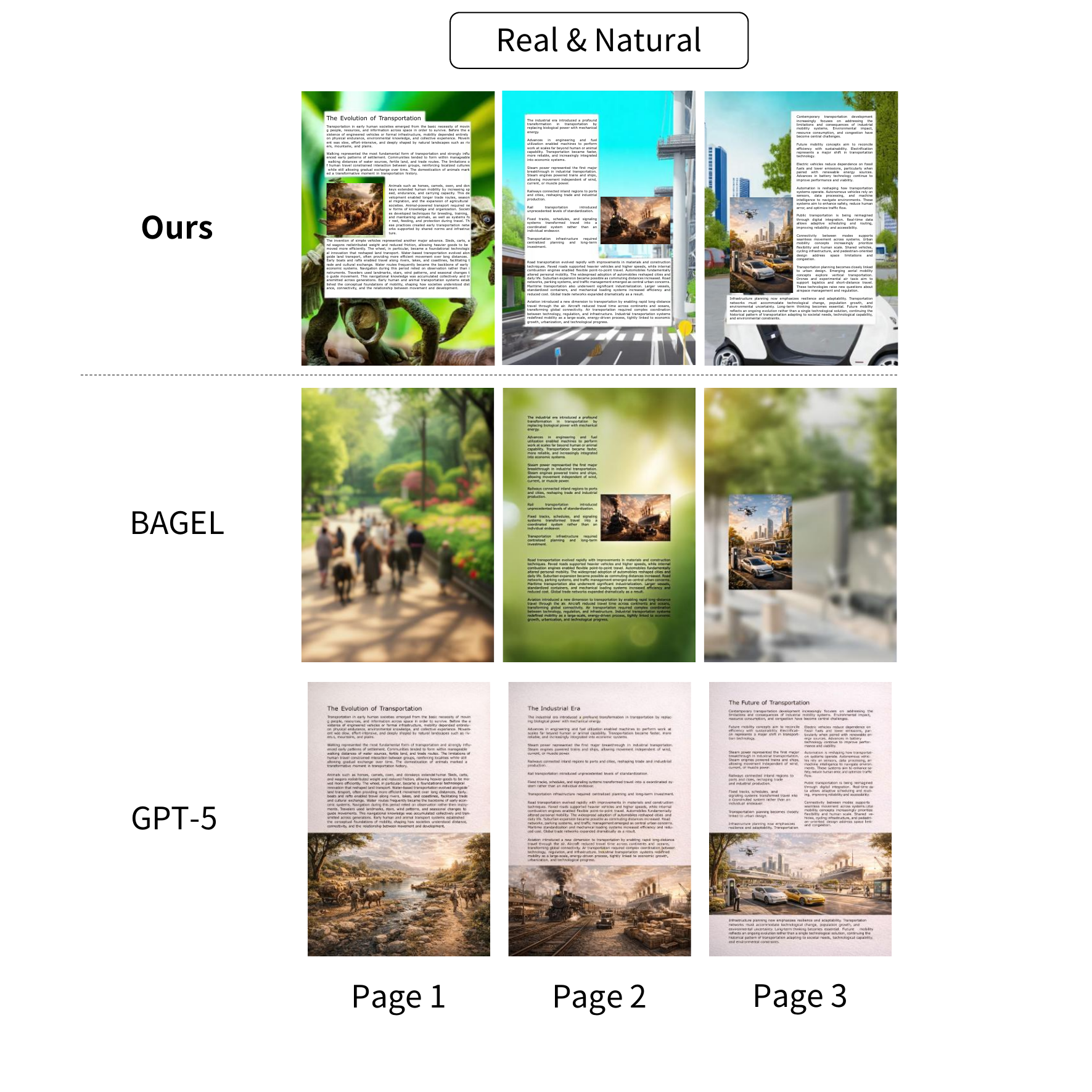}
    \caption{Qualitative comparison of background generation results for \emph{Real \& Natural}-style PDFs.}
\end{figure}

\newpage
\begin{figure}[t]
    \centering
    \includegraphics[width=\linewidth]{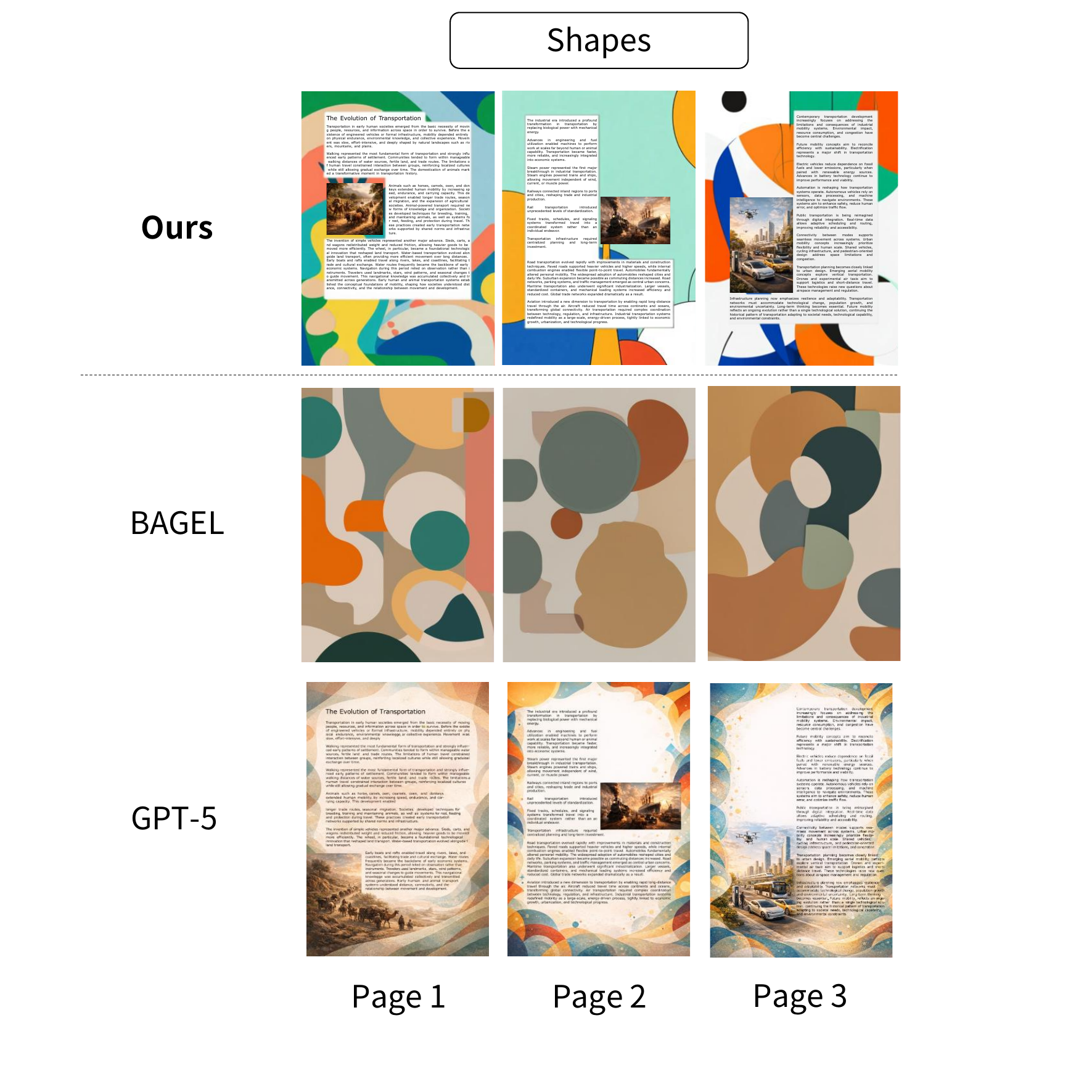}
    \caption{Qualitative comparison of background generation results for \emph{Shapes}-style PDFs.}
\end{figure}

\newpage
\begin{figure}[t]
    \centering
    \includegraphics[width=\linewidth]{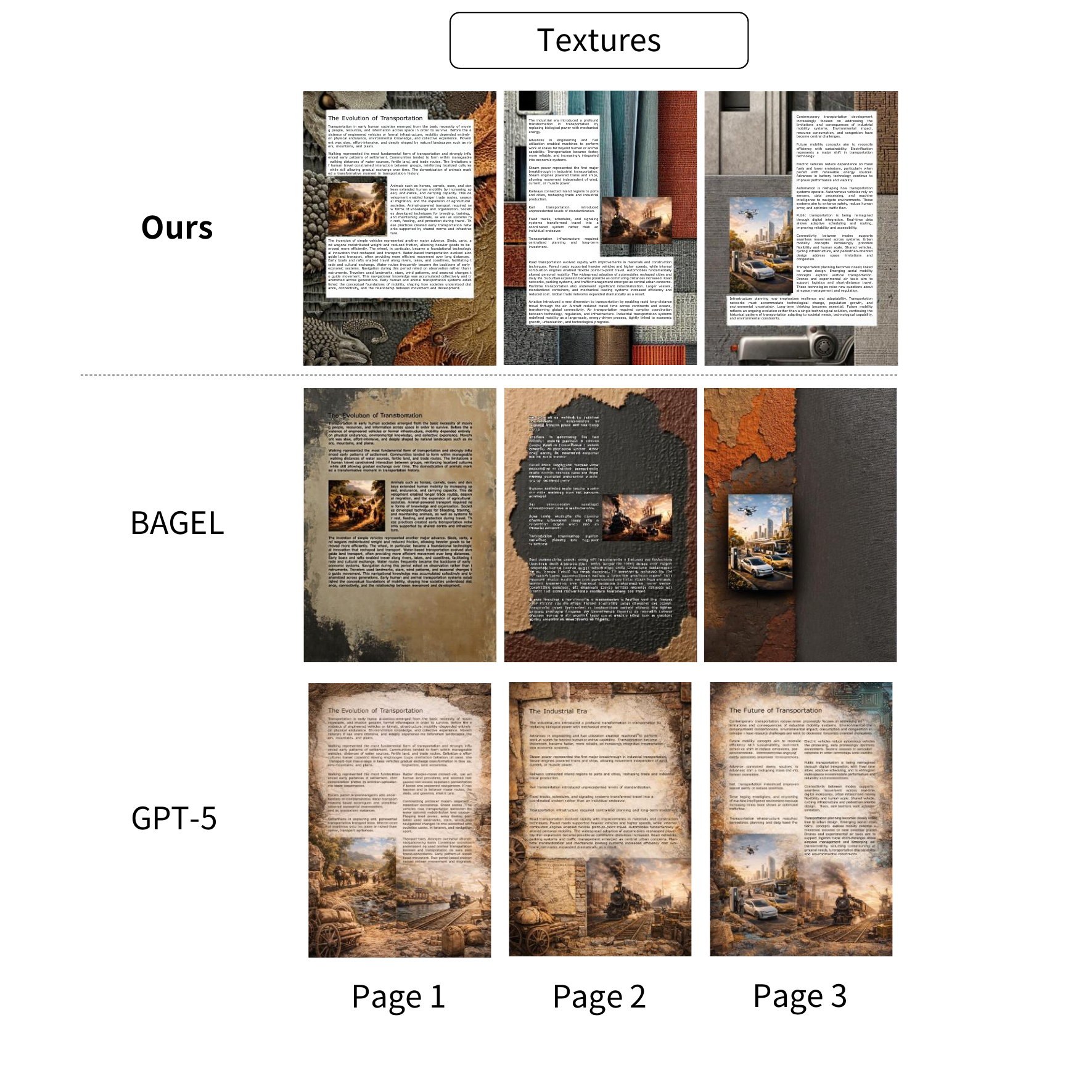}
    \caption{Qualitative comparison of background generation results for \emph{Textures}-style PDFs.}
\end{figure}

\newpage
\begin{figure}[t]
    \centering
    \includegraphics[width=\linewidth]{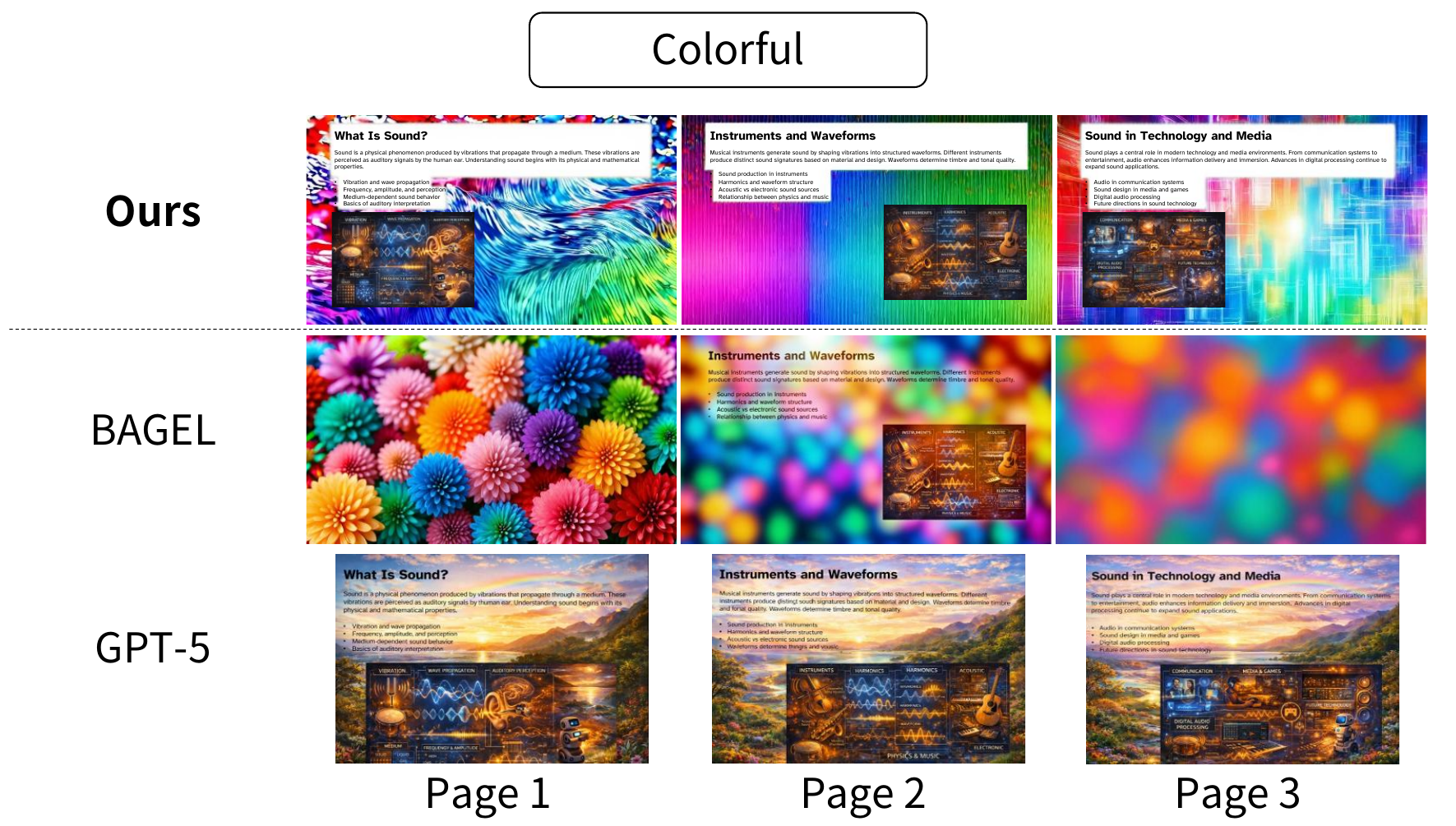}
    \caption{Qualitative comparison of background generation results for \emph{Colorful}-style slides.}
\end{figure}

\begin{figure}[t]
    \centering
    \includegraphics[width=\linewidth]{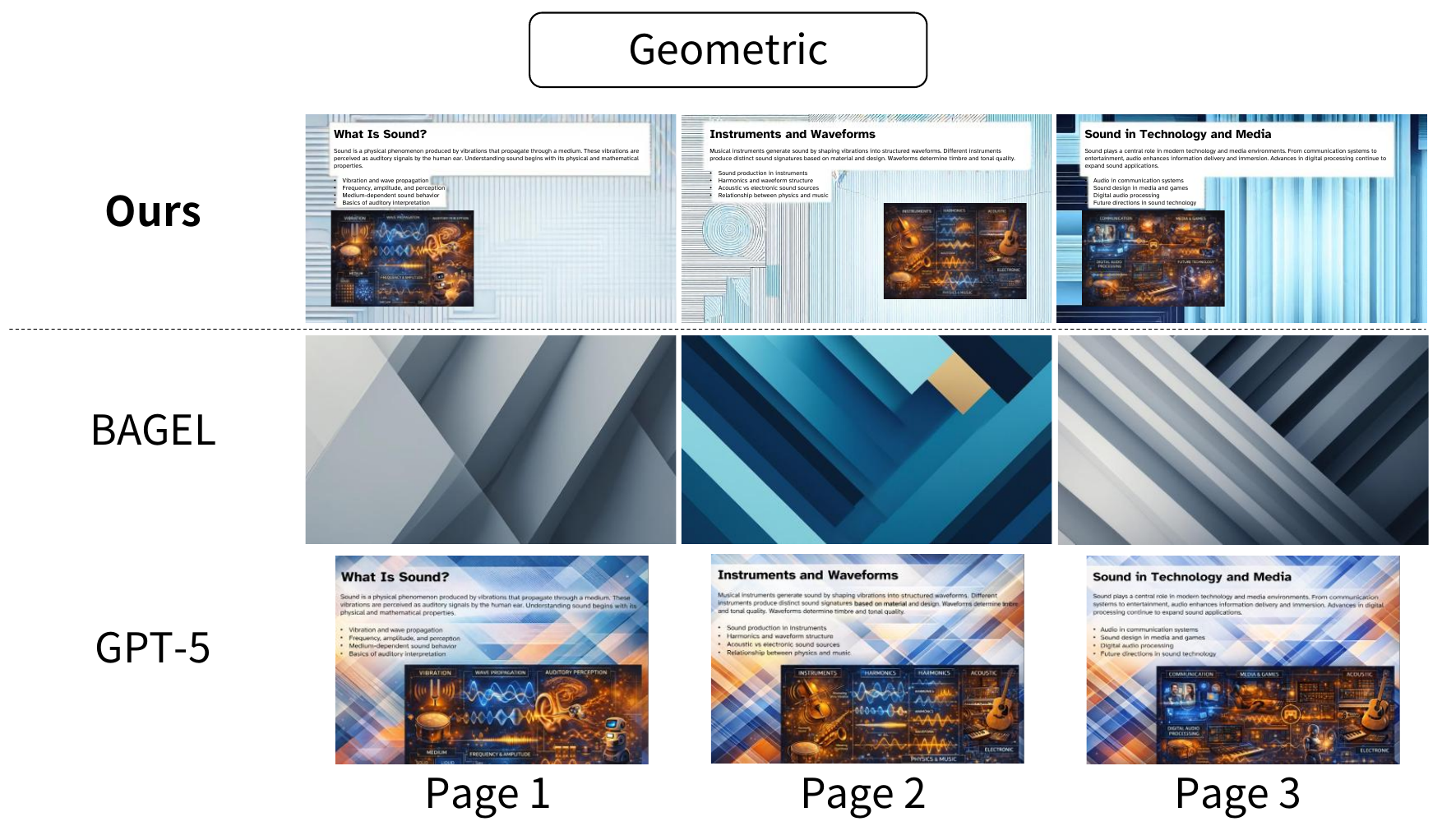}
    \caption{Qualitative comparison of background generation results for \emph{Geometric}-style slides.}
\end{figure}

\newpage
\begin{figure}[t]
    \centering
    \includegraphics[width=\linewidth]{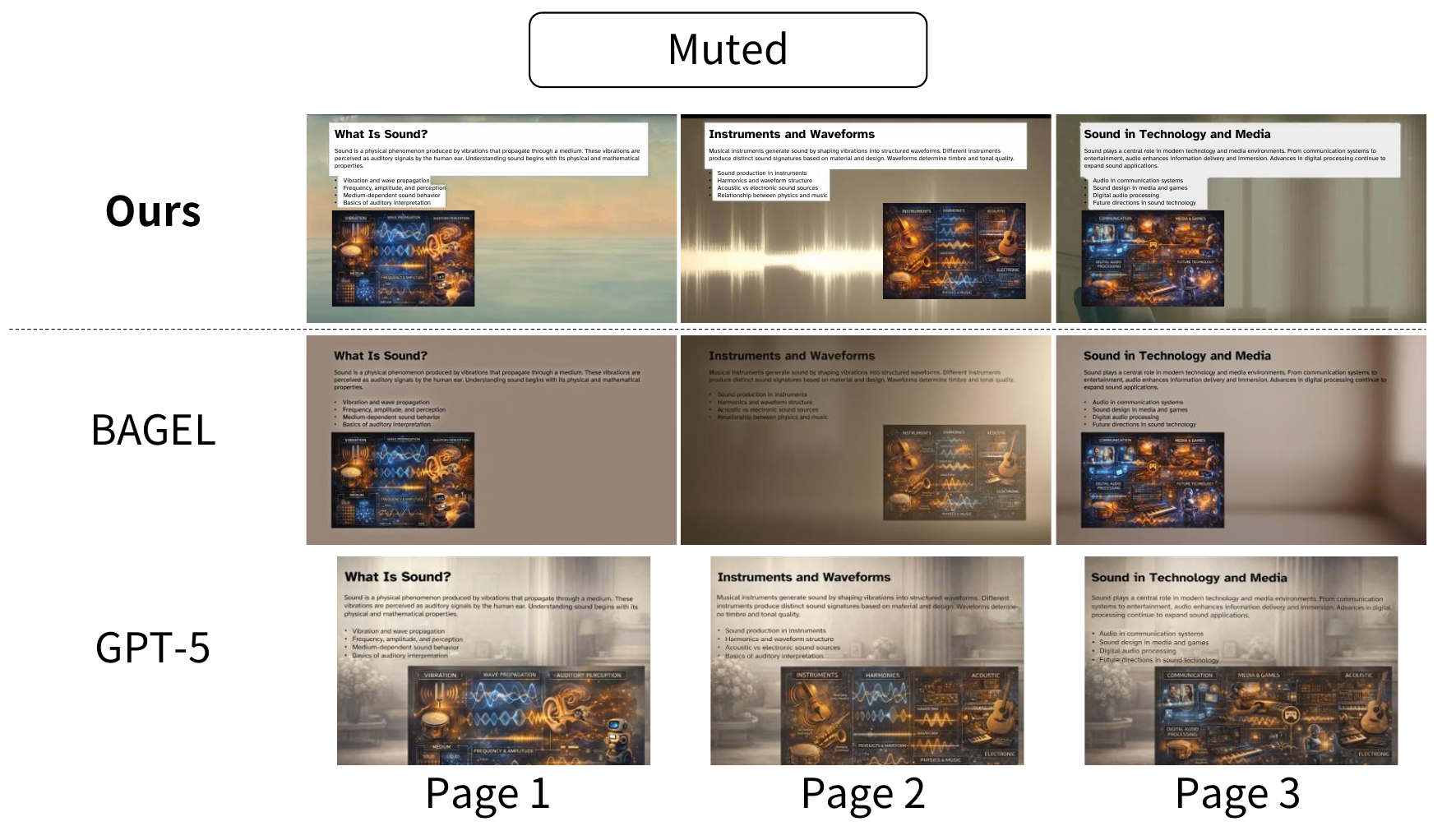}
    \caption{Qualitative comparison of background generation results for \emph{Muted}-style slides.}
\end{figure}

\begin{figure}[t]
    \centering
    \includegraphics[width=\linewidth]{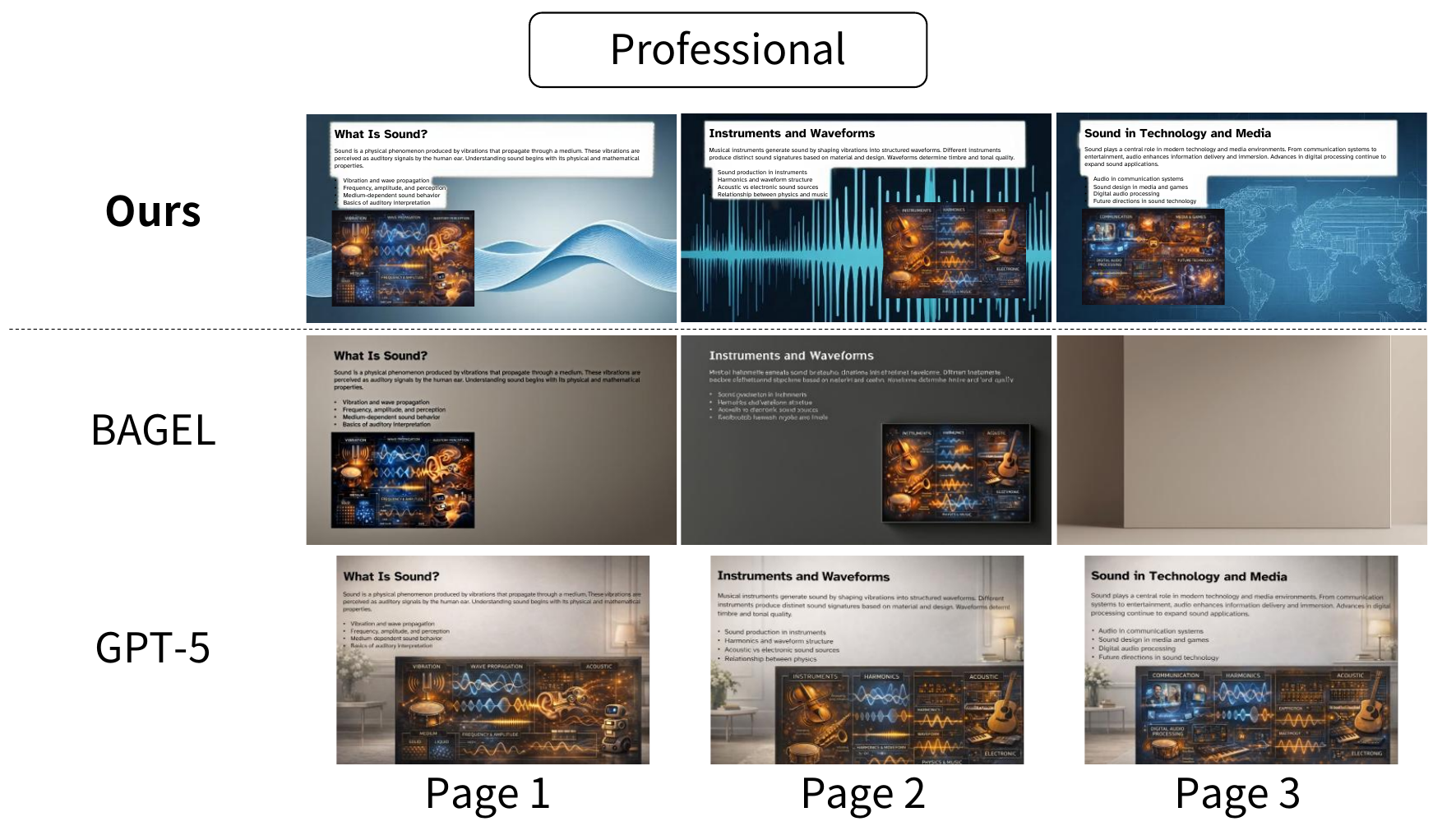}
    \caption{Qualitative comparison of background generation results for \emph{Professional}-style slides.}
\end{figure}

\newpage
\begin{figure}[t]
    \centering
    \includegraphics[width=\linewidth]{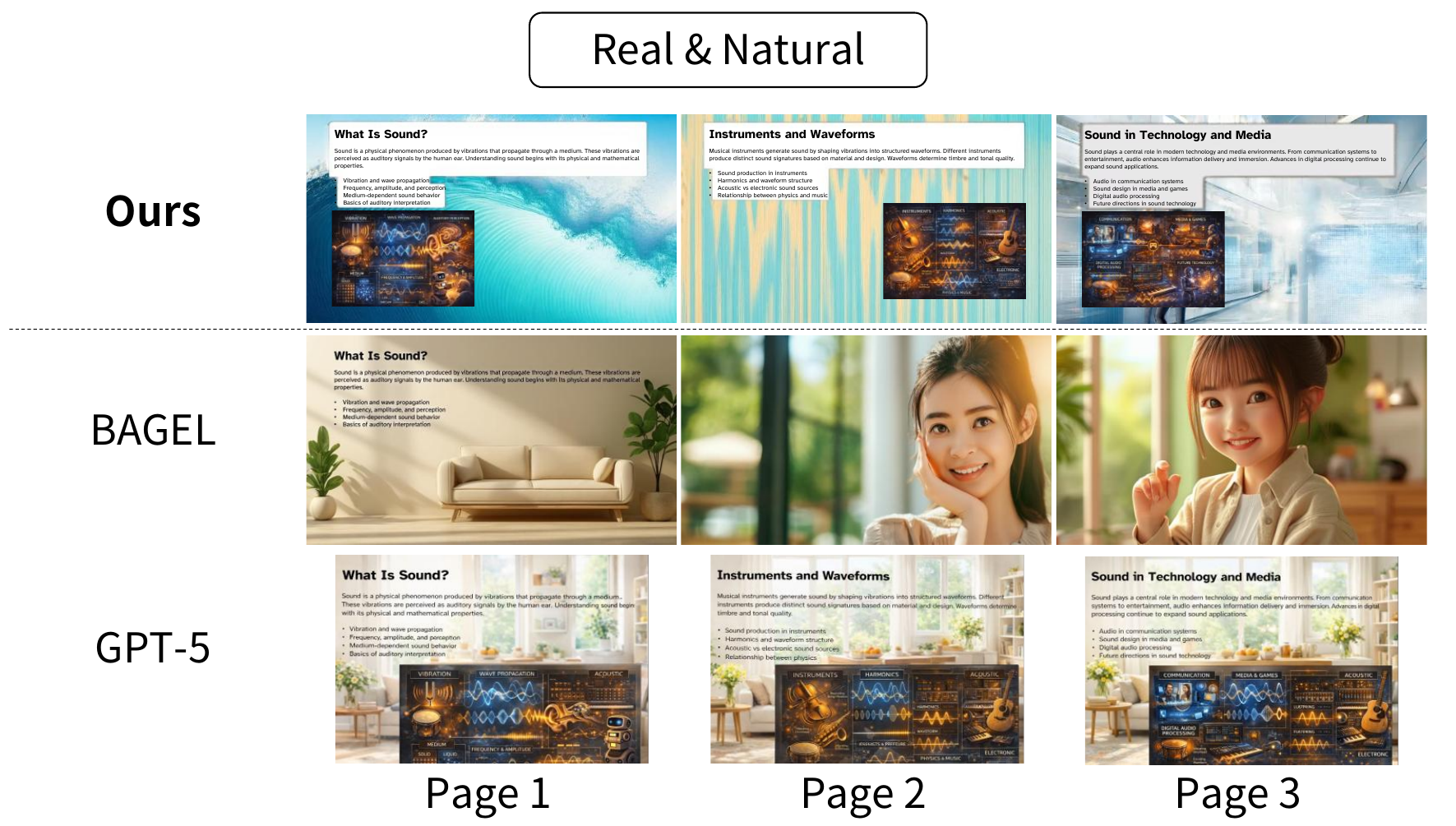}
    \caption{Qualitative comparison of background generation results for \emph{Real \& Natural}-style slides.}
\end{figure}

\begin{figure}[t]
    \centering
    \includegraphics[width=\linewidth]{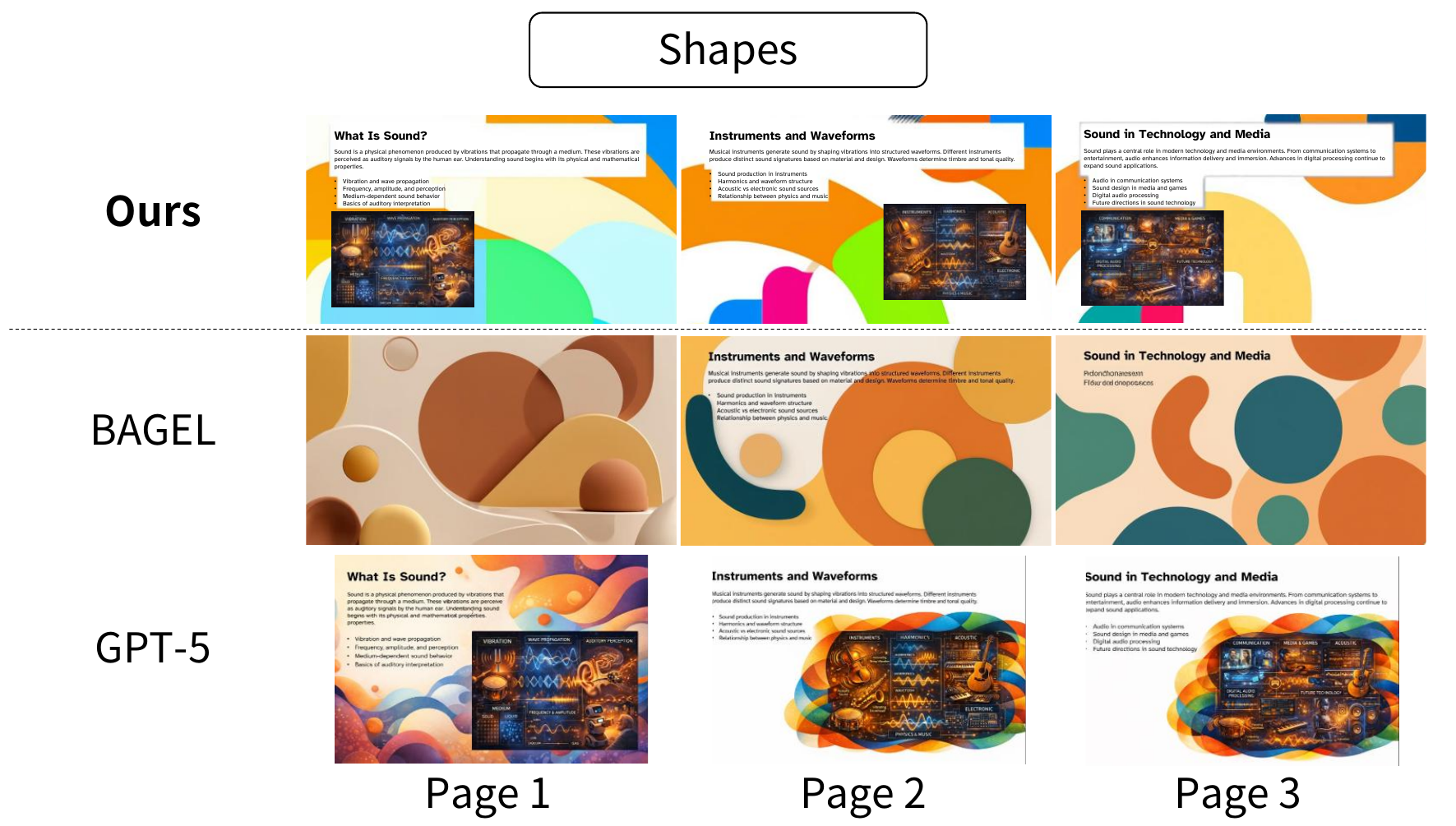}
    \caption{Qualitative comparison of background generation results for \emph{Shapes}-style slides.}
\end{figure}

\newpage
\begin{figure}[t]
    \centering
    \includegraphics[width=\linewidth]{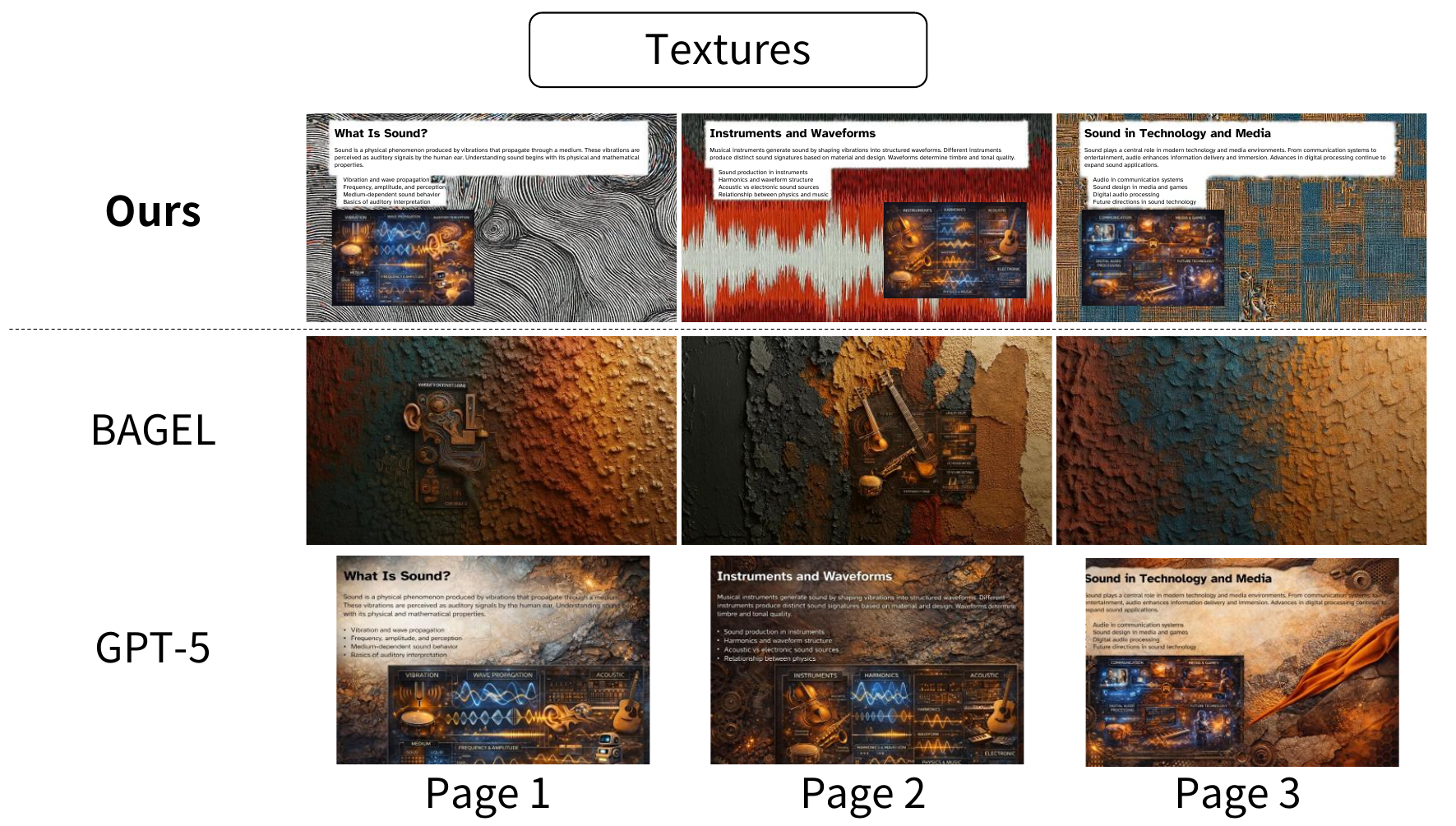}
    \caption{Qualitative comparison of background generation results for \emph{Textures}-style slides.}
\end{figure}

\end{document}